%% file: main.tex
\documentclass{article} 
\usepackage{iclr2026_conference,times}

\input{math_commands.tex}

\input{packages_macros.tex}

\title{\oursshort: Fast Optimization Through Whole Flow Processes for Training-Free Editing}

\author{Or Ronai, Vladimir Kulikov, Tomer Michaeli \\
Technion - Israel Institute of Technology \\
\texttt{\{or.ronai@campus, vladimir.k@campus, tomer.m@ee\}.technion.ac.il}
}

\iclrfinalcopy
\begin{document}

\maketitle

\input{sections/main_paper/abstract}
\input{sections/main_paper/introduction}
\input{sections/main_paper/realted_work}
\input{sections/main_paper/preliminaries}
\input{sections/main_paper/method}
\input{sections/main_paper/results}
\input{sections/main_paper/conclusions}
\input{sections/main_paper/ethics_statement}
\input{sections/main_paper/acknowledgements}

\bibliography{iclr2026_conference}
\bibliographystyle{iclr2026_conference}

\clearpage
\beginsupplement
\appendix
\mySuppTitle
\input{sections/supplementary/additional_results}
\input{sections/supplementary/comparisons}
\input{sections/supplementary/initialization}
\input{sections/supplementary/sd3}
\input{sections/supplementary/other_losses}
\input{sections/supplementary/contraction_mapping}
\input{sections/supplementary/ddim_coefficients}
\input{sections/supplementary/teaser_hyperparameters}
\input{sections/supplementary/editing_by_inversion}
\input{sections/supplementary/limitations}

\end{document}

%% file: math_commands.tex
\usepackage{amsmath,amsfonts,bm}

\def\eqref#1{Eq.~(\ref{#1})}

\def\1{\bm{1}}

\DeclareMathAlphabet{\mathsfit}{\encodingdefault}{\sfdefault}{m}{sl}
\SetMathAlphabet{\mathsfit}{bold}{\encodingdefault}{\sfdefault}{bx}{n}

\newcommand{\normltwo}{L^2}

\newcommand{\norm}[1]{\left\lVert#1\right\rVert}

\DeclareMathOperator*{\argmin}{arg\,min}

%% file: packages_macros.tex
\usepackage[utf8]{inputenc} 
\usepackage[T1]{fontenc}    
\usepackage[
  pagebackref=true,
  colorlinks=true,
  linkcolor=red,
  citecolor=Blue,
  urlcolor=magenta
]{hyperref}       
\usepackage{url}            
\usepackage{booktabs}       
\usepackage{amsfonts}       
\usepackage{nicefrac}       
\usepackage{microtype}      
\usepackage[dvipsnames]{xcolor}         
\usepackage{graphicx}
\usepackage{amsmath,bm,physics,mathtools,amssymb,amsthm}
\usepackage{wrapfig}
\usepackage{placeins}
\usepackage{enumitem}
\usepackage{subcaption}
\usepackage{xr}
\usepackage[algo2e,ruled,vlined]{algorithm2e}
\usepackage[labelfont=bf,textfont=md]{caption}

\newcommand{\oursshort}{FlowOpt}
\newcommand{\ourslong}{Flow Zero-Order Optimization}

\newcommand{\mySuppTitle}{
    {\vbox{\hsize\textwidth
    {\LARGE\sc 
    Supplementary Material\par}
    \vskip 0.3in minus 0.1in}}
}

\makeatletter
\DeclareRobustCommand\onedot{\futurelet\@let@token\@onedot}
\def\@onedot{\ifx\@let@token.\else.\null\fi\xspace}
\def\eg{\emph{e.g}\onedot, } 
\def\ie{\emph{i.e}\onedot, }

\makeatother

\newcommand{\beginsupplement}{%
    \renewcommand\thefigure{S\arabic{figure}}    
    \setcounter{figure}{0}
    \renewcommand\thetable{S\arabic{table}}    
    \setcounter{table}{0}  
    \renewcommand\theequation{S\arabic{equation}}    
    \setcounter{equation}{0}
    \renewcommand\theHequation{S\arabic{equation}}
}

\newtheorem{theorem}{Theorem}

%% file: sections/main_paper/abstract.tex
\begin{abstract}
The remarkable success of diffusion and flow-matching models has ignited a surge of works on adapting them at test time for controlled generation tasks. Examples range from image editing to restoration, compression and personalization. However, due to the iterative nature of the sampling process in those models, it is computationally impractical to use gradient-based optimization to directly control the image generated at the end of the process. As a result, existing methods typically resort to manipulating each timestep separately. Here we introduce \oursshort{} -- a zero-order (gradient-free) optimization framework that treats the entire flow process as a black box, enabling optimization through the whole sampling path without backpropagation through the model.
Our method is both highly efficient and allows users to monitor the intermediate optimization results and perform early stopping if desired. We prove a sufficient condition on \oursshort{}'s step-size, under which convergence to the global optimum is guaranteed. We further show how to empirically estimate this upper bound so as to choose an appropriate step-size.
We demonstrate how \oursshort{} can be used for image editing, showcasing two options: (\textit{i})~inversion (determining the initial noise that generates a given image), and (\textit{ii})~directly steering the edited image to be similar to the source image while conforming to a target text prompt. 
In both cases, \oursshort{} achieves state-of-the-art results while using roughly the same number of neural function evaluations (NFEs) as existing methods. Code and examples are available on the project's \href{https://orronai.github.io/FlowOpt/}{webpage}.
\end{abstract}

%% file: sections/main_paper/introduction.tex
\section{Introduction}
Diffusion and flow matching models have emerged as powerful generative frameworks, achieving state-of-the-art (SotA) results on image, video, and audio generation \citep{ho2020denoising,song2021denoising,rombach2022high,lipman2023flow,liu2023flow,albergo2023building}. However, as opposed to their generative adversarial network (GAN) predecessors, flow models generate samples through an iterative process that often involves dozens of neural function evaluations (NFEs). This makes it challenging to adapt them at inference time for solving controlled generation tasks. Indeed, while GANs naturally lend themselves to gradient-based optimization for directly minimizing losses on the generator's output \citep{menon2020pulse}, in flow models this approach is computationally impractical. As a result, methods that use pre-trained flow models for controlled generation typically intervene in each step of the sampling process separately, without employing any direct supervision on the final result. This strategy is used \eg for image restoration, image editing (using inversion techniques), and image compression \citep{kawar2022denoising,tumanyan2023plug,pan2023effective,qi2023fatezero,huberman2024edit,hong2024exact,cohen2024slicedit,garibi2024renoise,manor2024zero,elata2024zero,wang2025taming,martin2025pnpflow,deng2025fireflow,ohayon2025compressed,samuel2025lightningfast}.

Recently, \cite{ben2024d} demonstrated the great potential of employing optimization through the whole flow process in the context of solving inverse problems with pre-trained flow models. Unlike other methods, this approach directly controls the generated image, and thus avoids accumulation of approximation errors that can build up throughout the flow path. However, performing gradient-based optimization is not scalable to reasonably sized models and image dimensions. In fact, even with a small flow-matching model, small images ($128\times 128$), and memory-saving techniques like gradient checkpointing, this approach takes approximately $15$ minutes to run on a single input. 

In this work, we introduce \oursshort{} -- a zero-order (gradient-free) optimization framework for directly minimizing loss functions on the target image without backpropagating through the model.
Specifically, unrolling the sampling process, a flow model can be viewed as a chain of neural networks, which we refer to as ``denoisers''. Our approach treats this entire chain of denoisers as a black box, and enables optimization with respect to arbitrary loss functions. Here we specifically focus on image-editing objectives. The avoidance of backpropagation enables working with large flow models and treating large images.
Furthermore, it allows using a small number of flow timesteps, which is in contrast with inversion-based techniques that often require many timesteps to avoid error accumulation. Taken together, these features enable \oursshort{} to achieve SotA results at a number of NFEs comparable to existing methods. Additionally, \oursshort{} allows monitoring the intermediate optimization results. Thus, at the same budget of NFEs as existing methods, \oursshort{} in fact provides multiple candidate edited images (one per optimization step) from which the user can choose.

Zero-order optimization has been previously used in several computer vision contexts \citep{tao2017zero,milanfar2018rendition,chen2019zo,tu2019autozoom}. \oursshort{} is a generalization of the method of \citet{tao2017zero}, with the difference that the update in each optimization step is multiplied by a step-size $\eta$ (the method of \citet{tao2017zero} corresponds to \oursshort{} with $\eta=1$). As we show, this modification is of dramatic importance. Specifically, we prove a sufficient condition on~$\eta$ under which convergence to the global minimum is guaranteed, and show that for popular flow models this bound is orders of magnitude smaller than $1$. We demonstrate that \oursshort{} indeed converges when $\eta$ is chosen smaller than the bound, and fails to converge when it significantly exceeds the bound.

\begin{figure}[t]
    \centering
    \includegraphics[width=\linewidth]{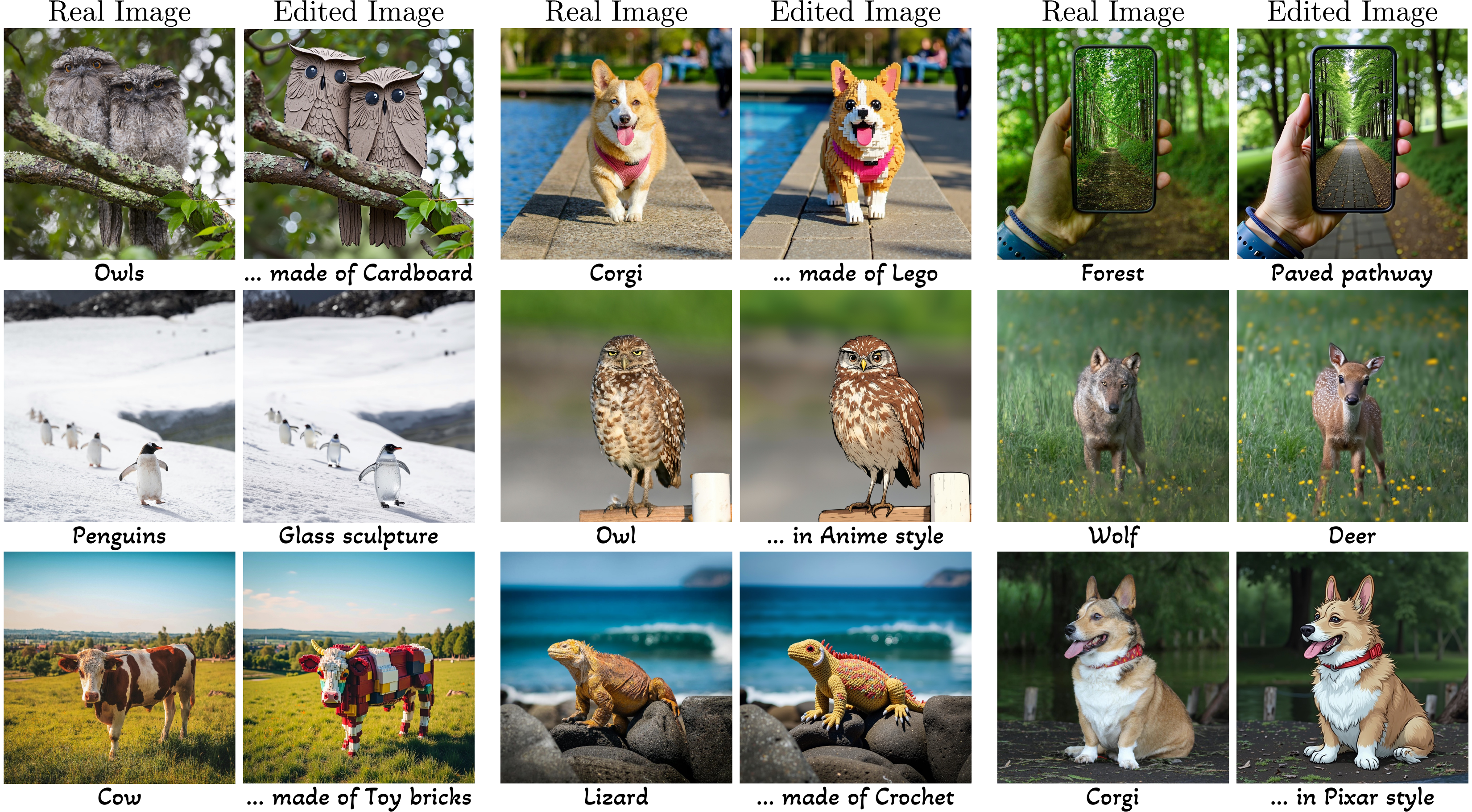}
    \caption{\textbf{\oursshort{}.} We propose a zero-order (gradient-free) framework for optimization through an unrolled flow sampling process. \oursshort{} can efficiently optimize losses on the target image, even when working with large models and high resolution images. We leverage our framework for text-based image editing, demonstrating state-of-the-art results on both FLUX (first and third rows) and Stable Diffusion 3 (second row). Fine details are visible upon zooming in.}
    \label{fig:teaser}
\end{figure}

We demonstrate the effectiveness of \oursshort{} for both image reconstruction (inversion) and direct image editing (Fig.~\ref{fig:teaser}), using the FLUX-1.dev \citep{flux2024} and Stable Diffusion 3 (SD3) \citep{esser2024scaling} text-to-image (T2I) models. We show that \oursshort{} provides an efficient solution to these tasks, delivering SotA performance at running times comparable to existing methods.

%% file: sections/main_paper/realted_work.tex
\section{Related Work}
\label{sec:related_work}
T2I diffusion and flow-based models \citep{saharia2022photorealistic,ramesh2022hierarchical} generate images by steering a diffusion or flow process according to a text prompt provided by the user.
Latent diffusion/flow variants \citep{rombach2022high,vahdat2021score,dao2023lfm} follow the same principle but operate in a lower-dimensional latent space, improving computational efficiency while preserving visual fidelity. Many methods utilize these T2I foundation models for downstream tasks like image editing in a zero-shot manner.

A common approach for performing image editing with pre-trained diffusion/flow models is to start with an inversion stage \citep{song2021denoising} (often referred to as DDIM or ODE inversion), whose goal is to extract the initial noise that would generate the input image if used in a regular sampling process. Once this initial noise is obtained, it is used for sampling a new image, but using a text prompt that describes the desired edit. However, inversion methods introduce approximation errors that accumulate across the flow timesteps, and lead to significant reconstruction inaccuracies \citep{mokady2023null,huberman2024edit}.

One line of work focuses on improving the precision of ODE-inversion. \citet{wang2025taming} employ a high-order Taylor expansion to more accurately approximate the nonlinear components of the flow. \citet{deng2025fireflow} propose a solver that reuses intermediate velocity vector approximations. Yet, despite improving numerical accuracy, such methods still operate on each timestep separately and do not promote direct alignment with the given image during the inversion. Therefore, they still suffer from accumulation of errors that can degrade overall performance.

A different approach is to optimize each denoising timestep independently \citep{mokady2023null,pan2023effective,hong2024exact,garibi2024renoise,miyake2025negative,samuel2025lightningfast}. For instance, \citet{mokady2023null} optimize the unconditional null prompt embedding used in classifier-free guidance (CFG) \citep{ho2021classifier} during the reverse process, aligning latent variables obtained through DDIM inversion. While effective, this approach requires storing all latent variables and optimized embeddings in memory, which becomes prohibitive for a large number of timesteps. Furthermore, repeated backward passes through each timestep render such methods impractical for interactive editing with large-scale models. \citet{hong2024exact} propose a gradient-based inversion scheme applied independently at each timestep, however their method is computationally expensive and time-intensive, particularly for modern large-scale T2I models. \citet{pan2023effective} and \citet{garibi2024renoise} mitigate this by introducing fixed-point iteration strategies that iteratively refine approximations of predicted states along the diffusion trajectory. However, all these methods rely on optimizing each timestep independently, ignoring the input image in each optimization step. This leads to accumulation of local approximation errors that degrade overall performance.

There exist several optimization-based methods that may superficially seem similar to \oursshort{}, as they neglect the Jacobian of the denoiser and thus avoid backpropagation through the model. These include Score Distillation Sampling (SDS) \citep{poole2023dreamfusion}, Delta Denoising Score (DDS) \citep{hertz2023delta}, Posterior Distillation Sampling (PDS) \citep{koo2024posterior}, and inverse Rectified Flow Distillation Sampling (iRFDS) \citep{yang2025texttoimage}. However, these methods still optimize each timestep separately by randomly sampling a timestep in each optimization step and performing an update based on that timestep alone. This is in contrast with \oursshort{}, which performs optimization through the whole chain of denoisers simultaneously.

Finally, \citet{ben2024d} proposed D-Flow, a method that like \oursshort{}, optimizes across the entire generative process. However, their framework relies on gradient-based optimization and requires repeated backpropagation through the entire chain of denoisers. This makes the method computationally intensive and impractical for high-resolution, real-world applications -- precisely the setting we aim to address with \oursshort{}.

%% file: sections/main_paper/preliminaries.tex
\section{Preliminaries and notation}
\label{sec:preliminaries}
Probability flow ODE \citep{song2021scorebased} and flow-matching models \citep{lipman2023flow,liu2023flow,albergo2023building} generate images by numerically solving an ODE over a time parameter $t$. Focusing for simplicity on the flow-matching formalism, the ODE takes the form 
\begin{equation}
    d\boldsymbol{z}_t = \boldsymbol{v}_t(\boldsymbol{z}_t, c)\,dt, \quad  t\in [0, 1]. 
\end{equation}
This ODE is designed such that when initialized at $t = 1$ with a sample from some source distribution (usually taken to be an isotropic Gaussian), $\boldsymbol{z}_1 \sim \pi_1$, and run backwards in time until $t = 0$, it yields a sample from a desired target distribution (\textit{e.g.}~the distribution of natural images), $\boldsymbol{z}_0 \sim \pi_0$. The function $\boldsymbol{v}_t(\cdot, \cdot)$ is a time dependent vector field that optionally accepts a condition $c$ (\eg a text prompt) in its second argument. In practice, this velocity field is implemented by a neural network, which we refer to as ``denoiser'', and the ODE is discretized and solved numerically as
\begin{equation}\label{eq:flow_discrete}
    \boldsymbol{z}_{t + \Delta t} = \boldsymbol{z}_t + \boldsymbol{v}_t(\boldsymbol{z}_t, c)\,\Delta t,
\end{equation}
where $\Delta t$ is the (negative) discretization step. 

Unrolling \eqref{eq:flow_discrete}, the sample $\boldsymbol{z}_0$ generated at the end of the flow process can be written as a function of the initial noise $\boldsymbol{z}_1$, namely $\boldsymbol{z}_0 = f(\boldsymbol{z}_1, c)$. This function is given by
\begin{equation}\label{eq:flow_black_box}
    f(\boldsymbol{z}_1, c) = \boldsymbol{z}_1 + \sum_{i}{ \boldsymbol{v}_{t_i}(\boldsymbol{z}_{t_i}, c) \, \Delta t},
\end{equation}
where $t_i=1+i\,\Delta t$ (see Fig.~\ref{fig:method}). For notational simplicity, we henceforth omit the condition $c$ whenever it is clear from the context. Furthermore, we sometimes use $f(\cdot)$ to denote the mapping from some intermediate timestep $t < 1$ to timestep $t=0$. Our method treats the function  $f(\cdot)$ as a black box in the sense that it can be evaluated but its Jacobian cannot be computed.

Commonly, the flow process is defined in the latent space of an encoder $\mathcal{E}(\cdot)$, so that the final image is obtained by passing the generated sample $\boldsymbol{z}_0$ through the corresponding decoder $\mathcal{D}(\cdot)$.

%% file: sections/main_paper/method.tex
\section{Method}
\label{sec:method}

\begin{figure}[t]
    \centering
    \includegraphics[width=\textwidth]{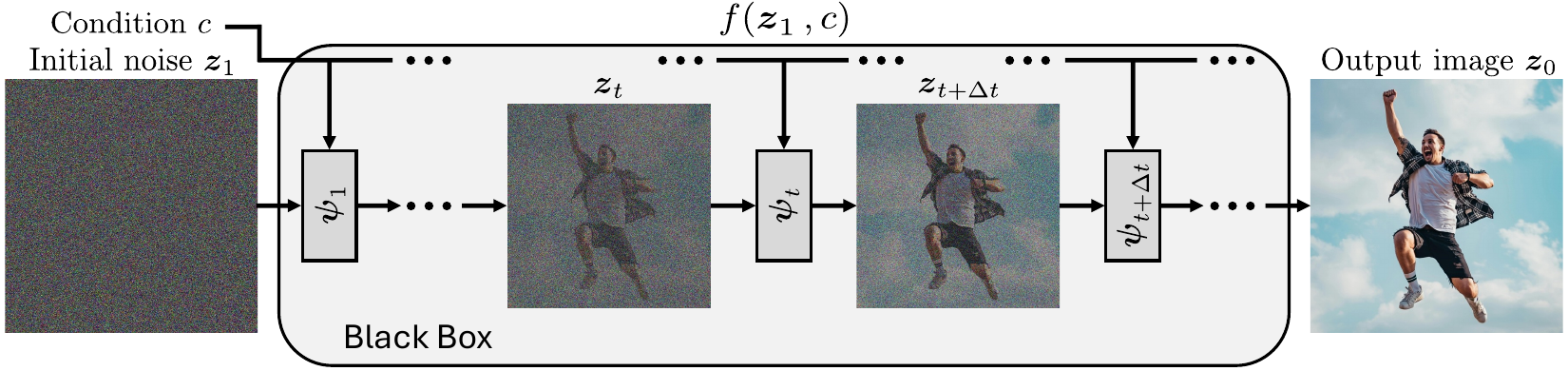}
    \caption{\textbf{A whole flow process as a black box.}
    We encapsulate the flow process as a black box function $f$, which receives an initial noise $\boldsymbol{z}_1$ and text conditioning $c$, and outputs a clean sample~$\boldsymbol{z}_0$. Each internal step within the black box is given by $\boldsymbol{\psi}_t(\boldsymbol{z}_t, c) = \boldsymbol{z}_t + \boldsymbol{v}_t(\boldsymbol{z}_t, c)\Delta t$, where $\boldsymbol{v}_t$ is the text-conditioned velocity predicting network.}
    \label{fig:method}
\end{figure}

Given a source image $\boldsymbol{y}$, a text prompt $c_{\text{src}}$ describing it, and a target text prompt $c_{\text{tar}}$ describing a desired edit, our goal is to generate an edited image $\boldsymbol{y}_{\text{edit}}$ that conforms to $c_{\text{tar}}$ while being as similar as possible to $\boldsymbol{y}$. Like previous approaches, we rely on a pre-trained flow model. However, in contrast to existing methods we propose to achieve this by directly optimizing over the vector $\boldsymbol{z}_t$ at some timestep $t$ (usually taken to be $1$), such that the image $\boldsymbol{z}_0$ at the end of the flow process is close to $\boldsymbol{y}$. 

Formalizing this mathematically, we are interested in $\boldsymbol{z}_t^*=\argmin_{\boldsymbol{z}_t}\mathcal{L}(f(\boldsymbol{z}_t,c), \boldsymbol{y})$, where $\mathcal{L}$ is some dissimilarity measure. Let us focus on the $L^2$ loss (see App.~\ref{ap:other_losses} for other losses). In this case, 
\begin{equation}
    \boldsymbol{z}_t^*=\argmin_{\boldsymbol{z}_t} {\frac{1}{2} \norm{f(\boldsymbol{z}_t, c) - \boldsymbol{y}}^2}.
\label{eq:mse_loss}
\end{equation}
This optimization problem can be used in two distinct ways. ($i$) \textbf{Inversion:} setting $c=c_{\text{src}}$ in \eqref{eq:mse_loss} leads to a $\boldsymbol{z}_t^*$ that reconstructs the input image with the source prompt. ($ii$) \textbf{Direct editing:} setting $c=c_{\text{tar}}$ in \eqref{eq:mse_loss} leads to a $\boldsymbol{z}_t^*$ that directly approximates the input image with the target prompt. In both cases, once $\boldsymbol{z}_t^*$ is obtained, it can be used to generate the edited image by performing sampling with the target prompt, $\boldsymbol{y}_{\text{edit}}=f(\boldsymbol{z}_t^*,c_{\text{tar}})$.

\begin{figure}[t]
    \centering
    \includegraphics[width=\textwidth]{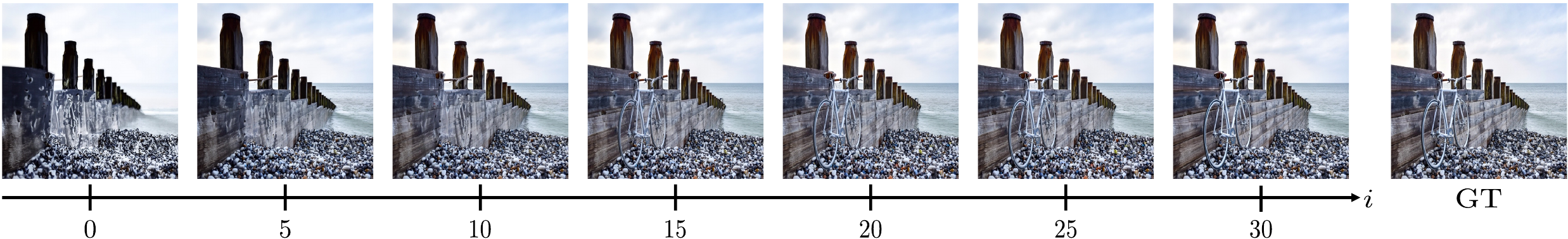}
    \caption{\textbf{Image inversion with \oursshort{}.} Intermediate samples $\boldsymbol{z}_0^{(i)} = f(\boldsymbol{z}^{(i)}_t, c)$ attained during our zero-order optimization through a chain of $10$ denoising steps (FLUX) for the task of reconstruction (inversion), \ie with $c=c_{\text{src}}$. Notice the missing details in the early steps, such as the bicycle and the horizon. As the iterations progress, the reconstruction converges to the ground truth image.}
    \label{fig:inversion_mse}
\end{figure}

\begin{figure}[t]
    \centering
    \includegraphics[width=\textwidth]{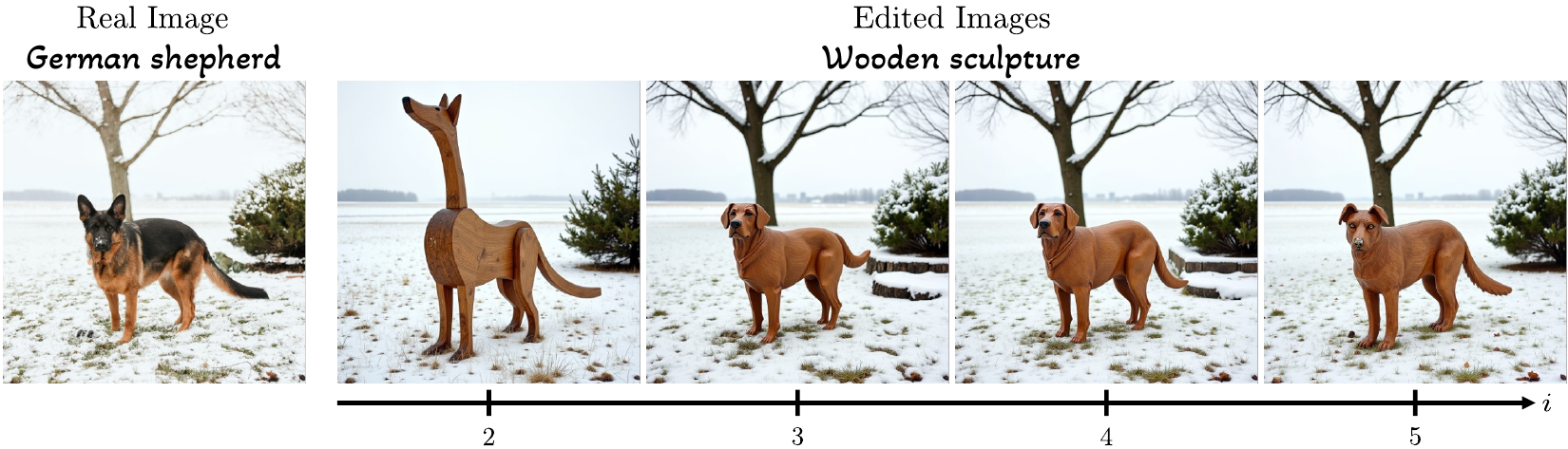}
    \caption{\textbf{Direct image editing with \oursshort{}.} Intermediate samples $\boldsymbol{z}_0^{(i)} = f(\boldsymbol{z}^{(i)}_t, c)$ attained during our zero-order optimization through a chain of $15$ denoising steps (FLUX) for direct image editing, \ie with $c=c_{\text{tar}}$. Notice the misalignment in the dog's body structure in the first iterations.}
    \label{fig:editing_iterations}
\end{figure}

Using gradient descent to solve \eqref{eq:mse_loss} would lead to the iterations
\begin{equation}
    \boldsymbol{z}^{(i+1)}_t \gets \boldsymbol{z}^{(i)}_t - \eta \,\boldsymbol{J}( \boldsymbol{z}^{(i)}_t)^{\top}  \left( f(\boldsymbol{z}^{(i)}_t) - \boldsymbol{y} \right),
\label{eq:mse_loss_update_rule_jacobian}
\end{equation}
where $\eta$ is the step size and $\boldsymbol{J} ( \boldsymbol{z}^{(i)}_t )$ is the Jacobian of $f(\cdot)$ with respect to $\boldsymbol{z}^{(i)}_t$. However, as mentioned above, backpropagation through whole flow processes is computationally impractical. Therefore, as an alternative, here we propose to simply ignore the Jacobian. This leads to the zero-order (gradient-free) iterations
\begin{equation}
    \boldsymbol{z}^{(i+1)}_t \gets \boldsymbol{z}^{(i)}_t - \eta \left( f(\boldsymbol{z}^{(i)}_t) - \boldsymbol{y} \right).
\label{eq:mse_update_rule}
\end{equation}
Figure \ref{fig:inversion_mse} demonstrates the progression of those iterates when used for inversion (with the source prompt).
Figure \ref{fig:editing_iterations} demonstrates the progression of the iterates when used for direct editing (with the target prompt). 
Algorithm \ref{algorithm:ZERO} summarizes the proposed method.

\begin{algorithm2e*}[t]
\SetKwInput{KwInput}{Inputs}
\DontPrintSemicolon
\textbf{Require:} step size $\eta$, number of iterations $N$, condition $c$, input image $\boldsymbol{y}$ \\
\textbf{Initialization:} $\boldsymbol{z}^{(0)}_t \in \mathbb{R}^d$ \\
\For{$i \gets 0, \hdots ,N - 1$}{
    $\boldsymbol{z}^{(i)}_0 = f(\boldsymbol{z}^{(i)}_t, c)$

    $\boldsymbol{z}^{(i+1)}_t \gets \boldsymbol{z}^{(i)}_t - \eta ( \boldsymbol{z}^{(i)}_0 - \boldsymbol{y})$
}
$\boldsymbol{z}^{(N)}_0 = f(\boldsymbol{z}^{(N)}_t, c)$

Return $\{ \boldsymbol{z}^{(i)}_0 \}_{i=0}^N$
\caption{\ourslong{} (\oursshort{})}
\label{algorithm:ZERO}
\end{algorithm2e*}

Before providing a theoretical convergence guarantee, two comments are in place. 
First, when $\eta = 1$, \eqref{eq:mse_update_rule} degenerates to the method of \citet{tao2017zero}. However, as we show below, $\eta$ is of crucial importance, as the maximal step size allowing convergence is much smaller than~$1$ for modern flow-matching models. Second, it is insightful to note that for flow-matching models, \eqref{eq:mse_update_rule} is equivalent to using gradient descent with step-size $\eta$ while applying the $\texttt{stop-grad}$ operator on the output of the velocity prediction network. Similarly, for probability flow ODE models \citep{song2021scorebased}, (a.k.a.~DDIM \citep{song2021denoising}), \eqref{eq:mse_update_rule} is equivalent to using gradient descent with step size $\sqrt{\alpha_T} \eta$ while applying $\texttt{stop-grad}$ on the noise prediction network (following the notation of \citet{song2021denoising}). The derivations of those observations are provided in App.~\ref{ap:pf_coeffs}.

The iterations of \eqref{eq:mse_update_rule} can be written as $\boldsymbol{z}_t^{(i + 1)} = g(\boldsymbol{z}_t^{(i)})$, where $g(\boldsymbol{u}) \triangleq \boldsymbol{u} - \eta ( f(\boldsymbol{u}) - \boldsymbol{y})$. By the Banach fixed-point theorem, if $g(\cdot)$ is a contractive mapping\footnote{$g(\cdot)$ is a contractive mapping if it satisfies $\|g(\boldsymbol{u}_1)-g(\boldsymbol{u}_2)\|\leq \gamma \|\boldsymbol{u}_1-\boldsymbol{u}_2\|$ for some $\gamma <1$ and all $\boldsymbol{u}_1,\boldsymbol{u}_2$.} then there exists a unique point satisfying $\boldsymbol{z}_t^* = g(\boldsymbol{z}_t^*)$, and thus $f(\boldsymbol{z}_t^*)=\boldsymbol{y}$. Furthermore, in this case the iterations converge to this unique solution. This fact can be used to obtain a sufficient condition on the step size $\eta$ under which the iterations are guaranteed to converge to the global minimum (see proof in App.~\ref{ap:contraction_mapping}).
\begin{theorem}
    \label{theorem:contraction_mapping}
    Assume that $\inf_{\boldsymbol{u}_1 \neq \boldsymbol{u}_2}\frac{\left\langle \boldsymbol{u}_1 - \boldsymbol{u}_2, f(\boldsymbol{u}_1) - f(\boldsymbol{u}_2) \right\rangle}{\norm{\boldsymbol{u}_1 - \boldsymbol{u}_2}\norm{f(\boldsymbol{u}_1) - f(\boldsymbol{u}_2)}} > 0$.
    If the step size $\eta$ satisfies
        \begin{equation}
           0 < \eta < 2  \inf_{\boldsymbol{u}_1, \boldsymbol{u}_2}\frac{\left\langle \boldsymbol{u}_1 - \boldsymbol{u}_2, f(\boldsymbol{u}_1) - f(\boldsymbol{u}_2) \right\rangle}{\norm{f(\boldsymbol{u}_1) - f(\boldsymbol{u}_2)}^2}
        \label{eq:contraction_mapping_sufficient}
        \end{equation}
        then there is a unique $\boldsymbol{z}_t^*$ satisfying $f(\boldsymbol{z}_t^*) = \boldsymbol{y}$ and the iterations of \eqref{eq:mse_update_rule} converge to this $\boldsymbol{z}_t^*$.
\end{theorem}

\begin{wrapfigure}{r}{0.5\linewidth}
  \centering
  \includegraphics[width=\linewidth]{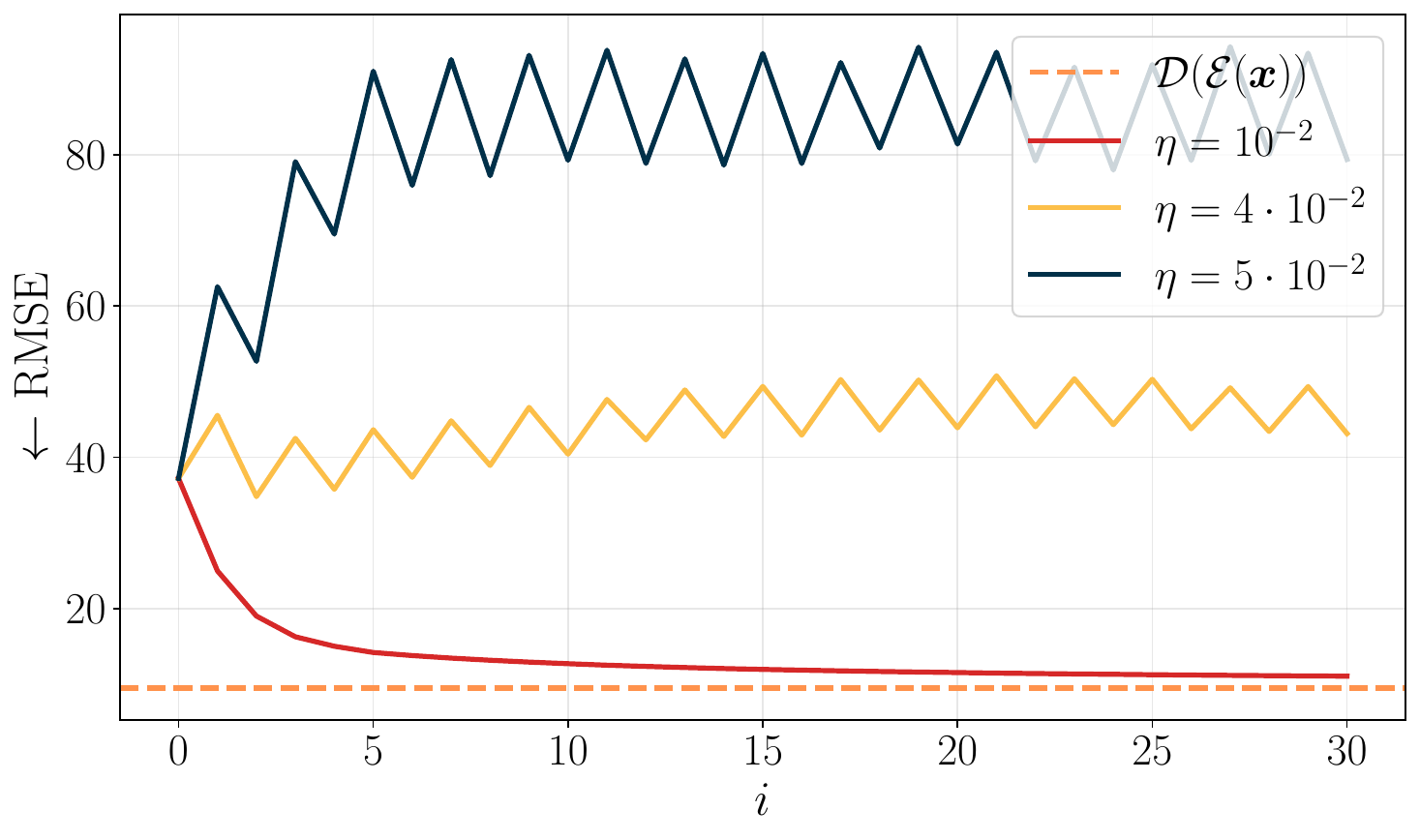}
    \caption{\textbf{Convergence analysis}. The plot shows RMSE in pixel space vs.~number of iterations for the task of inversion, averaged over a dataset. The step size we use (red) satisfies the sufficient condition of \eqref{eq:contraction_mapping_sufficient} and thus leads to convergence. Step sizes that are $ 4\times$ and $5\times$ larger (yellow and black) do not satisfy the condition and do not lead to convergence. The dashed orange line is the minimal RMSE achievable in this setting. It corresponds to passing images through the encoder and decoder.}
  \label{fig:eta_comparisons_sd3}
\end{wrapfigure}

The bound in \eqref{eq:contraction_mapping_sufficient} depends only on the flow model $f(\cdot)$. It can thus be computed once for each model in order to choose the step size. In App.~\ref{ap:contraction_mapping} we approximate this upper bound for the FLUX and SD3 models by drawing many pairs of samples $\boldsymbol{u}_1, \boldsymbol{u}_2$. As we show, the right-hand side of \eqref{eq:contraction_mapping_sufficient} is smallest when $\|\boldsymbol{u}_1 - \boldsymbol{u}_2\|$ is small. Table \ref{tab:contraction_mapping_upper_bounds} shows the bounds estimated for the two models, and the step sizes we chose for our experiments.

As can be seen, the bounds in Tab.~\ref{tab:contraction_mapping_upper_bounds} are significantly smaller than $1$, suggesting that the method of \citet{tao2017zero} is inapplicable in our setting. Indeed, Fig.~\ref{fig:eta_comparisons_sd3} shows the reconstruction error along the iterations for several choices of $\eta$ when used for inversion with SD3 (results for FLUX are presented in App.~\ref{ap:contraction_mapping}). When setting $\eta = 10^{-2}$, which is below the bound of $1.67\cdot10^{-2}$, the iterations converge. However, when using larger step sizes, like $4 \cdot 10^{-2}$ or $ 5 \cdot 10^{-2}$, the iterations fail to converge. The setting of this experiment is as in Sec.~\ref{subsec:reconstruction}.

\begin{table*}[t]
    \centering
    \caption{\textbf{Step sizes guaranteeing convergence}. Column 2 shows the estimated sufficient condition of \eqref{eq:contraction_mapping_sufficient} and column 3 reports the step size we chose for each model (see App.~\ref{ap:contraction_mapping} for details).}
    \label{tab:contraction_mapping_upper_bounds}
    \begin{tabular}{@{}ccc@{}}
    \toprule
     Model & Sufficient condition (\eqref{eq:contraction_mapping_sufficient})  & Our chosen step size  \\ \midrule
     FLUX & $\eta<2.70 \cdot 10^{-3}$ & $\eta=2.5 \cdot 10^{-3}$  \\
     SD3 & $\eta<1.67 \cdot 10^{-2}$ & $\eta=1.0\cdot10^{-2}$  \\
    \bottomrule
    \end{tabular}
\end{table*}

%% file: sections/main_paper/results.tex
\section{Experiments}
\label{sec:experiments}
We compare \oursshort{} against competing methods on two tasks: image reconstruction (inversion) and text-based image editing.
We show results with FLUX-1.dev in the main text and with SD3 in App.~\ref{ap:sd3}. We use the step sizes reported in Tab.~\ref{tab:contraction_mapping_upper_bounds} and initialize our algorithm with the UniInv \citep{jiao2025uniedit} inversion method (see App.~\ref{ap:initialization} for details).
All images are of dimension $1024 \times 1024$.

\subsection{Image Reconstruction (Inversion)}
\label{subsec:reconstruction}

\begin{figure}[t]
\centering
\begin{subfigure}[t]{\linewidth}
  \centering 
  \includegraphics[width=.49\textwidth]{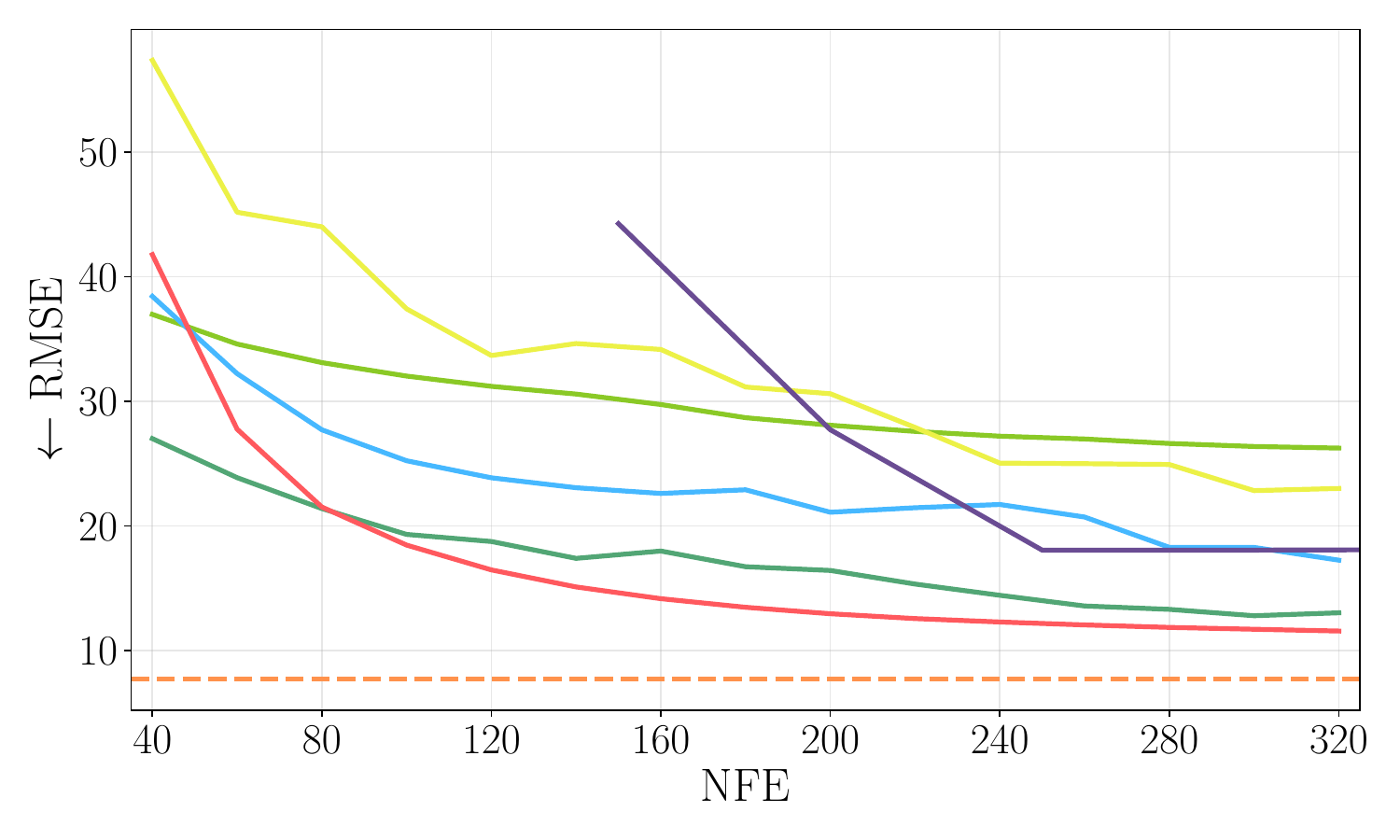}
  \includegraphics[width=.49\textwidth]{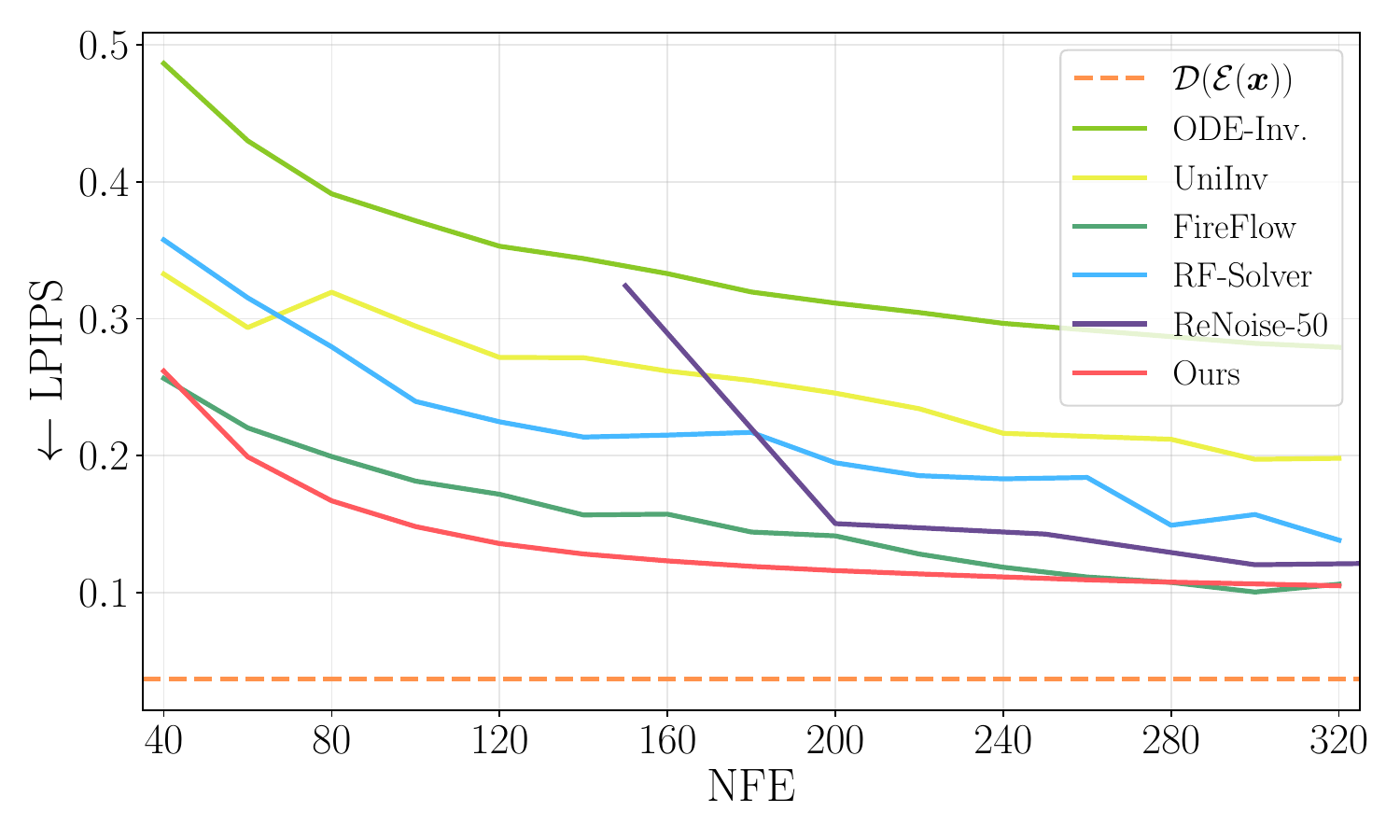}
\end{subfigure}
\begin{subfigure}[t]{\linewidth}
  \centering 
  \includegraphics[width=.49\textwidth]{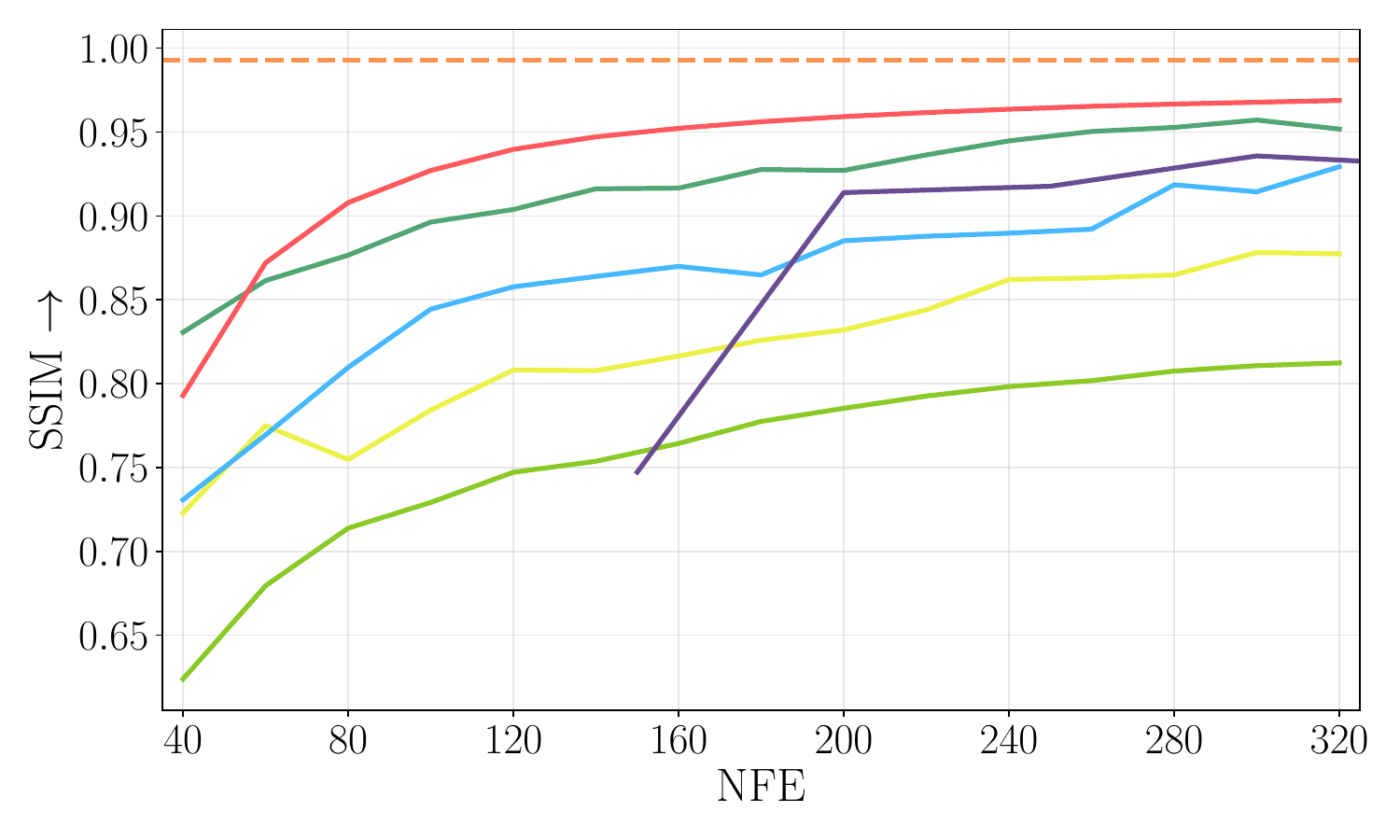}
  \includegraphics[width=.49\textwidth]{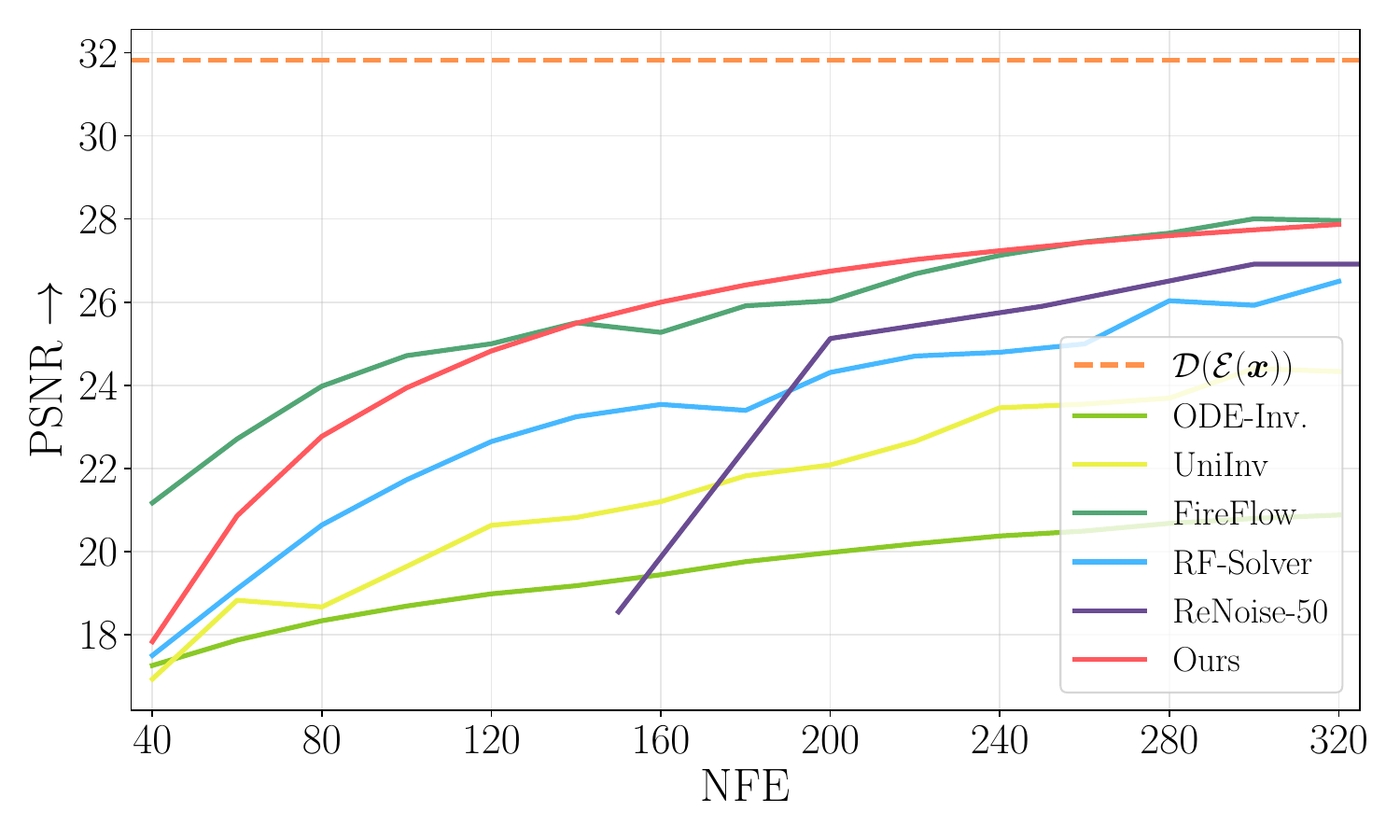}
\end{subfigure}
\caption{\textbf{Reconstruction accuracy vs.~NFEs for inversion}. The plots depict pixel-space RMSE, LPIPS, SSIM, and PSNR as a function of the number of NFEs for several inversion methods. The dashed bound corresponds to passing the images through the encoder and decoder. \oursshort{} achieves favorable reconstruction quality under 240 NFEs, which is the regime of practical interest.}
\label{fig:reconstruction_flux}
\end{figure}

For inversion, we use $c = c_{\text{src}}$ in \eqref{eq:mse_loss}, setting it to a text prompt describing the source image.
We set the number of flow steps in FLUX (number of denoisers) to $T = 10$ and evaluate the reconstruction error for various numbers of NFEs by varying the number of \oursshort{} iterations $N$. Specifically, we have $\text{NFE} = T (N + 2)$, as $T$ NFEs are used for the initialization, $N T$ NFEs for the optimization process, and $T$ NFEs for the final sampling process.

We randomly choose $100$ real images from the DIV2K dataset \citep{agustsson2017ntire}, and resize and center-crop them to dimension $1024 \times 1024$. For the source prompts, we caption each image with BLIP \citep{li2022blip} and then manually refine the prompt.

We compare \oursshort{} to several inversion methods: naive ODE Inversion, RF-Solver \citep{wang2025taming}, FireFlow \citep{deng2025fireflow}, UniInv \citep{jiao2025uniedit}, and ReNoise \citep{garibi2024renoise}. We use the official implementations of all methods except for ODE Inversion and ReNoise (that lacks an implementation for flow models), which we implemented by ourselves. 
To ensure a fair comparison, we set the number of timesteps for each method such that the total NFE count is the same for all methods. Specifically, for FireFlow and UniInv, which use a single forward pass per timestep, we set \smash{$T = \tfrac{\text{NFE}}{2}$}. For RF-Solver, which uses two forward passes per timestep for inversion and two for sampling, we set \smash{$T = \tfrac{\text{NFE}}{4}$}. 
For ReNoise, we used $T = 50$ and set the number of ReNoise steps so as to achieve the desired NFE count. We note that we evaluated ReNoise with various hyperparameter settings and chose the one that achieved the best results.

Figure \ref{fig:reconstruction_flux} shows the reconstruction accuracy achieved by all methods as a function of the NFEs. The figure reports pixel-space RMSE, PSNR, SSIM \citep{wang2004image}, and LPIPS \citep{zhang2018unreasonable}. As can be seen, \oursshort{} achieves the best reconstruction results over a wide range of NFE counts. In App.~\ref{ap:comparisons} we show that the same trend is obtained with empty text prompts, both with the CFG parameter of FLUX set to $0$ and with it set to $1$ (these options differ as FLUX is a distilled model).

\subsection{Image Editing}
\label{subsec:editing}

\begin{figure}[t]
\centering
\begin{subfigure}[t]{\linewidth}
  \centering 
  \includegraphics[width=.327\textwidth]{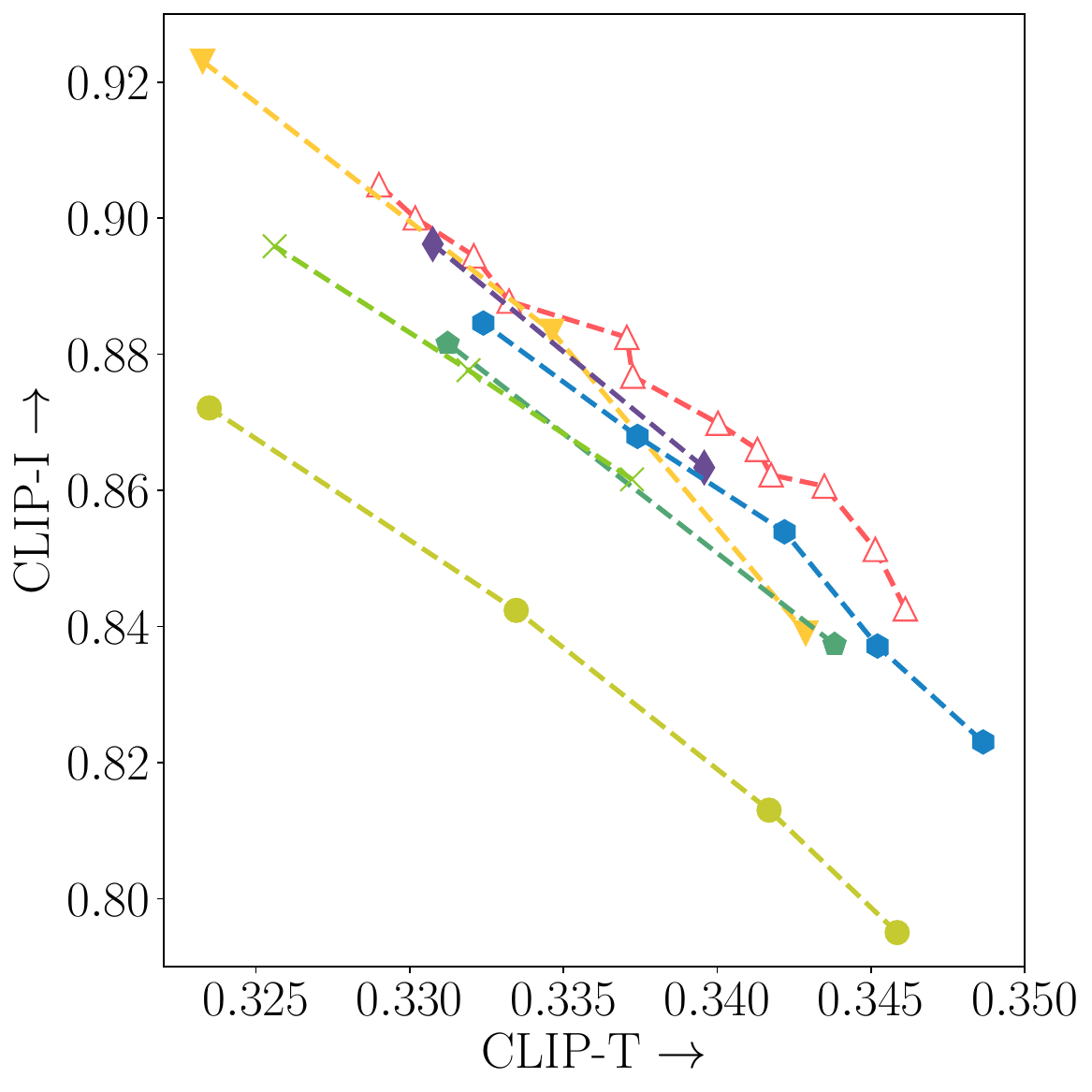}
  \includegraphics[width=.327\textwidth]{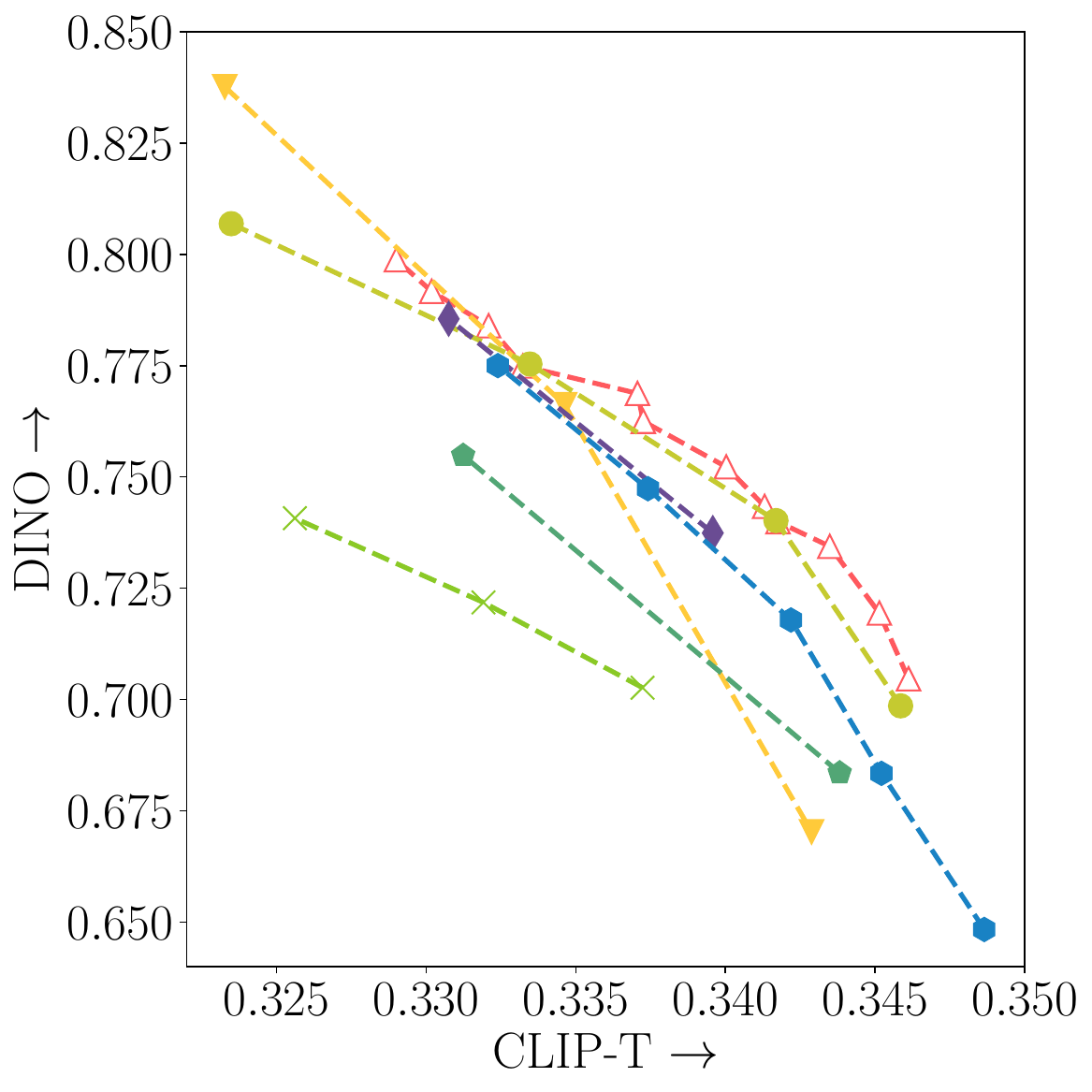}
  \includegraphics[width=.327\textwidth]{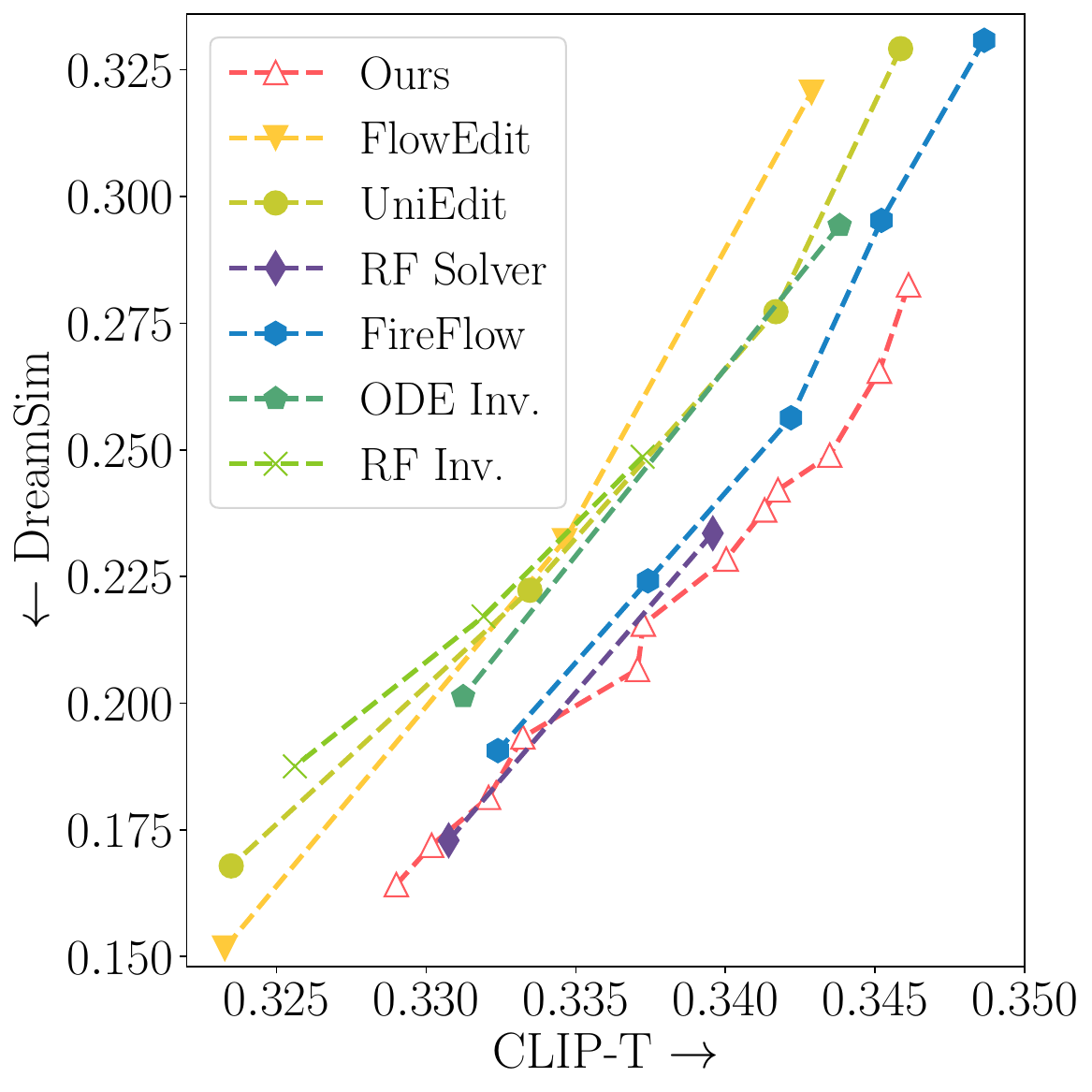}
\end{subfigure}
\caption{\textbf{Editing quantitative comparisons}. Semantic preservation of different editing methods evaluated using CLIP-Image, DINOv3 and DreamSim as functions of text adherence, measured by CLIP-Text. Connected markers represent different set of hyperparameters (see App.~\ref{ap:comparisons}). Our method achieves the most favorable balance between semantic preservation and text adherence.}
\label{fig:editing_quantitative_flux}
\end{figure}

\begin{figure}[t!]
    \centering
    \includegraphics[width=\linewidth]{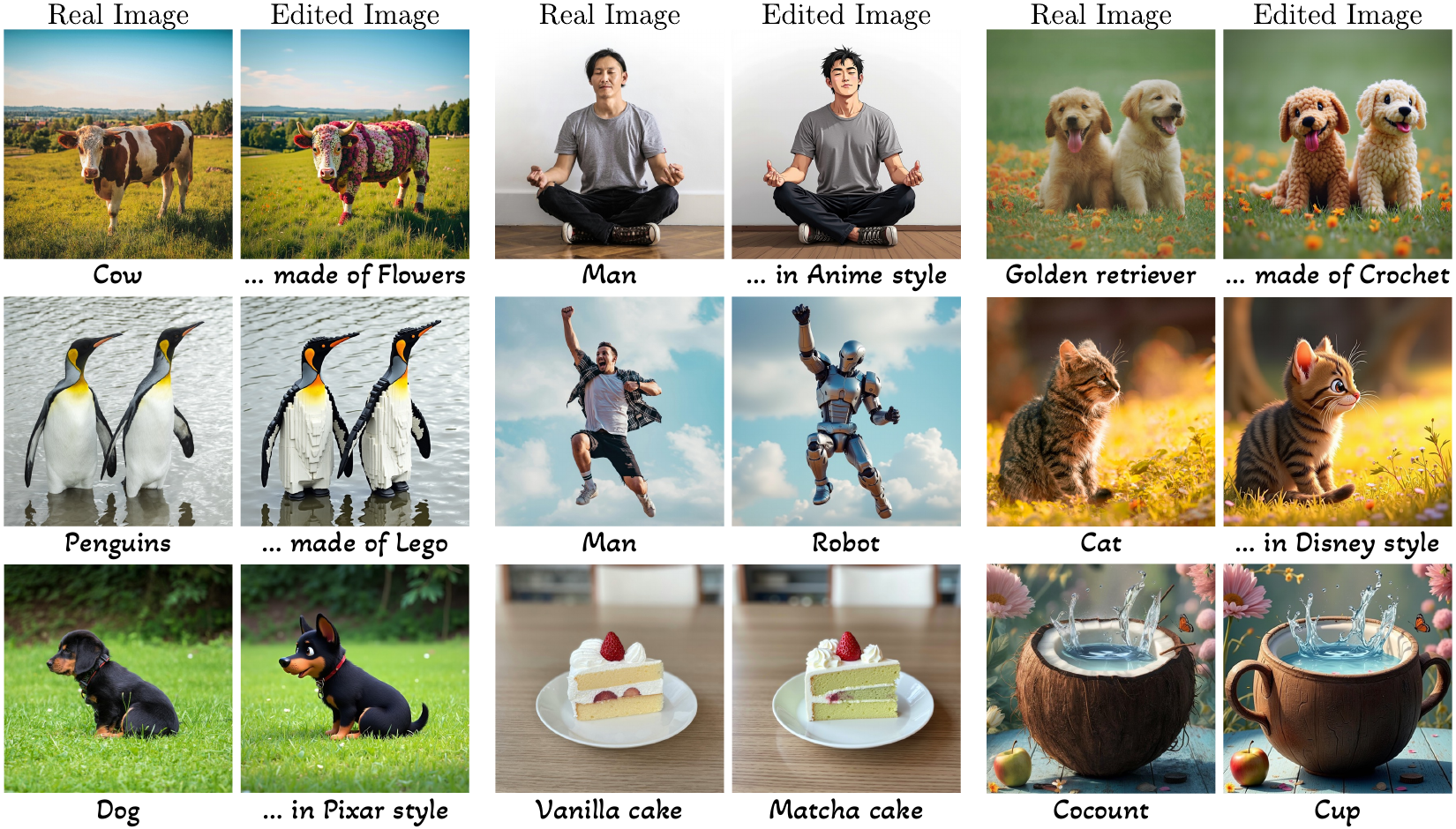}
    \caption{\textbf{\oursshort{} editing results}. Our method successfully preserves the object's semantics and structure, as well as the background details, all the while loyally adhering to the target text prompt. Fine details are visible upon zooming in.}
    \label{fig:main_results}
\end{figure}

Accurate inversion does not necessarily lead to good editing results. Indeed, even for synthetic images, for which the initial noise map is known, plain editing-by-inversion leads to unsatisfactory results \citep{kulikov2024flowedit,huberman2024edit} (see App.~\ref{ap:editing_by_inversion} for further discussion).  Accordingly, for the task of editing we employ our direct optimization approach, where the target text prompt $c = c_{\text{tar}}$ is used in \eqref{eq:mse_loss}.
In this case, we do not necessarily want a large number of iterations, to avoid getting too close to the original image. We therefore use $N \in \{ 2, 3, 4, 5 \}$.  We set the number of flow steps to $T = 15$ and perform the optimization on the latent vector at timestep $n_{\max} \in \{ 14, 13, 12 \}$ (corresponding to $t$ in \eqref{eq:mse_loss}). The total number of NFEs is given by $\text{NFE} = n_{\max} (N + 2)$. We use the default CFG of $3.5$.
All visual results in the paper were obtained with $n_{\max} = 13$, except for Fig.~\ref{fig:teaser}, whose hyperparameters are provided in App.~\ref{ap:teaser_hyperparameters}.

We evaluate all methods on the dataset of \cite{kulikov2024flowedit}, which we enriched with additional images and editing prompts. In total, our dataset consists of $90$ real images of dimensions $1024 \times 1024$ from the DIV2K dataset and from royalty free online sources \citep{pexels,pxhere}.
Each image was captioned by LLaVA-1.5 \citep{liu2024improved} and manually refined.
For each image, we manually created target editing prompts.
Overall, this led to about $400$ text-image pairs.

We compare our method against all aforementioned methods, in addition to FlowEdit \citep{kulikov2024flowedit} and RF-Inversion \citep{rout2025semantic}. These two methods were excluded from the inversion experiments of Sec.~\ref{subsec:reconstruction} as they do not use inversion in the regular sense (FlowEdit is inversion-free and RF-Inversion explicitly incorporates the source image into the denoising process). 
For ODE Inversion, we apply the same number of NFEs as our method. For other methods, we use the hyperparameters reported in the papers or in the official implementations. We performed a hyperparameter search for all methods that provided more than a single set of hyperparameters. Additional details and the final hyperparameters chosen for each method are provided in App.~\ref{ap:comparisons}.

\begin{figure}[t]
    \centering
    \includegraphics[width=\linewidth]{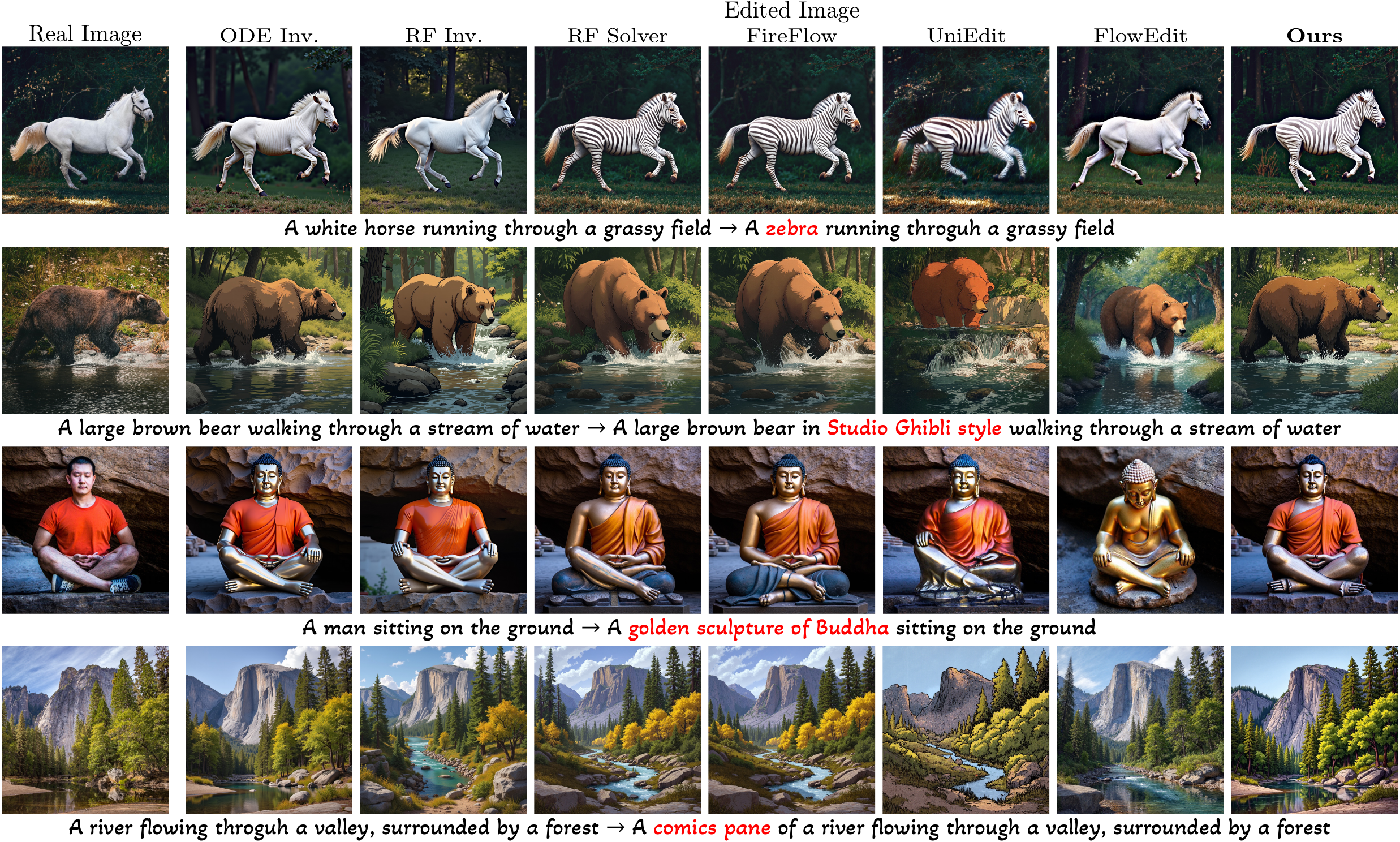}
    \caption{\textbf{Qualitative comparisons}. \oursshort{} is the only method to consistently adhere both to target text prompt, and to the original image. Fine details are visible upon zooming in. For instance, the back legs of the zebra in the first row, the posture of the bear in the second  row, the statue's limbs in the third row, and the structure of the scene in the last row.}
    \label{fig:main_qualitative_comparison_flux}
\end{figure}

Figures \ref{fig:teaser}, \ref{fig:main_results} and~\ref{fig:supp_results_flux} showcase the diverse editing capabilities of our method, including object replacement, style changes, and texture editing. \oursshort{} achieves high quality, text adherent edits that also remain loyal to the source image semantics. 
Figure \ref{fig:main_qualitative_comparison_flux} presents qualitative comparisons between \oursshort{} and other methods. As can be observed, our edits maintain superior alignment with the source image's structure while simultaneously adhering to the target text.
For example, when turning the horse into a zebra (first row), \oursshort{} successfully preserves the leg positions. Similarly, when replacing the sitting man (third row) with a golden sculpture of Buddha, \oursshort{} preserves the original limb orientations and scene background. For additional comparisons, see App.~\ref{ap:comparisons}.

Figure \ref{fig:editing_quantitative_flux} presents a numerical evaluation of the results obtained for various hyperparameters. We use cosine similarity on CLIP image and text embeddings \citep{radford2021learning} to measure adherence to the original image and to the target text prompt, respectively. For image adherence, we also use cosine similarity between DINOv3 embeddings \citep{caron2021emerging,simeoni2025dinov3}, as well as DreamSim \citep{fu2023dreamsim}.
As can be seen, our method achieves the best tradeoff between text adherence and structure preservation.

%% file: sections/main_paper/conclusions.tex
\section{Conclusions}
We presented a zero-order (gradient-free) framework that allows efficient optimization over the initial noise in a flow process while minimizing a loss over the sample generated at the end of the process. We demonstrated the effectiveness of our approach for performing image editing using pre-trained flow models. In particular, extensive comparisons showed that our \oursshort{} method achieves SotA performance on both image reconstruction and editing. We note that, similarly to other training-free editing methods, our approach still encounters difficulties in certain settings, like modifying large regions of the image (see App.~\ref{ap:limitations}). However, taking a broader perspective, we believe that our zero-order framework opens the door for exploiting pre-trained flow-models in diverse applications (\eg restoration, compression, and personalization) and for diverse modalities (\eg image, video, and audio). We leave those extensions for future work.

%% file: sections/main_paper/ethics_statement.tex
\subsubsection*{Ethics statement}
This work builds upon pre-trained generative models, and thus inherits the broader ethical considerations associated with their use. Such models may reflect or amplify societal biases present in the training data, and their outputs could be misinterpreted or misused in sensitive applications. In addition, our approach involves large-scale flow matching models, which carry the potential risk of being repurposed for harmful or malicious purposes. We emphasize that our contributions are intended solely for advancing research in generative modeling.

%% file: sections/main_paper/acknowledgements.tex
\subsubsection*{Acknowledgments}
This research was supported by the Israel Science Foundation (grant no.~2318/22) and by the Ollendorff Minerva Center, ECE faculty, Technion. The authors thank Matan Kleiner for his insightful suggestions throughout this work.

%% file: sections/supplementary/additional_results.tex
\section{Additional results}
\label{ap:additional_results}
\begin{figure}[htbp]
    \centering
    \includegraphics[width=\linewidth]{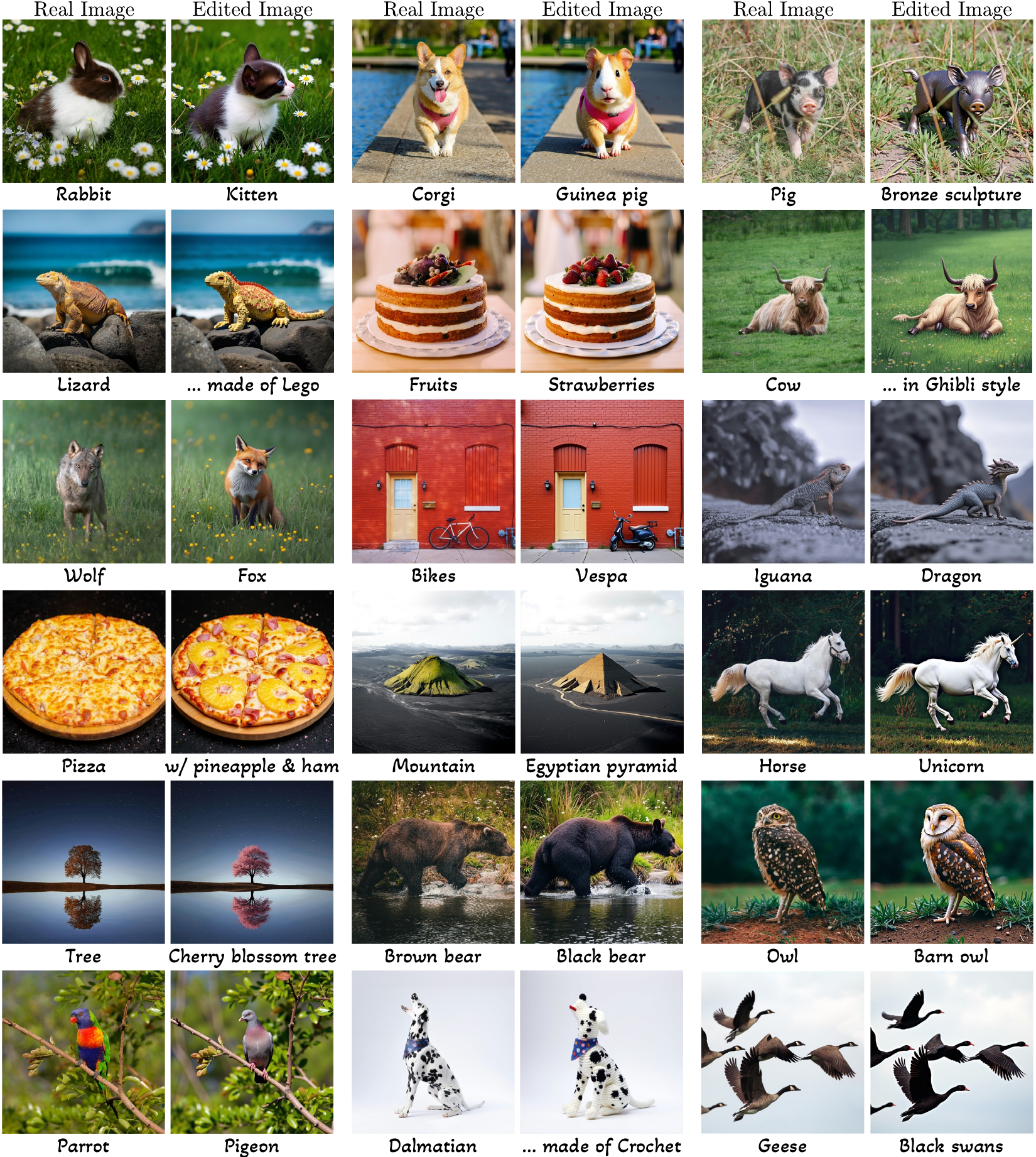}
    \caption{\textbf{Additional \oursshort{} results (FLUX).}}
    \label{fig:supp_results_flux}
\end{figure}

Additional results obtained by our method for image editing with FLUX are presented in Fig.~\ref{fig:supp_results_flux}.

\clearpage

%% file: sections/supplementary/comparisons.tex
\section{Comparisons}
\label{ap:comparisons}
\subsection{Image reconstruction (inversion)}
\begin{figure}[htbp]
\centering
\begin{subfigure}[t]{\linewidth}
  \centering 
  \includegraphics[width=.49\textwidth]{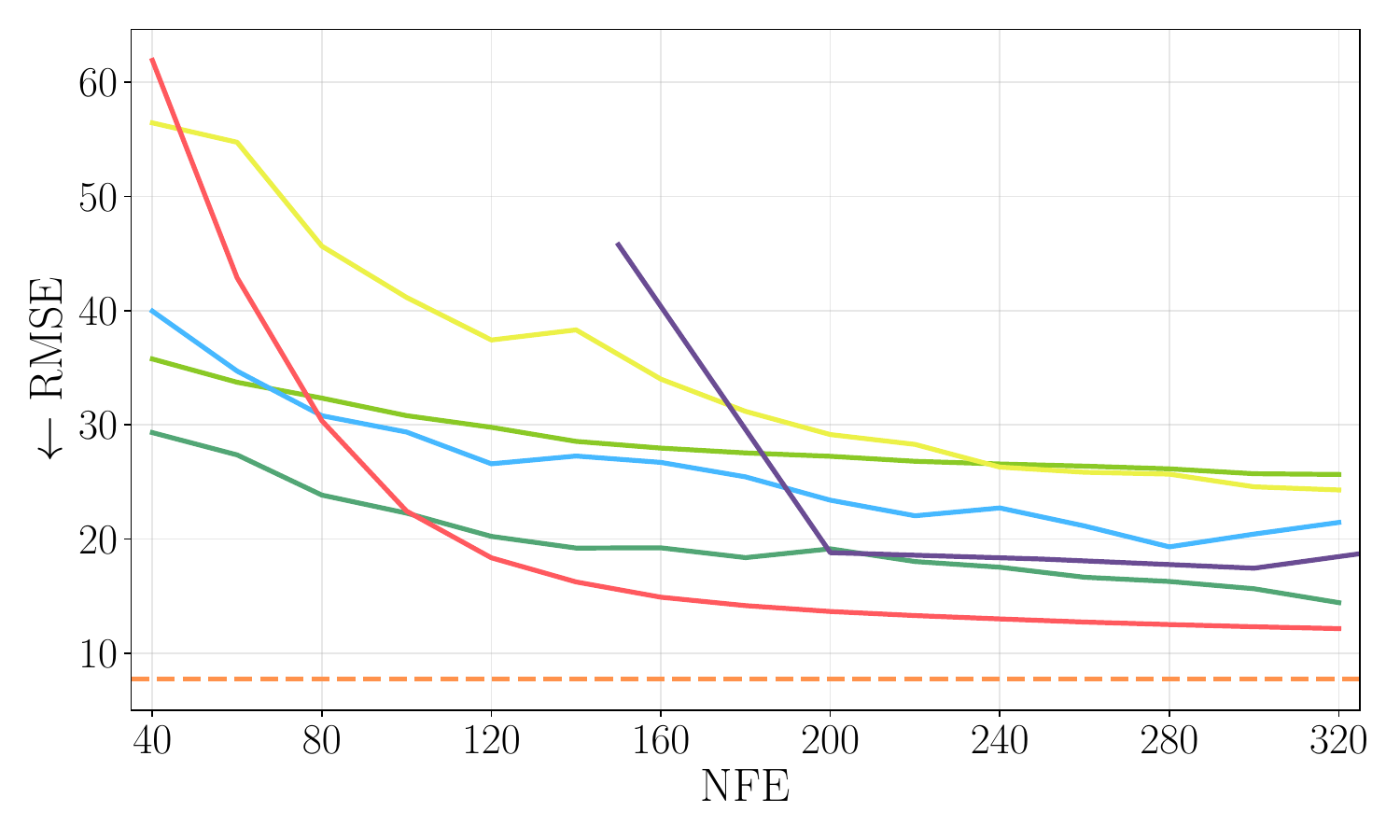}
  \includegraphics[width=.49\textwidth]{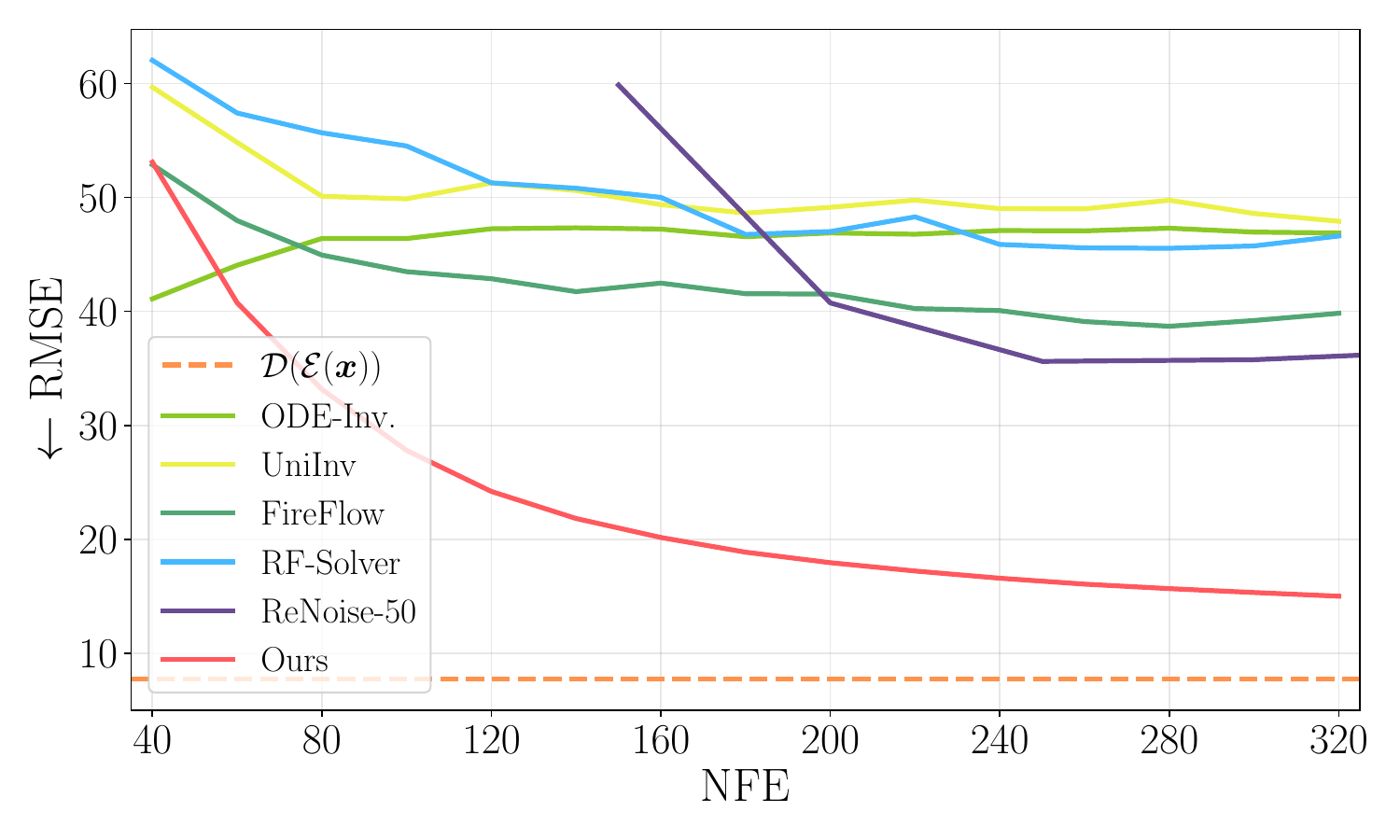}
\end{subfigure}
\hfill
\begin{subfigure}[t]{\linewidth}
  \centering 
  \includegraphics[width=.49\textwidth]{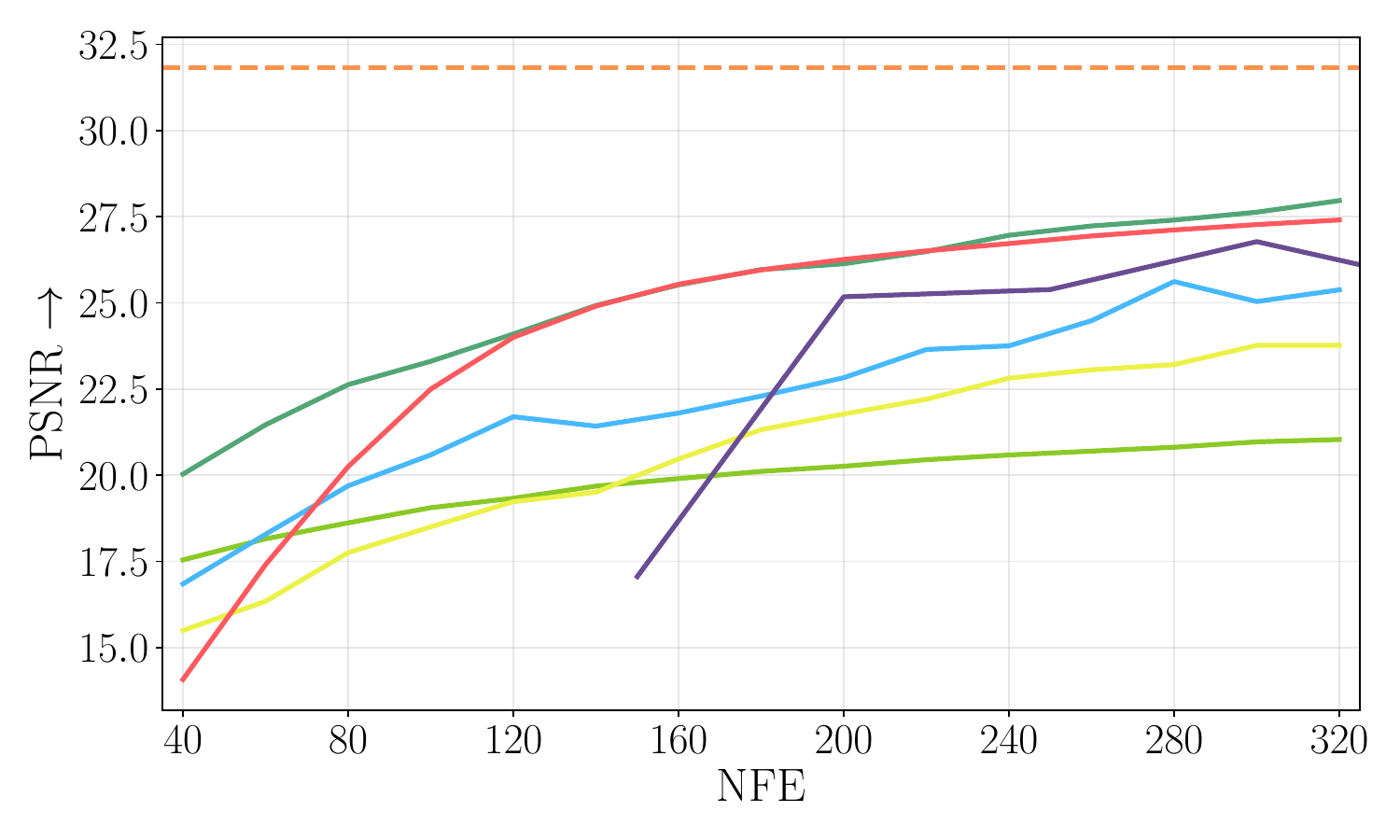}
  \includegraphics[width=.49\textwidth]{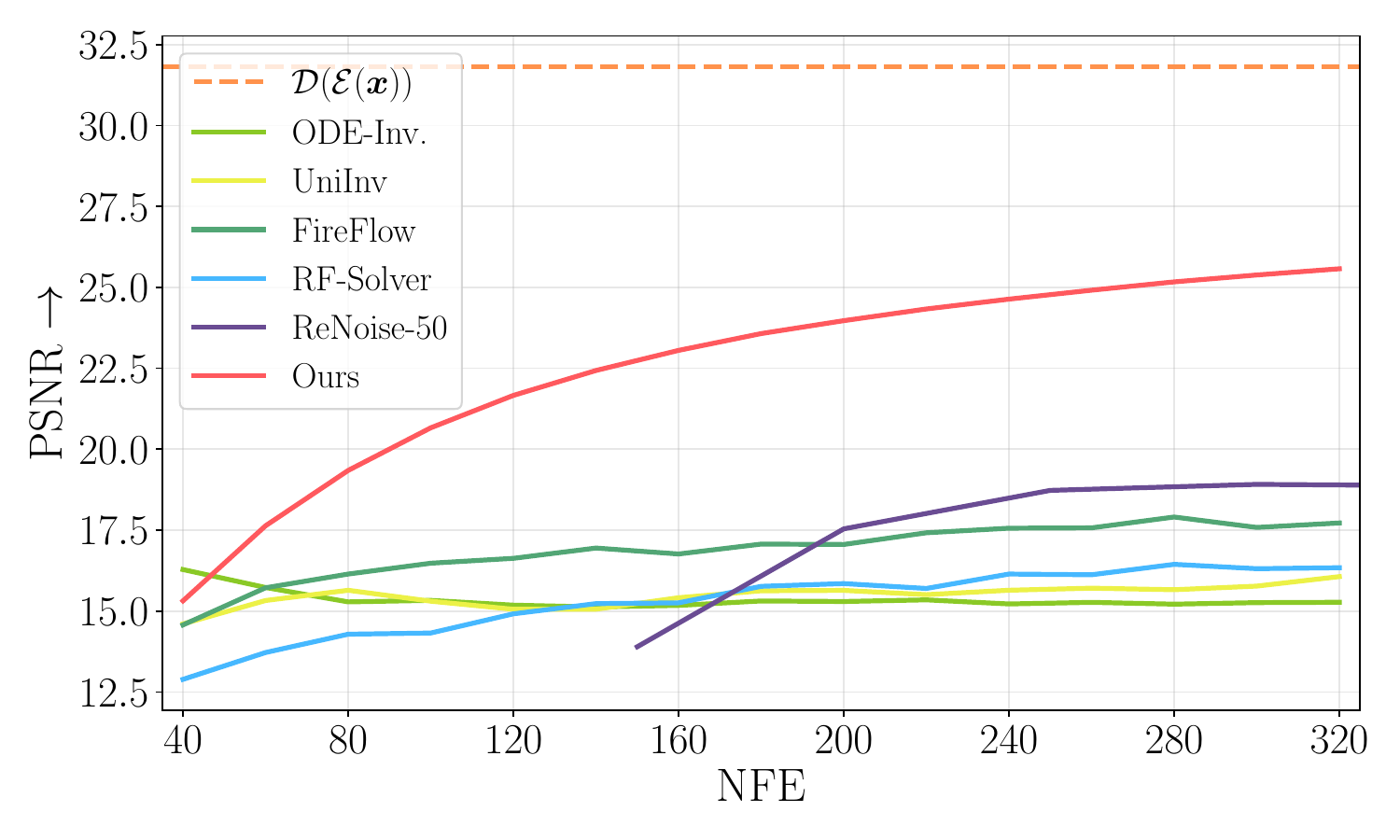}
\end{subfigure}
\hfill
\begin{subfigure}[t]{\linewidth}
  \centering 
  \includegraphics[width=.49\textwidth]{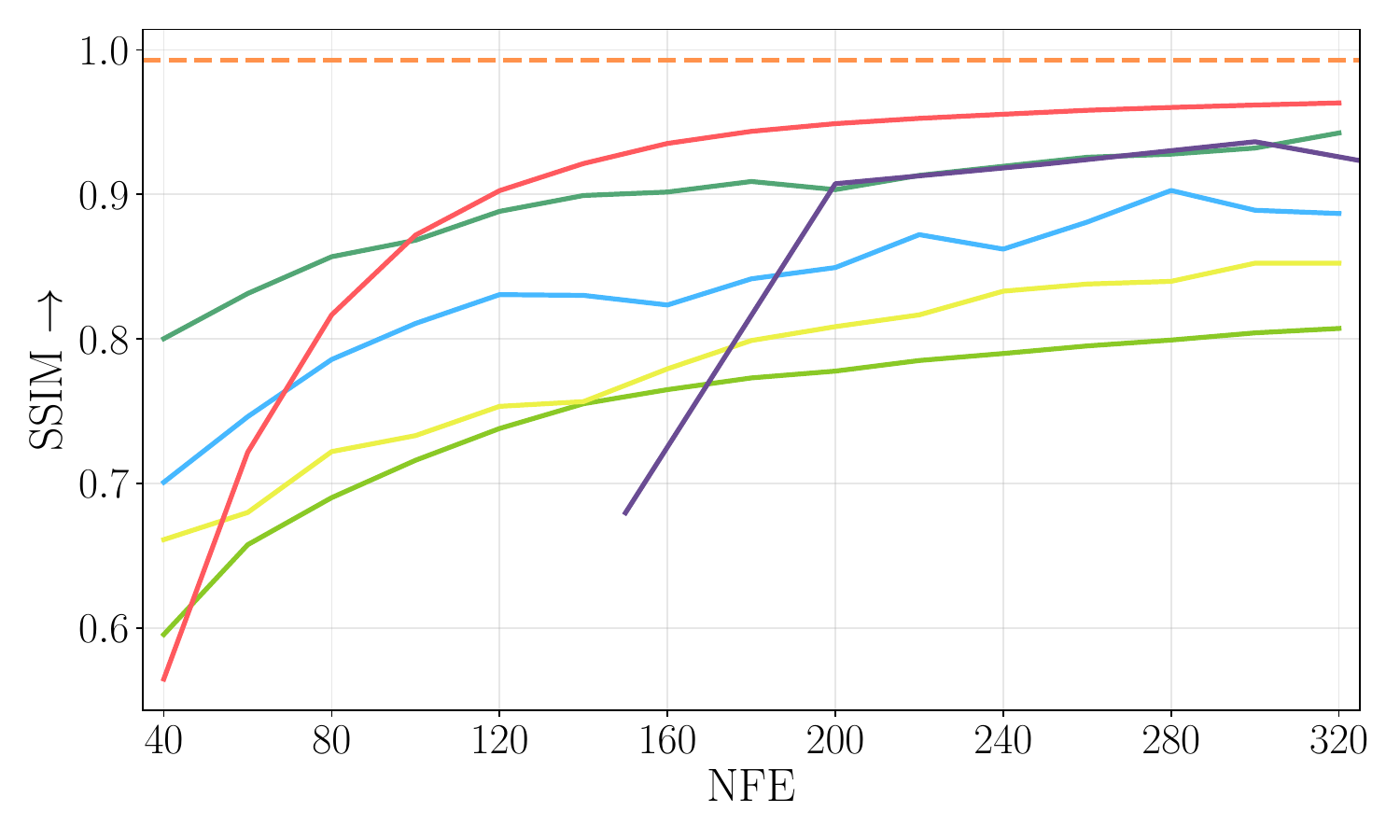}
  \includegraphics[width=.49\textwidth]{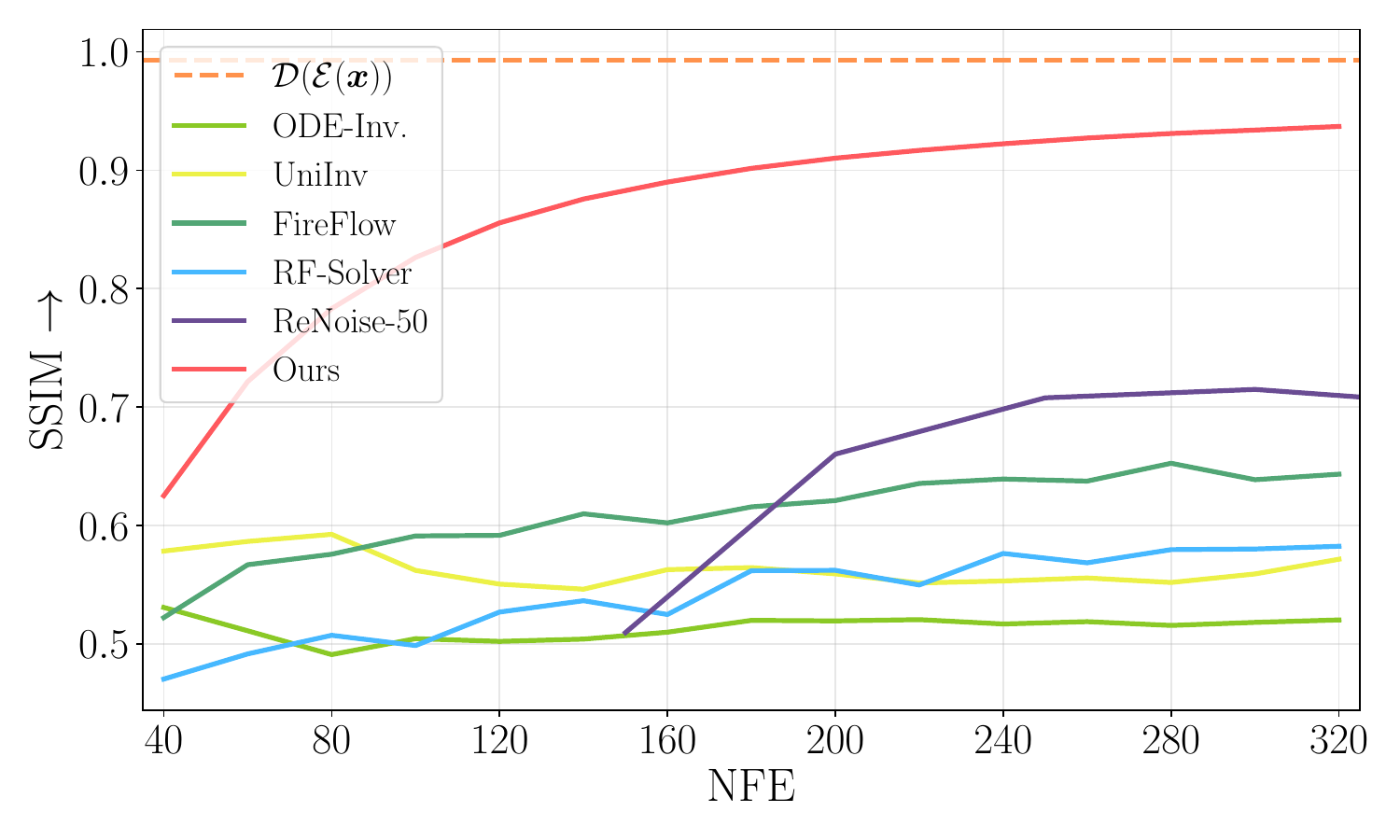}
\end{subfigure}
\hfill
\begin{subfigure}[t]{\linewidth}
  \centering 
  \includegraphics[width=.49\textwidth]{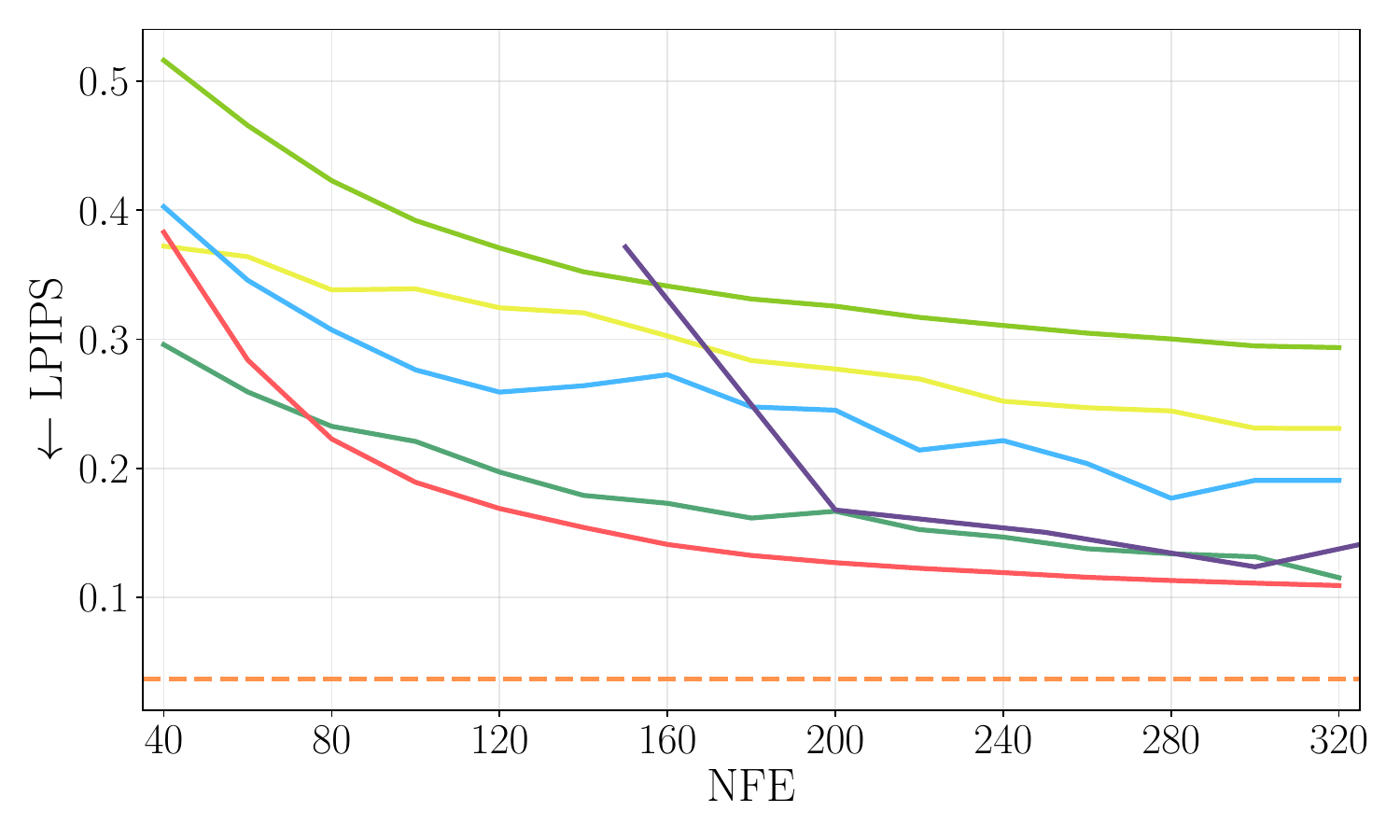}
  \includegraphics[width=.49\textwidth]{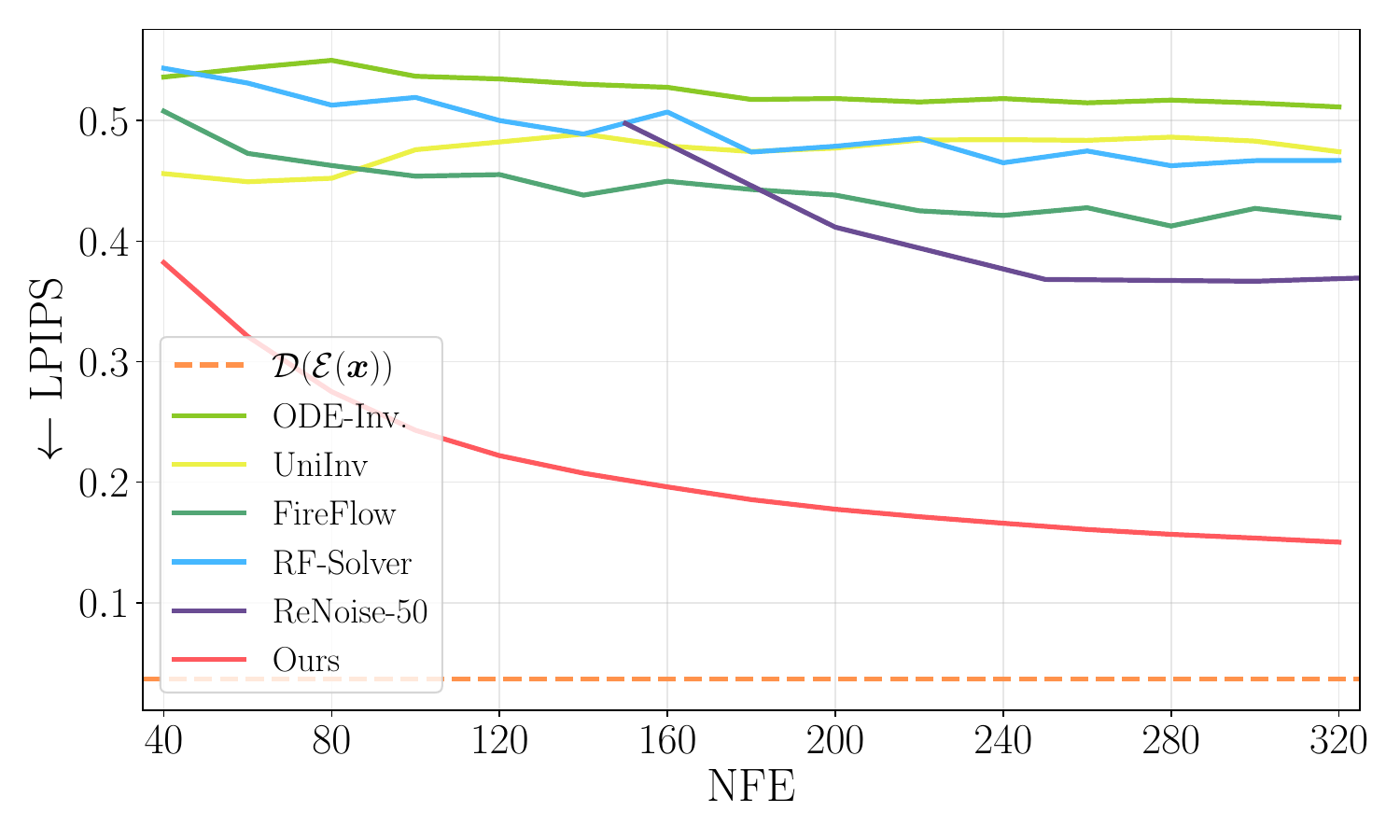}
\end{subfigure}
\caption{\textbf{Reconstruction quantitative comparisons (FLUX).} Pixel-space RMSE (first row), PSNR (second row), SSIM (third row), and LPIPS (last row) as functions of the number of NFEs for several inversion methods, for unconditional sampling with $\text{CFG} = 1$ (left) and with $\text{CFG} = 0$ (right). The dashed orange horizontal line is the average of forwarding the images through the encoder and decoder of the model.}
\label{fig:unconditional_reconstruction_flux}
\end{figure}

Figure \ref{fig:unconditional_reconstruction_flux} displays the results for the unconditional case for FLUX, evaluated by pixel-space RMSE, PSNR, SSIM and LPIPS as a function of the NFEs. The left column in this figure is obtained by using $\text{CFG} = 1$, and the right column is obtained by using $\text{CFG} = 0$. The details of this experiments are the same as in Sec.~\ref{sec:experiments}.

\subsection{Image editing}
\subsubsection{Additional qualitative comparisons}
\begin{figure}[htbp]
    \centering
    \includegraphics[width=\linewidth]{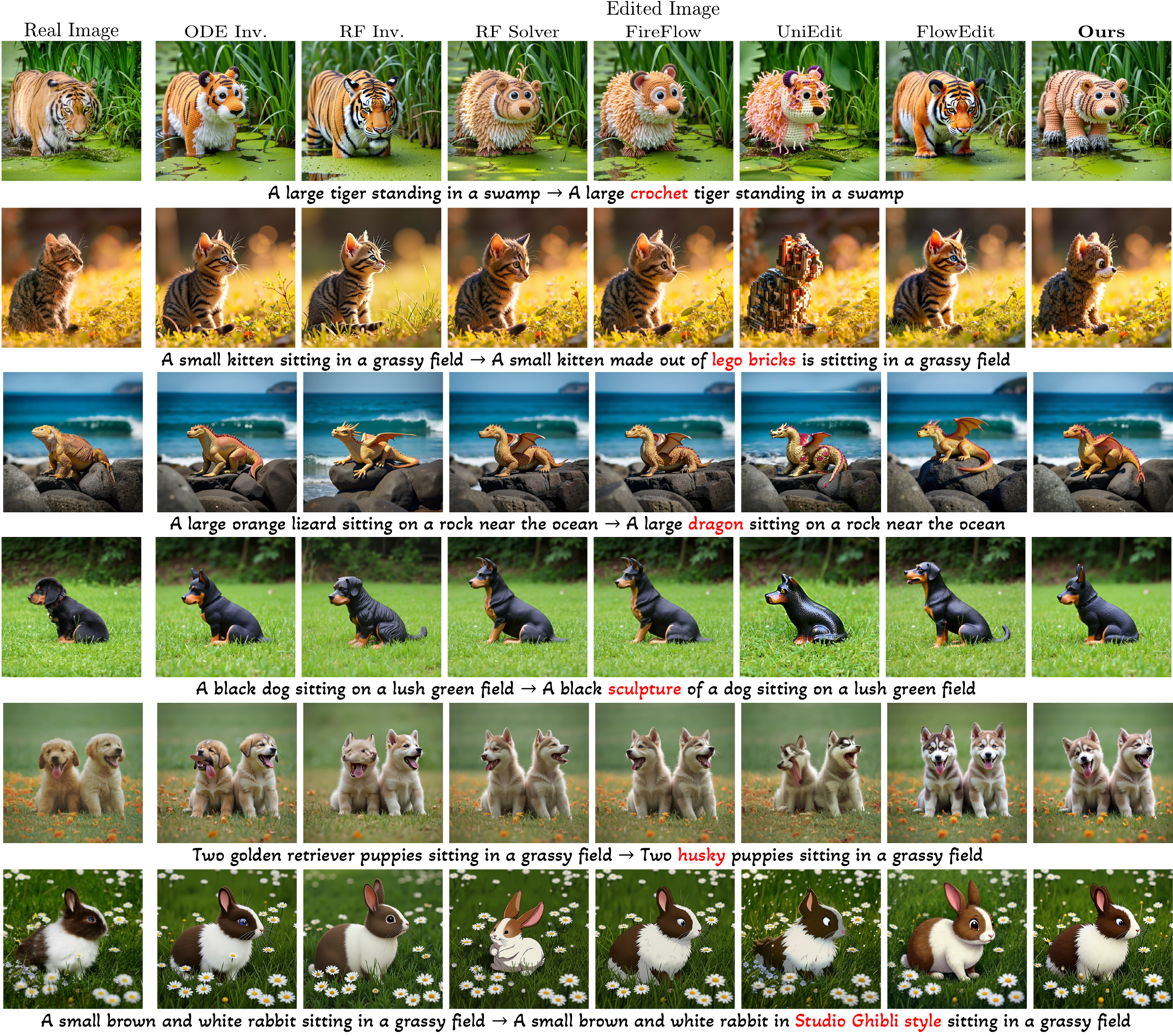}
    \caption{\textbf{Additional qualitative comparisons (FLUX).} Fine details are visible upon zooming in.}
    \label{fig:supp_qualitative_comparison_flux}
\end{figure}

Figure \ref{fig:supp_qualitative_comparison_flux} presents additional comparisons on image editing. We can see that \oursshort{} achieves consistently the best results both in terms of source image adherence and in terms of text adherence. For example, in the third row our method is the only one that managed to preserve the background and the structure of the rocks in the foreground. Similarly, in the fifth row, our method is the only one that preserved the posture of the dogs.

\subsubsection{Details of the experiment settings}
\begin{figure}[htbp]
\centering
\begin{subfigure}[t]{\linewidth}
  \centering 
  \includegraphics[width=.327\textwidth]{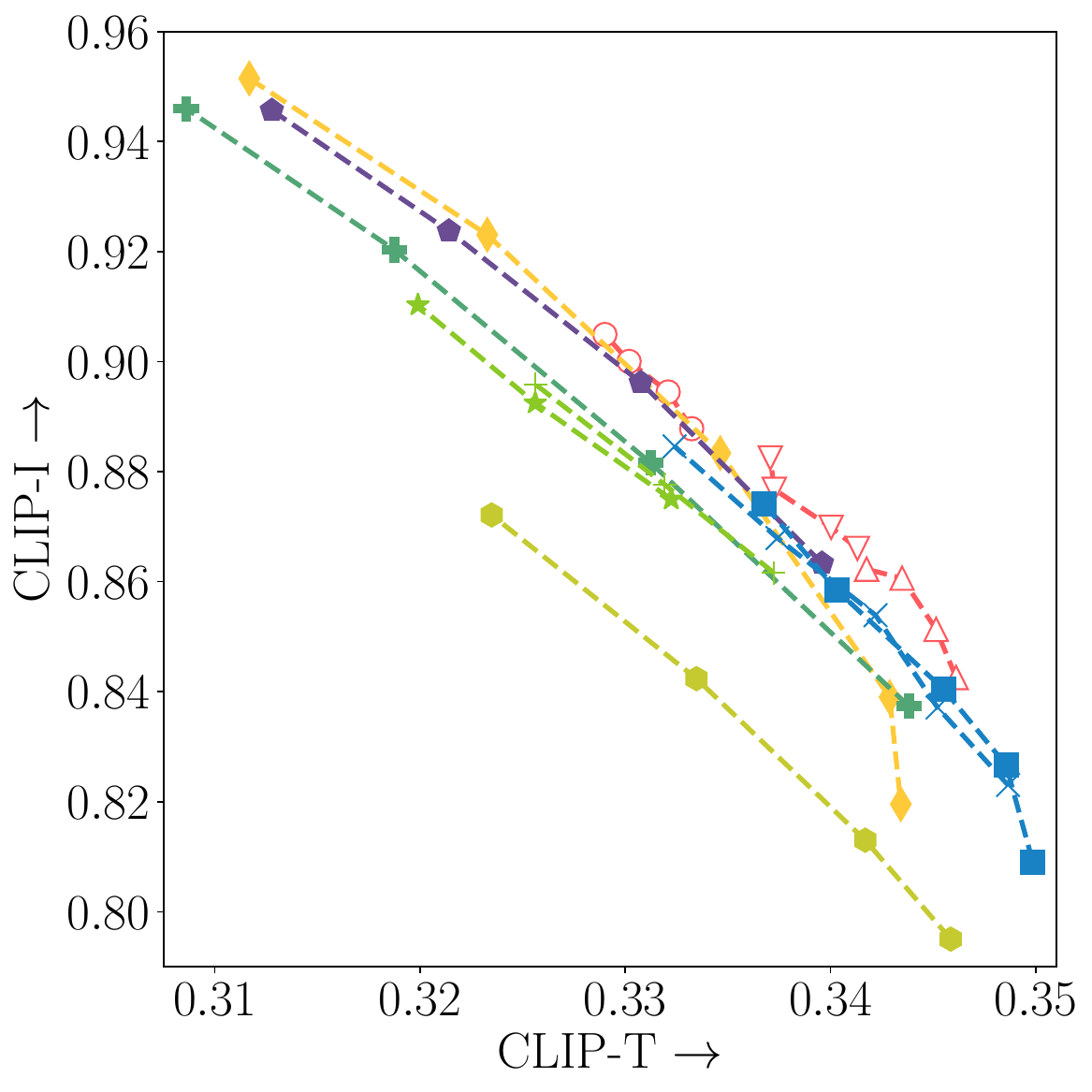}
  \includegraphics[width=.327\textwidth]{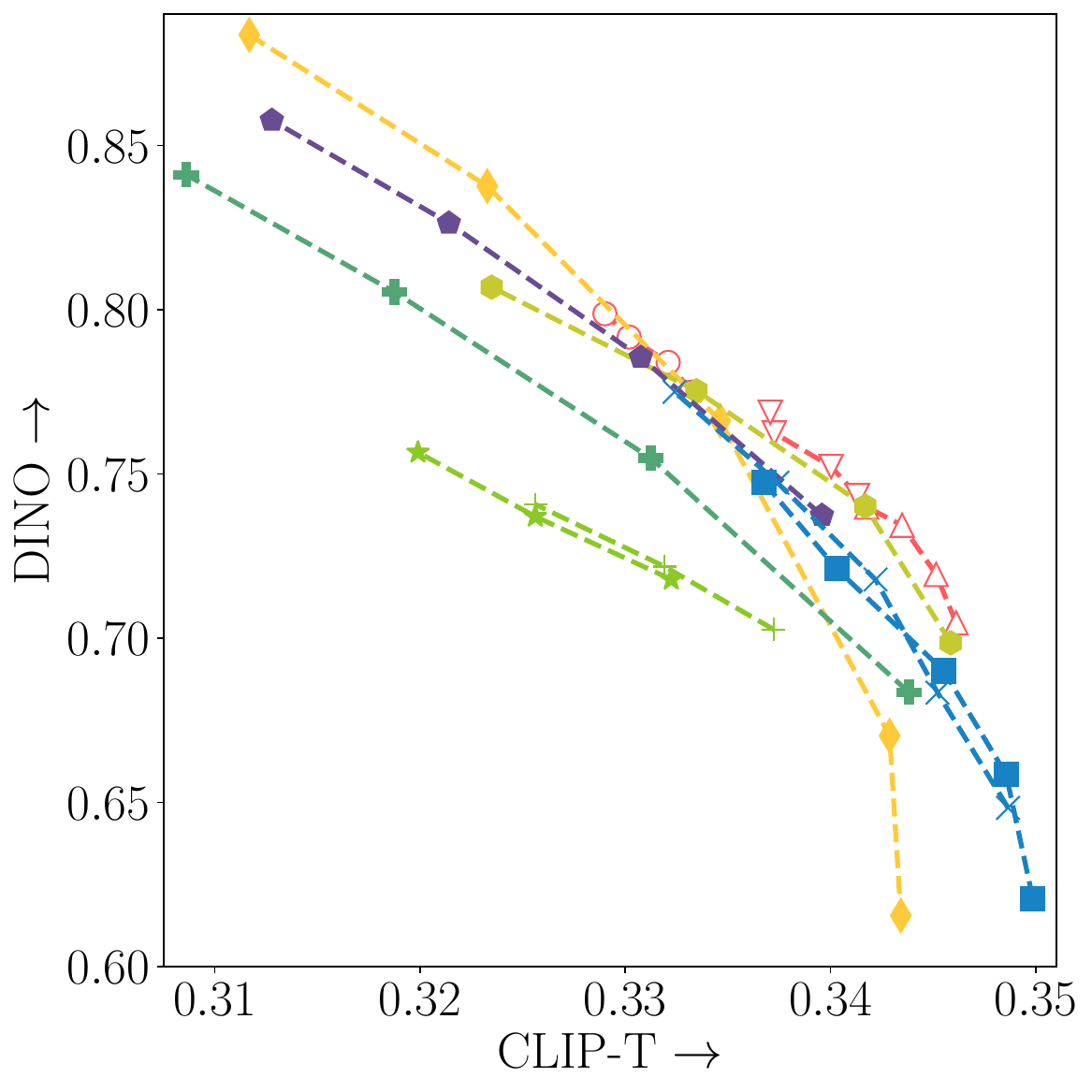}
  \includegraphics[width=.327\textwidth]{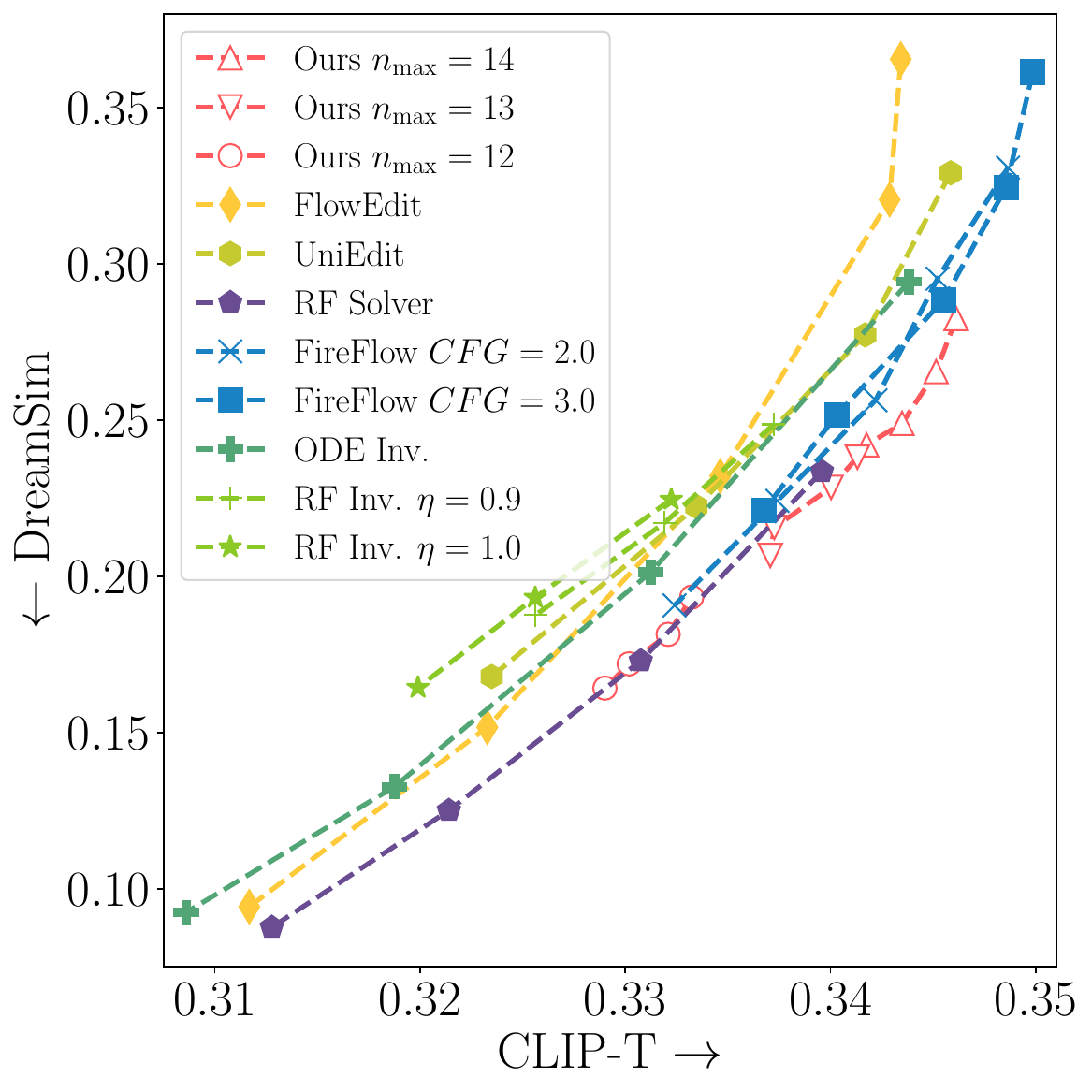}
\end{subfigure}
\caption{\textbf{Editing quantitative comparisons (FLUX).}  Text adherence is measured by CLIP-Text (x-axis) for all figures. Image adherence (y-axis) is measured by CLIP-Image (left), DINOv3 (center), and DreamSim (right). Connected markers represent different hyperparameters.}
\label{fig:supp_editing_quantitative_flux}
\end{figure}

Figure~\ref{fig:main_qualitative_comparison_flux} compares between all methods in terms of text adherence (CLIP Text) and image adherence measures (CLIP Image, DINOv3 and DreamSim). Figure~\ref{fig:supp_editing_quantitative_flux} provides more detailed comparisons between all methods.

\paragraph{FLUX hyperparameters.}
Table~\ref{tab:hyperparams_flux} lists the settings with which FlowEdit, ODE Inversion, and \oursshort{} were run in Fig.~\ref{fig:supp_editing_quantitative_flux}. The hyperparameters for all figures in the main text (except for Fig.~\ref{fig:teaser}) are marked in bold in this table.

\FloatBarrier

\begin{table*}[htbp!]
\centering
\caption{\textbf{FLUX hyperparameters.}}
\label{tab:hyperparams_flux}
\begin{tabular}{@{}ccccccc@{}}
\toprule
 & $T$ & $n_{\max}$ & CFG @ source & CFG @ target & $N$ iterations   \\ \midrule
FlowEdit & $28$ & $27,\;26,\;\textbf{24},\;22,\;20$ & $1.5$ & $5.5$ & -  \\
ODE Inversion & $50$ & $45,\;\textbf{40},\;35,\;30$  & $1$ & $3.5$ & -  \\
\oursshort{} & $15$ & $14,\;\textbf{13},\;12$  & $1$ & $3.5$ & $2,\;3,\;4,\;5$  \\
\bottomrule
\end{tabular}
\end{table*}

For UniEdit, the evaluated hyperparmeters are presented in Tab.~\ref{tab:hyperparams_uniedit_flux}, with the chosen value for their $\alpha$ parameter marked in bold. In our notation, $\alpha = n_{\max}/ T$.

\begin{table*}[htbp!]
\centering
\caption{\textbf{FLUX UniEdit hyperparameters.}}
\label{tab:hyperparams_uniedit_flux}
\begin{tabular}{@{}ccc@{}}
\toprule
 $T$ & $\alpha$ delay rate & $\omega$ guidance scale  \\ \midrule
 $15$ & $\frac{2}{5},\;\frac{2}{3},\;\mathbf{\frac{11}{15}},\;\frac{3}{5}$ & $5$  \\
\bottomrule
\end{tabular}
\end{table*}

For RF-Solver and FireFlow, the hyperparameters that were evaluated are presented in Tab.~\ref{tab:hyperparams_rf_solver_fireflow}, following their official implementation.

\begin{table*}[htbp!]
\centering
\caption{\textbf{RF-Solver and FireFlow hyperparameters.}}
\label{tab:hyperparams_rf_solver_fireflow}
\begin{tabular}{@{}ccccc@{}}
\toprule
 & $T$ & CFG & Injection step \\ \midrule
RF-Solver & $15$ & $2$ & $\textbf{2},\;3,\;4,\;5$  \\
FireFlow & $30$ & $\textbf{2},\;3$ & $1,\;2,\;3,\;\textbf{4},\;5$  \\
\bottomrule
\end{tabular}
\end{table*}

For RF-Inversion, the paper provides a set of hyperparameters for each kind of editing. We evaluate the sets of hyperparameters provided in their supplementary material. These are reported in Tab.~\ref{tab:hyperparams_rf_inv}. We choose the set that achieved the best results.

\begin{table*}[htbp!]
\centering
\caption{\textbf{RF-Inversion hyperparameters.}}
\label{tab:hyperparams_rf_inv}
\begin{tabular}{@{}ccccc@{}}
\toprule
$T$ & $s$ starting time & $\tau$ stopping time & $\eta$ strength \\ \midrule
$28$ &  $0$ & $\textbf{6},\;7,\;8$  & $\textbf{0.9}, 1.0$  \\
\bottomrule
\end{tabular}
\end{table*}

%% file: sections/supplementary/initialization.tex
\section{Initialization}
\label{ap:initialization}
\begin{figure}[htbp]
\centering
\begin{subfigure}[t]{\linewidth}
  \centering 
  \includegraphics[width=.327\textwidth]{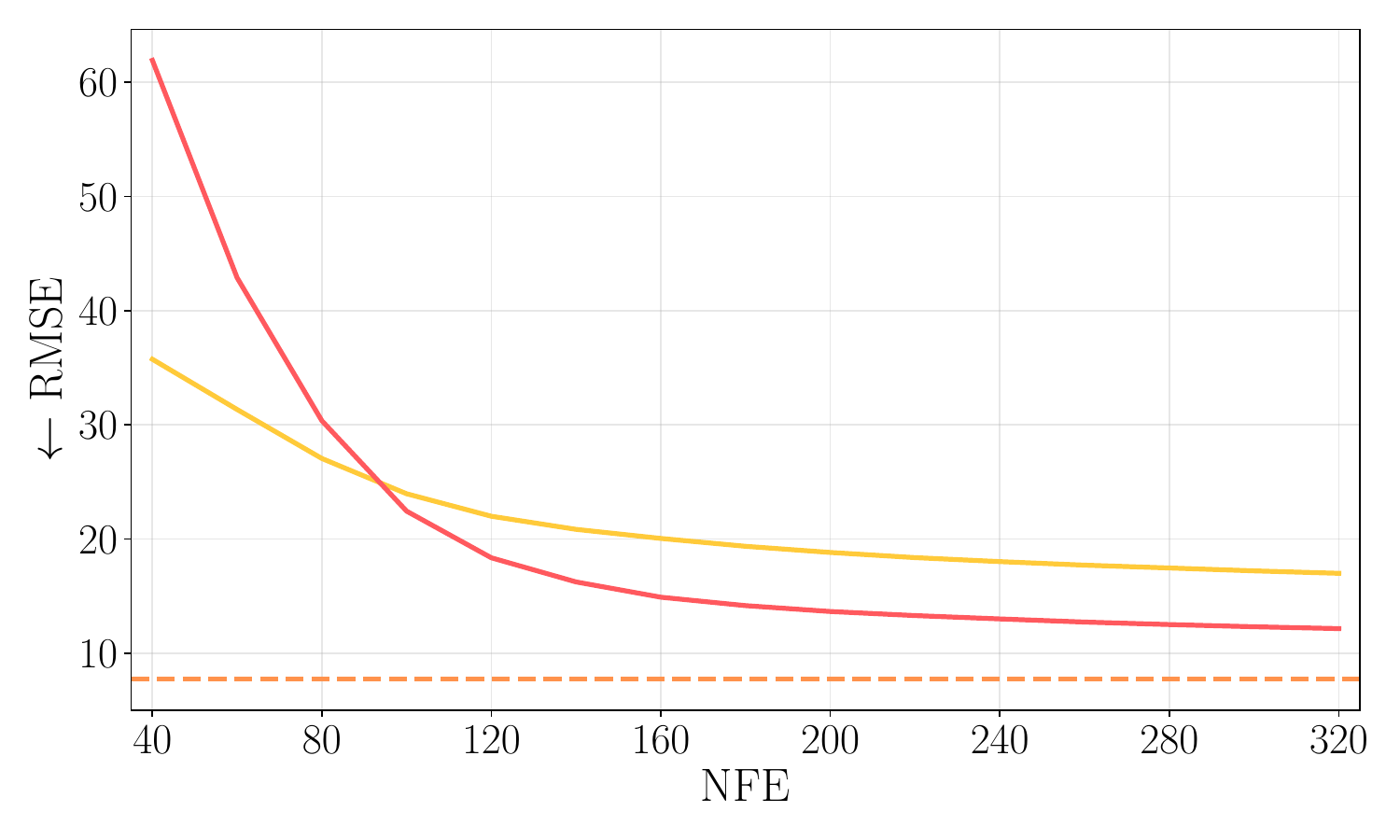}
  \includegraphics[width=.327\textwidth]{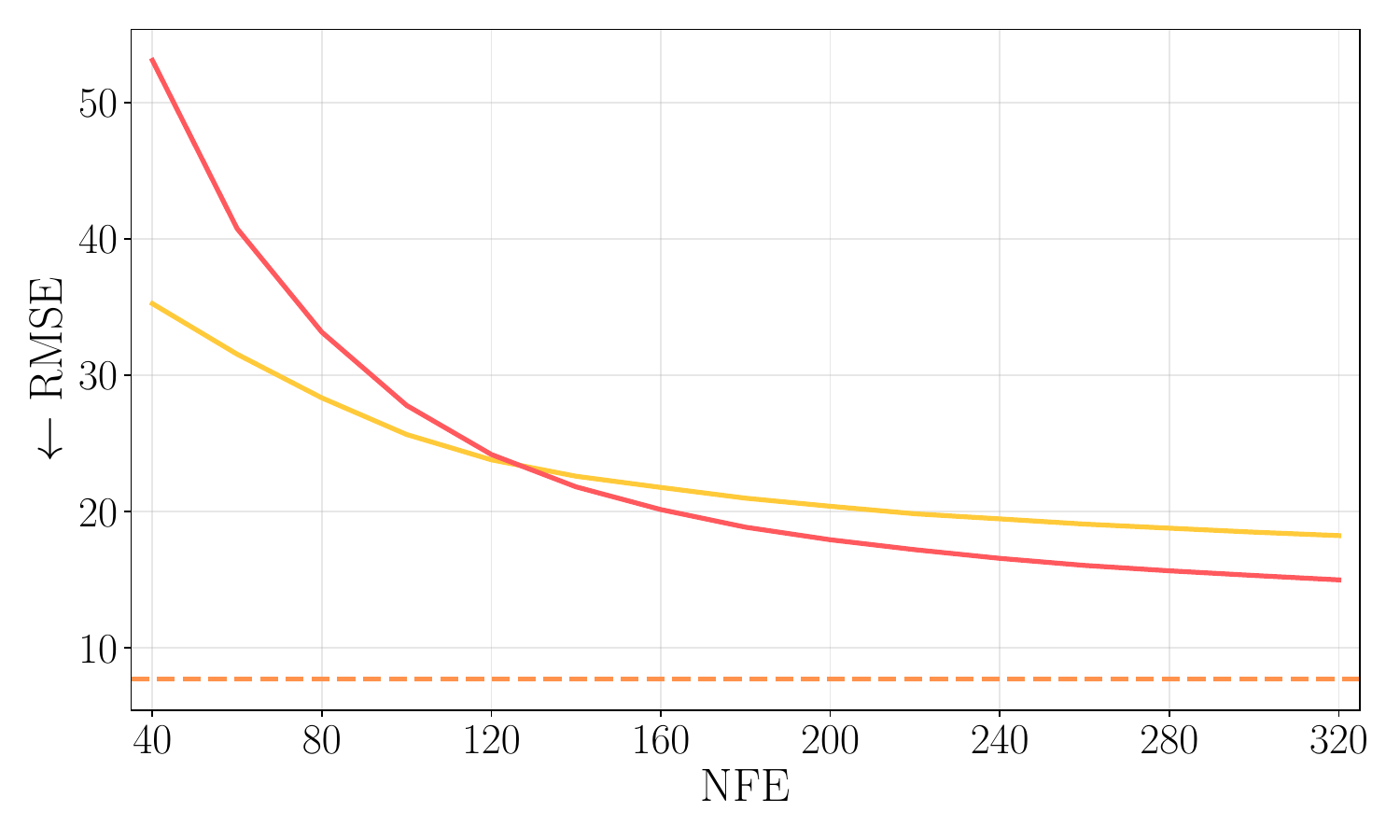}
  \includegraphics[width=.327\textwidth]{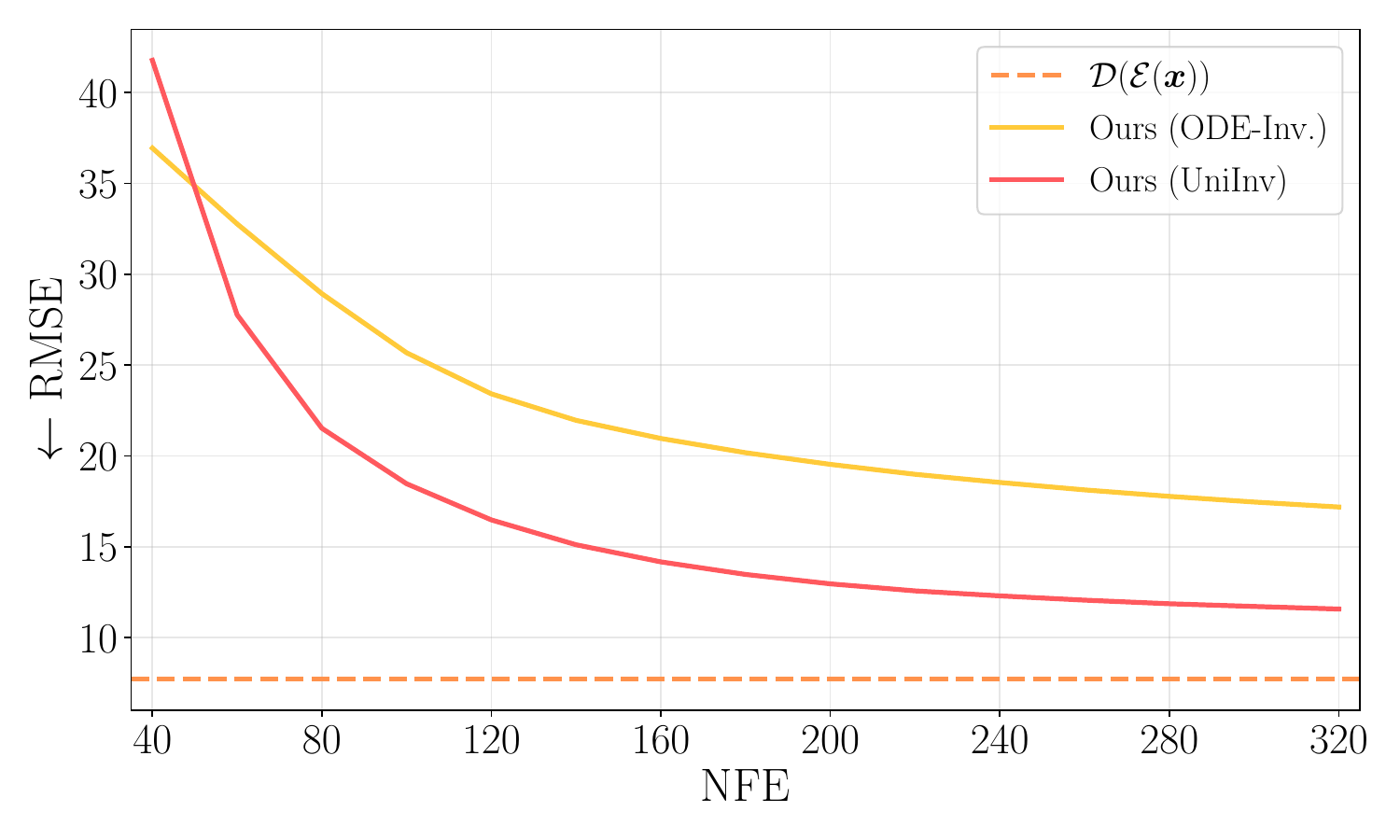}
\end{subfigure}
\hfill
\begin{subfigure}[t]{\linewidth}
  \centering 
  \includegraphics[width=.327\textwidth]{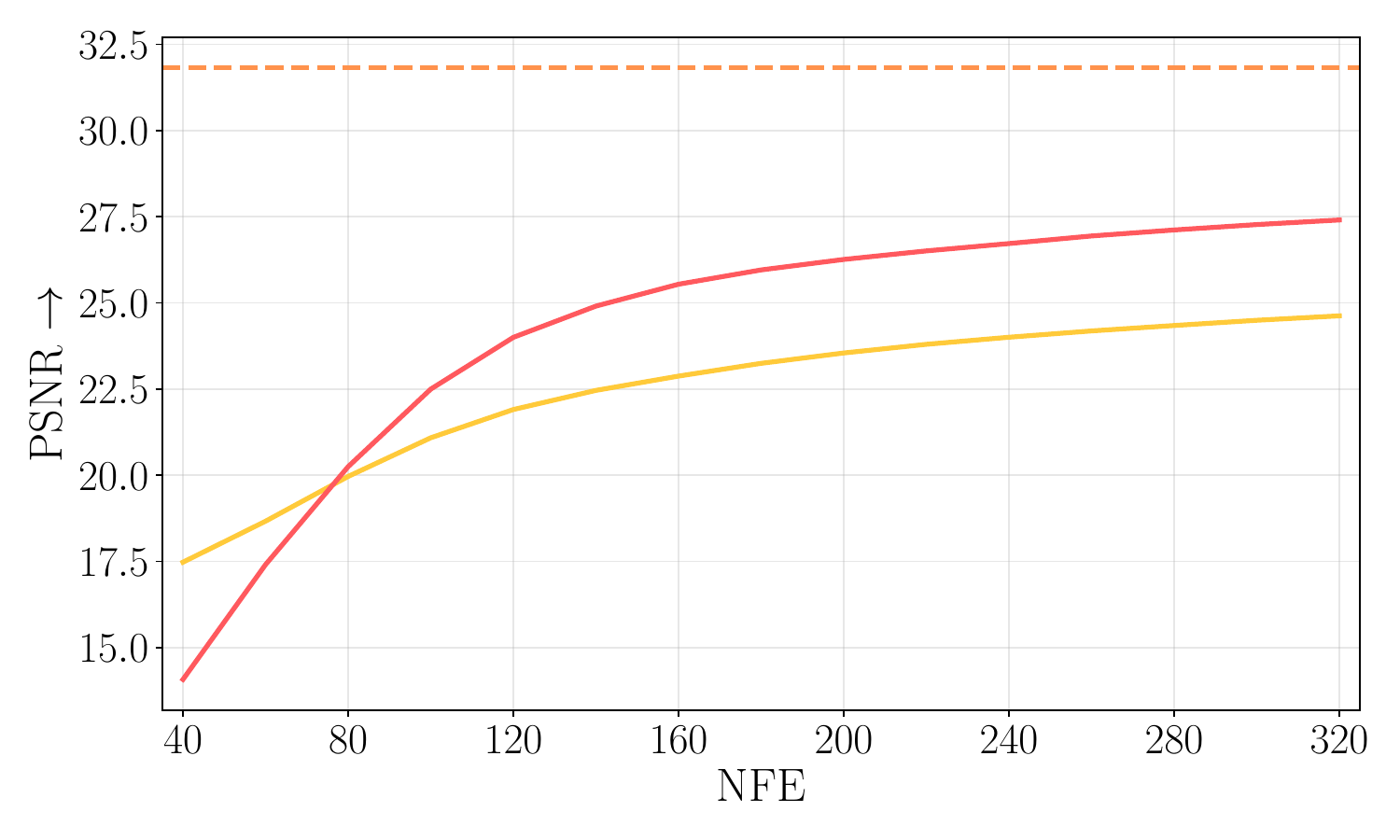}
  \includegraphics[width=.327\textwidth]{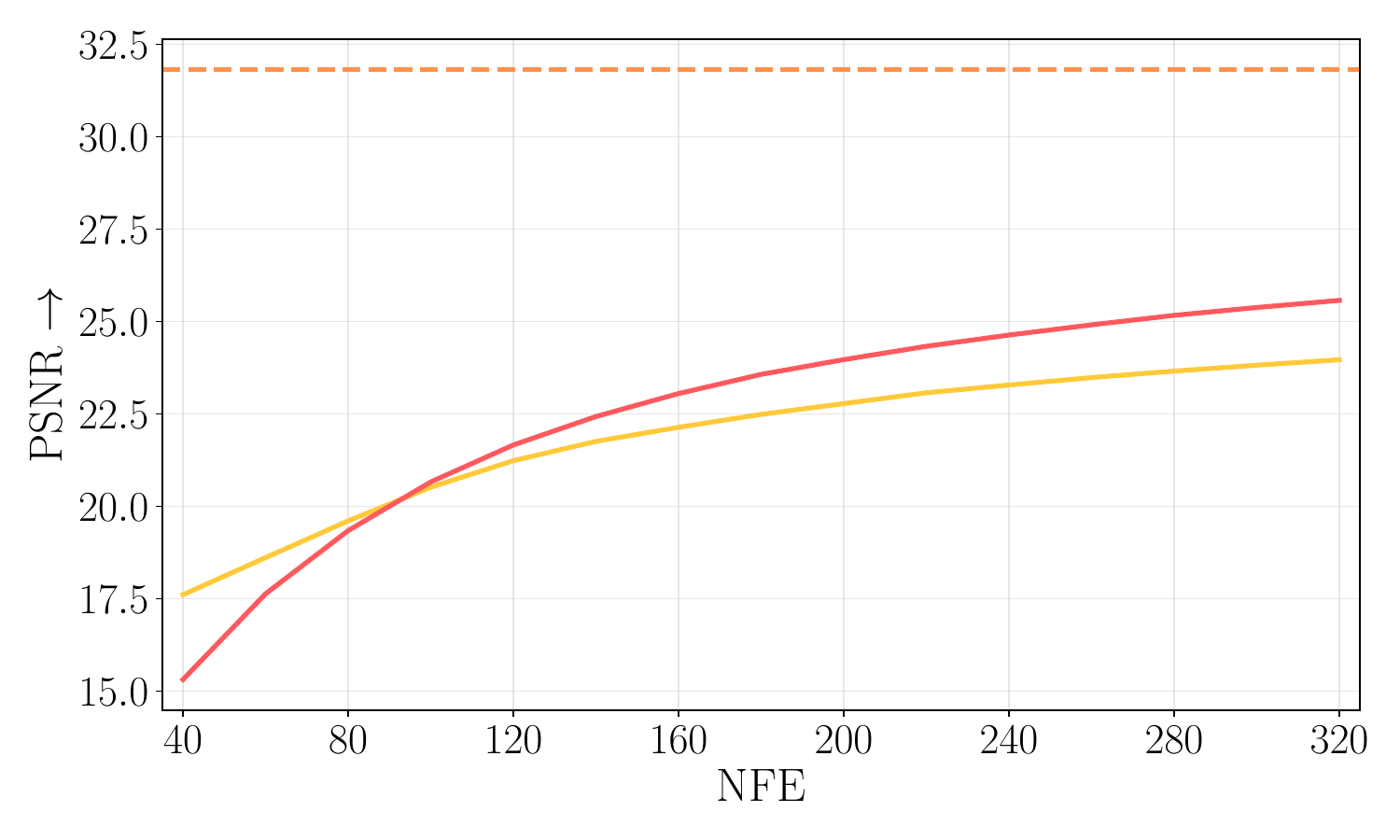}
  \includegraphics[width=.327\textwidth]{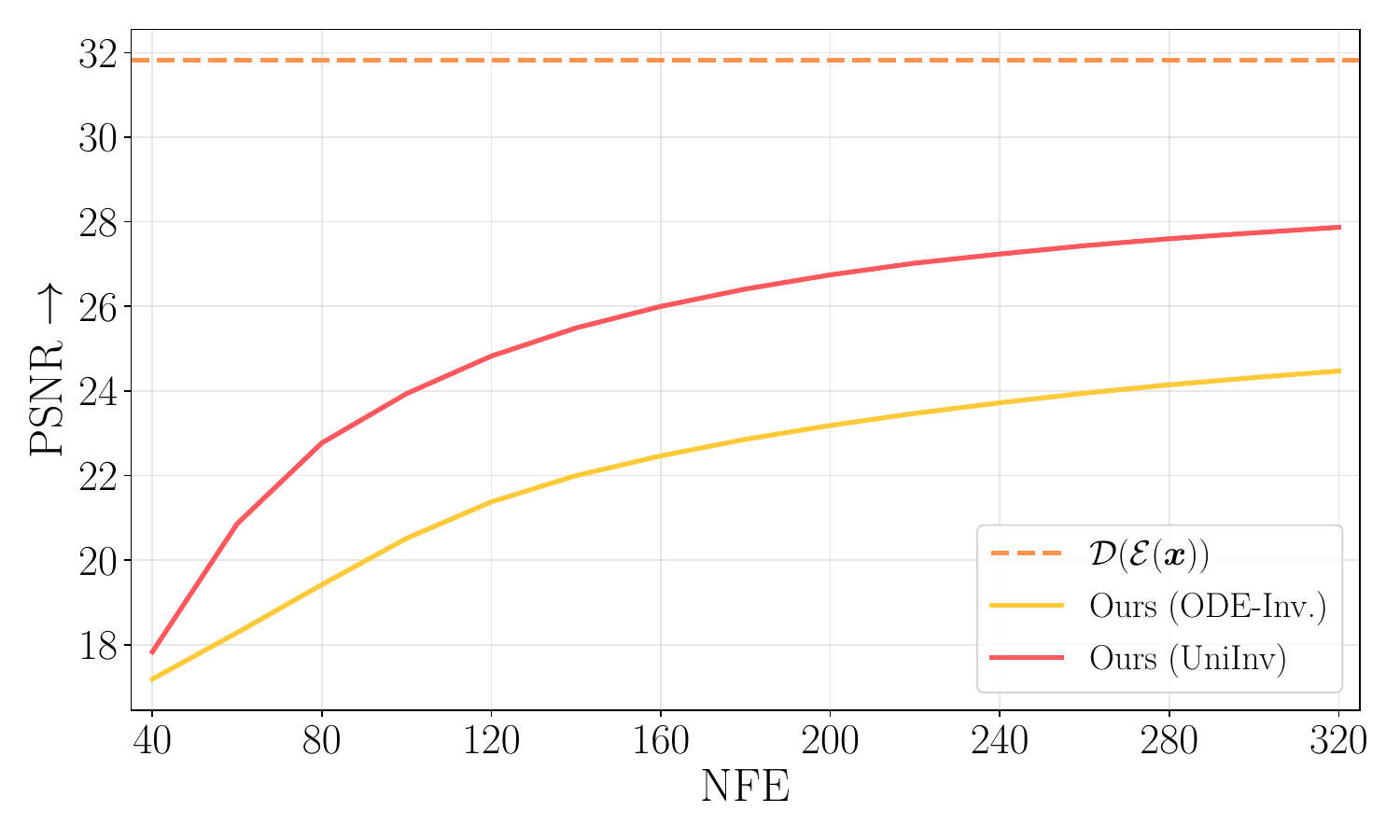}
\end{subfigure}
\hfill
\begin{subfigure}[t]{\linewidth}
  \centering 
  \includegraphics[width=.327\textwidth]{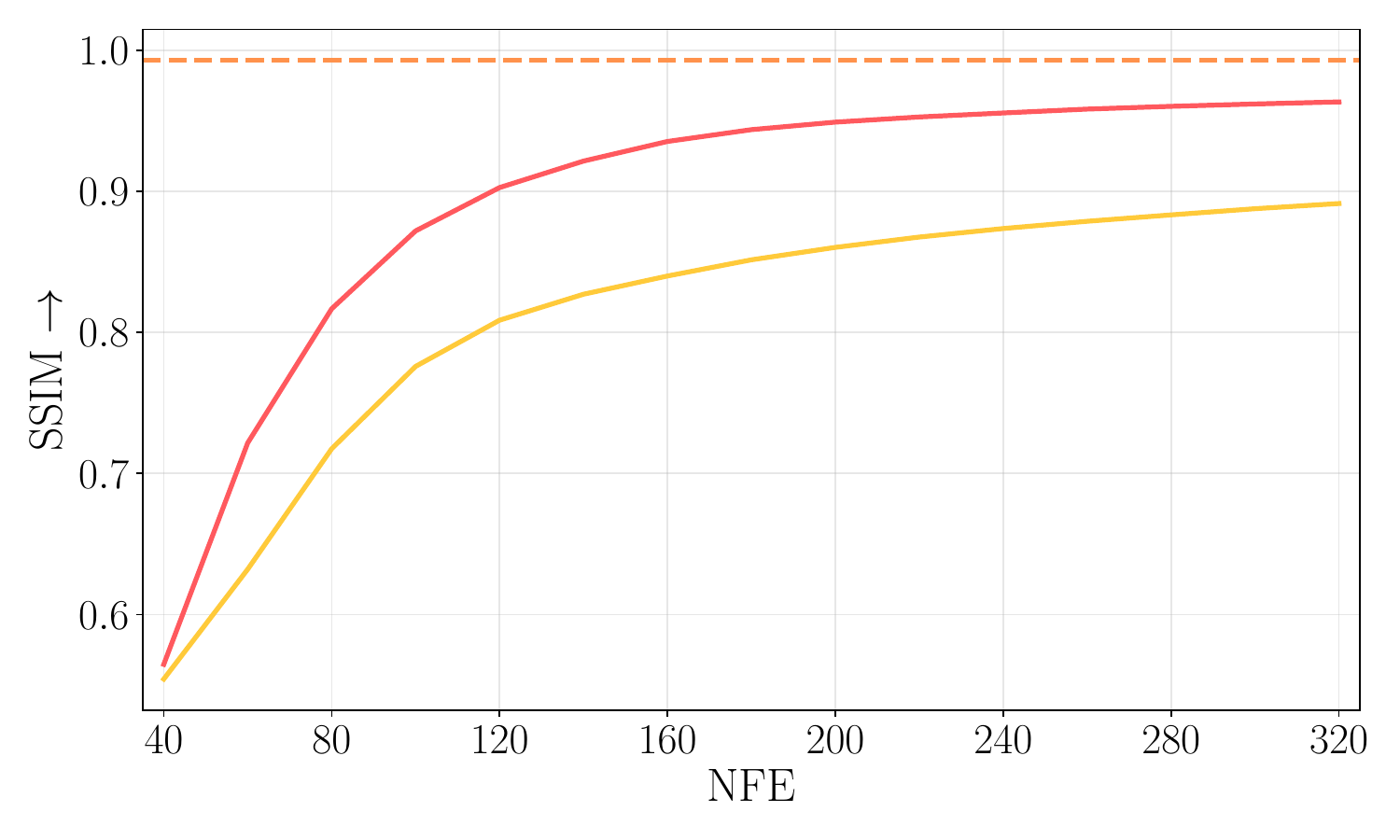}
  \includegraphics[width=.327\textwidth]{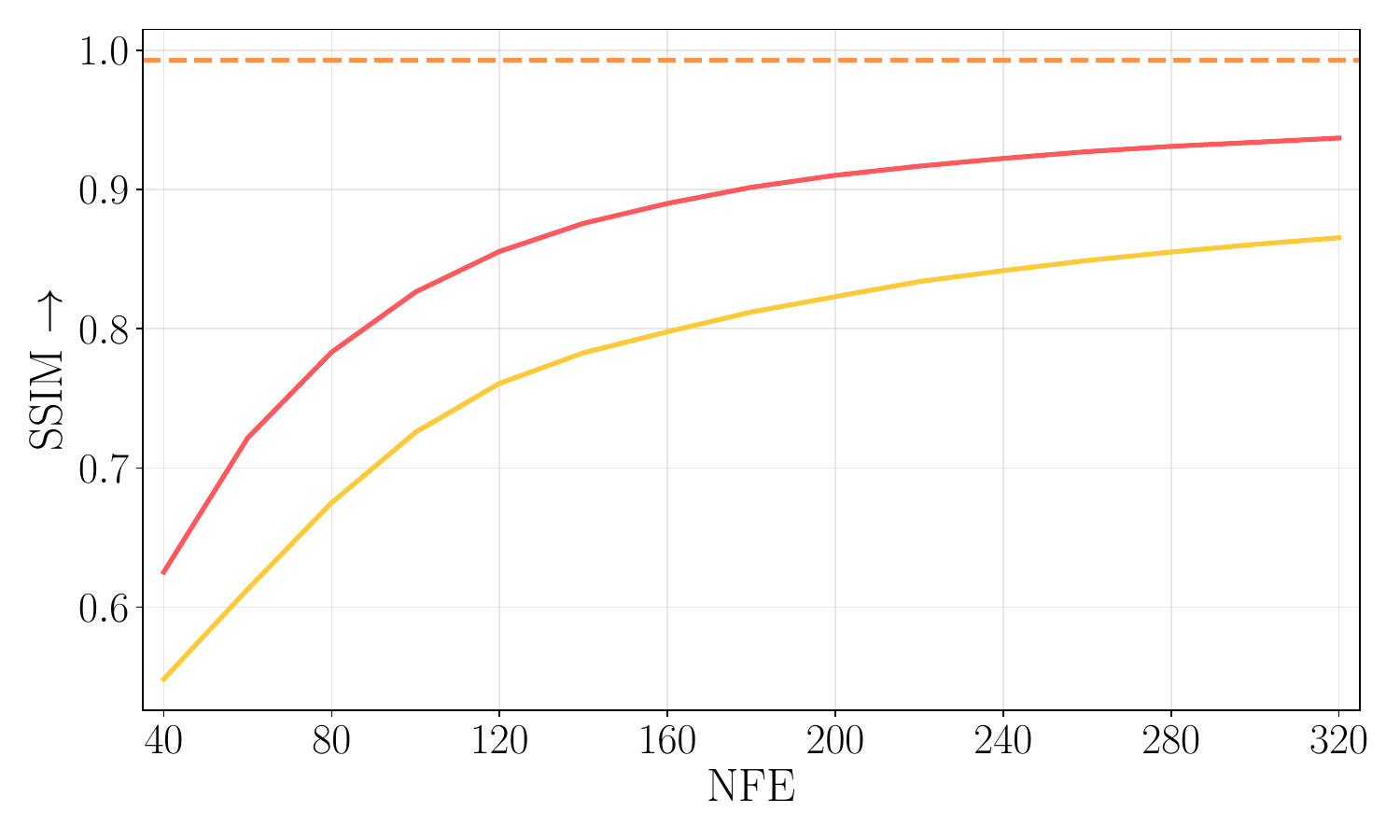}
  \includegraphics[width=.327\textwidth]{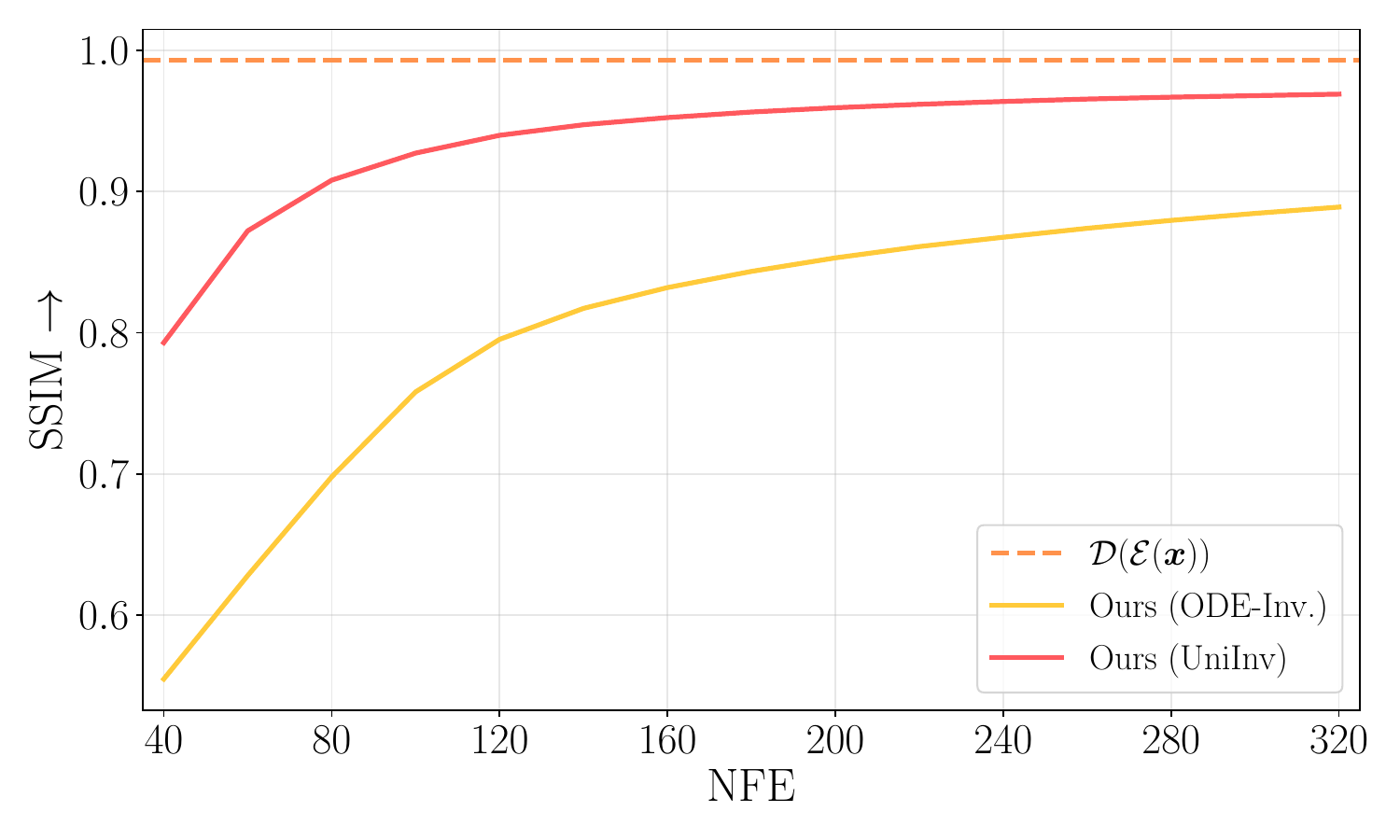}
\end{subfigure}
\hfill
\begin{subfigure}[t]{\linewidth}
  \centering 
  \includegraphics[width=.327\textwidth]{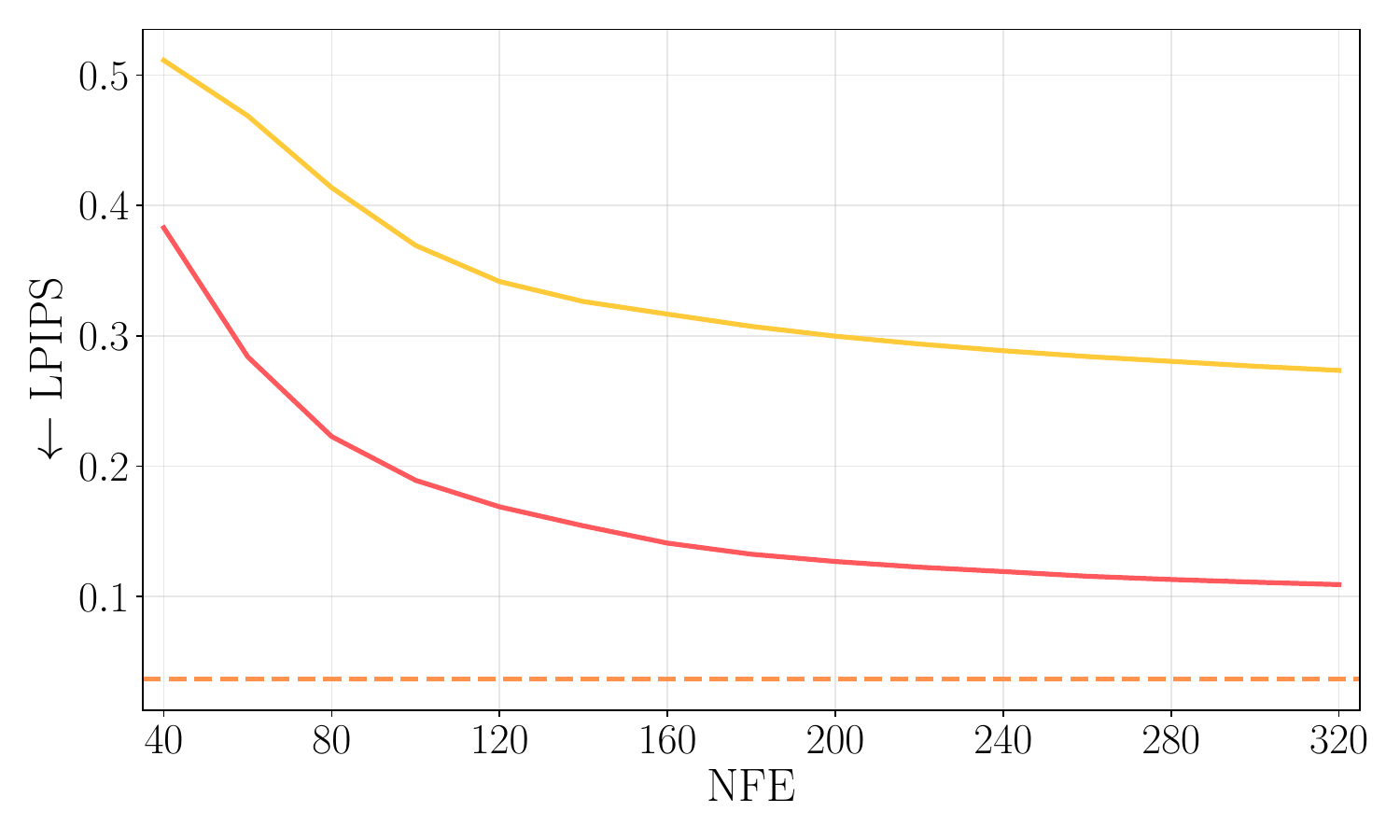}
  \includegraphics[width=.327\textwidth]{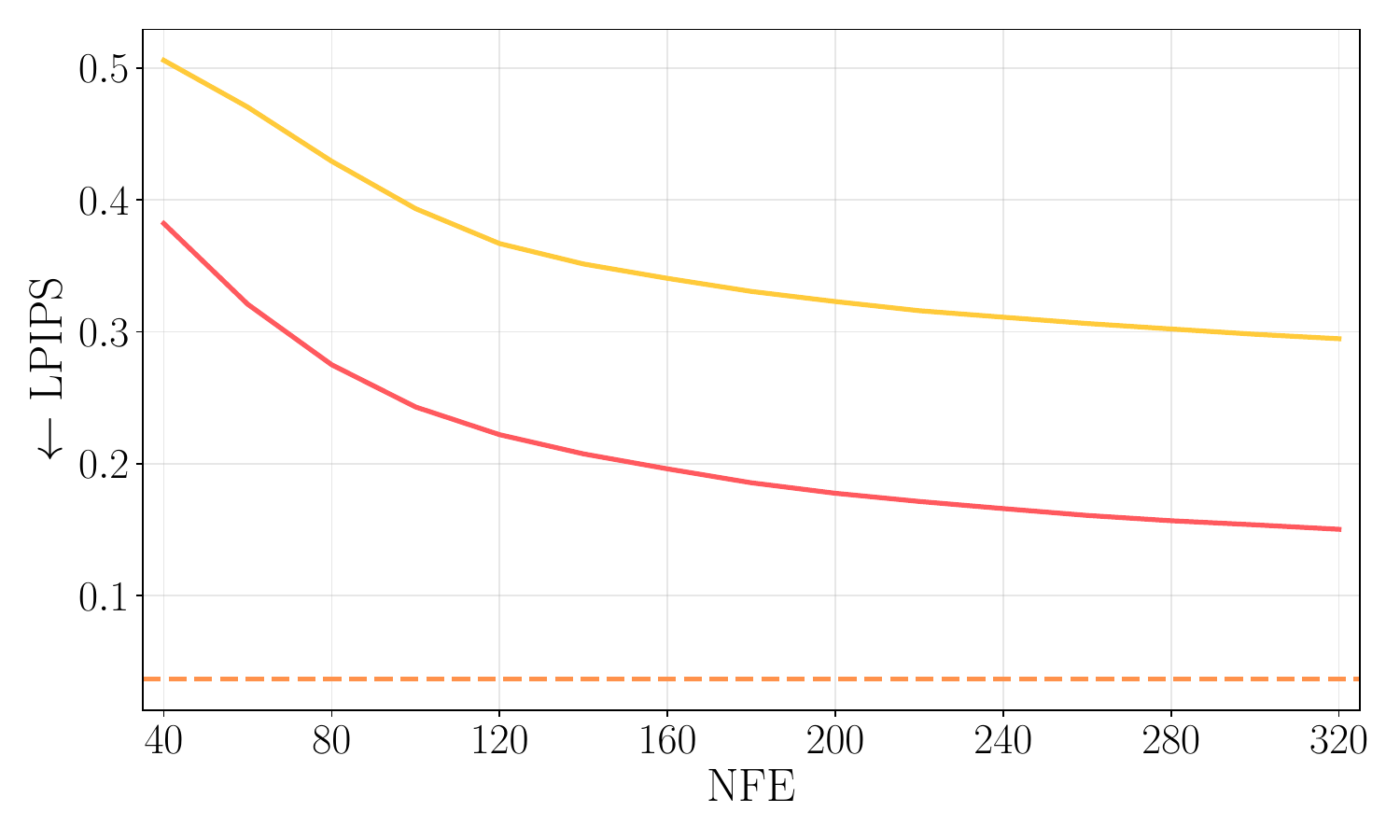}
  \includegraphics[width=.327\textwidth]{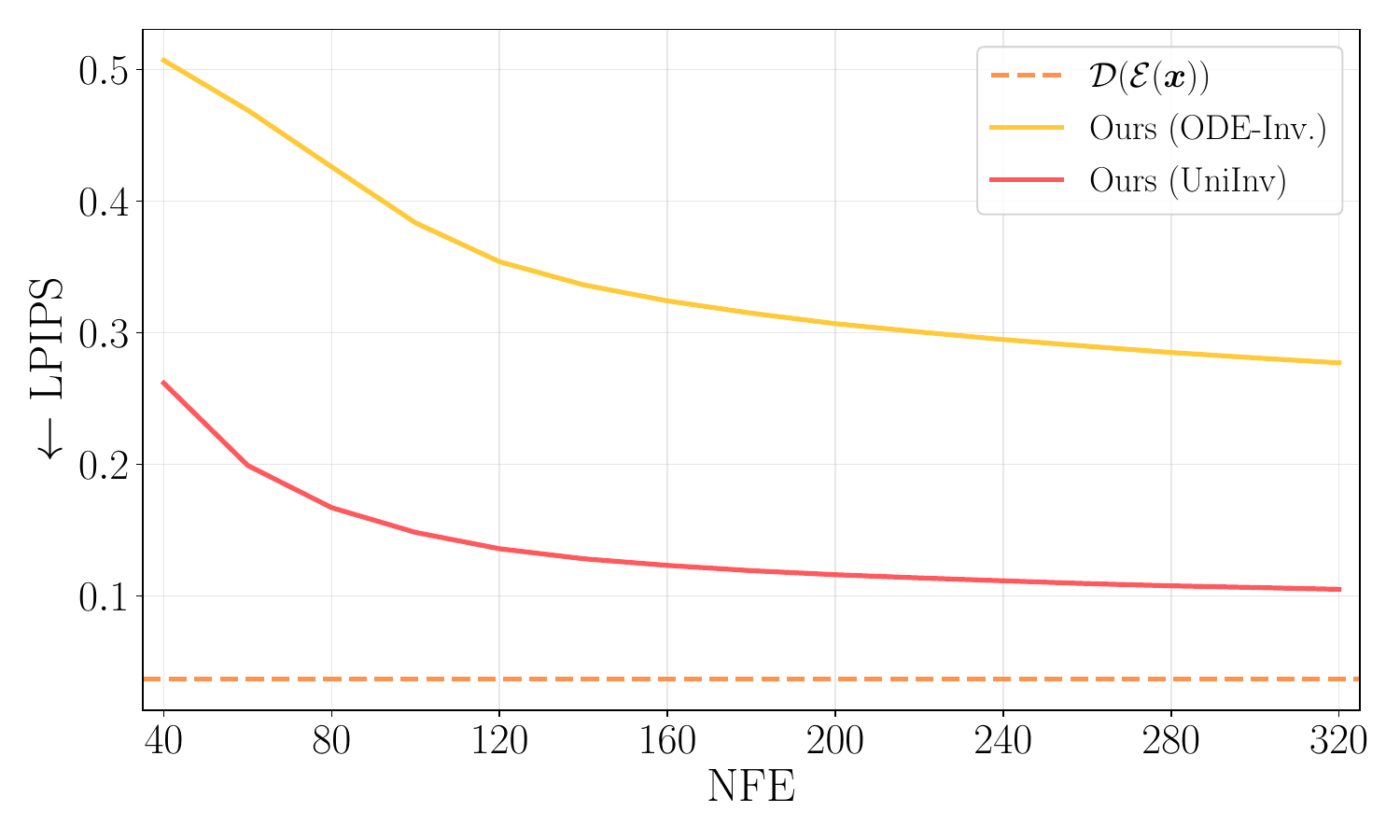}
\end{subfigure}
\caption{\textbf{Reconstruction quantitative comparisons (FLUX).}  Pixel-space RMSE (first row), PSNR (second row), SSIM (third row), and LPIPS (last row) as functions of the number of NFEs, for unconditional sampling with $\text{CFG} = 1$ (left), unconditional sampling with $\text{CFG} = 0$ (center), and text-conditional sampling (right). The red and yellow curves correspond to \oursshort{} initialized with the UniInv and ODE Inversion methods, respectively. The dashed orange horizontal line is the average of forwarding the images through the encoder and decoder of the model.}
\label{fig:reconstruction_flux_init}
\end{figure}

We proved that if the step size is chosen appropriately, then \oursshort{} necessarily converges to the unique global minimum of our optimization problem. However, for any finite number of iterations, the initialization does have an impact on the result.  This is illustrated in Figs.~\ref{fig:reconstruction_flux_init} and~\ref{fig:reconstruction_sd3}, where the red and yellow curves correspond to initialization with the UniInv and ODE Inversion methods, respectively.
As the results obtained with the UniInv initialization are better than with ODE inversion, we chose the former for all experiments in the paper.

\clearpage

%% file: sections/supplementary/sd3.tex
\section{Stable Diffusion 3 (SD3)}
\label{ap:sd3}
In this appendix, we repeat all the experiments of Sec.~\ref{sec:experiments}, but with SD3 instead of FLUX. In this case, we choose the step size $\eta = 10^{-2}$ in the update rule of \eqref{eq:mse_update_rule}.

\subsection{Image Reconstruction (Inversion)}
\paragraph{Implementation details.}
We use the implementation details provided for FLUX. We set the number of denoisers to $T = 10$, and evaluate the reconstruction error for various NFE values.

\paragraph{Dataset.}
We use the same dataset used for evaluating FLUX -- randomly chosen same $100$ real images of dimension $1024 \times 1024$ from the DIV2K dataset, automatically captioned by BLIP and manually refined.

\paragraph{Competing methods.}
As with the experiments on FLUX, we compare our method to ODE Inversion, RF-Solver, FireFlow and UniInv. For methods, like RF-Solver, which use two forward passes per timestep, we set $T = \frac{\text{NFE}}{4}$. For methods that use a single forward pass per timestep, we set $T = \frac{\text{NFE}}{2}$. We also evaluated ReNoise, with both $T = \{ 10, 28 \}$, and set the number of ReNoise steps so as to achieve the desired NFE count. We evaluated various hyperparameters for ReNoise and report the results with those that worked best. It should be noted that, for the fixed point iterations of ReNoise, one would expect that the final iteration would provide the best results. However, we observe that this does not necessarily happen in practice. 
We also note that as there was no official implementation for any method for SD3, we implemented all of them by ourselves.

\begin{figure}[ht]
\centering
\begin{subfigure}[t]{\linewidth}
  \centering
  \includegraphics[width=.49\textwidth]{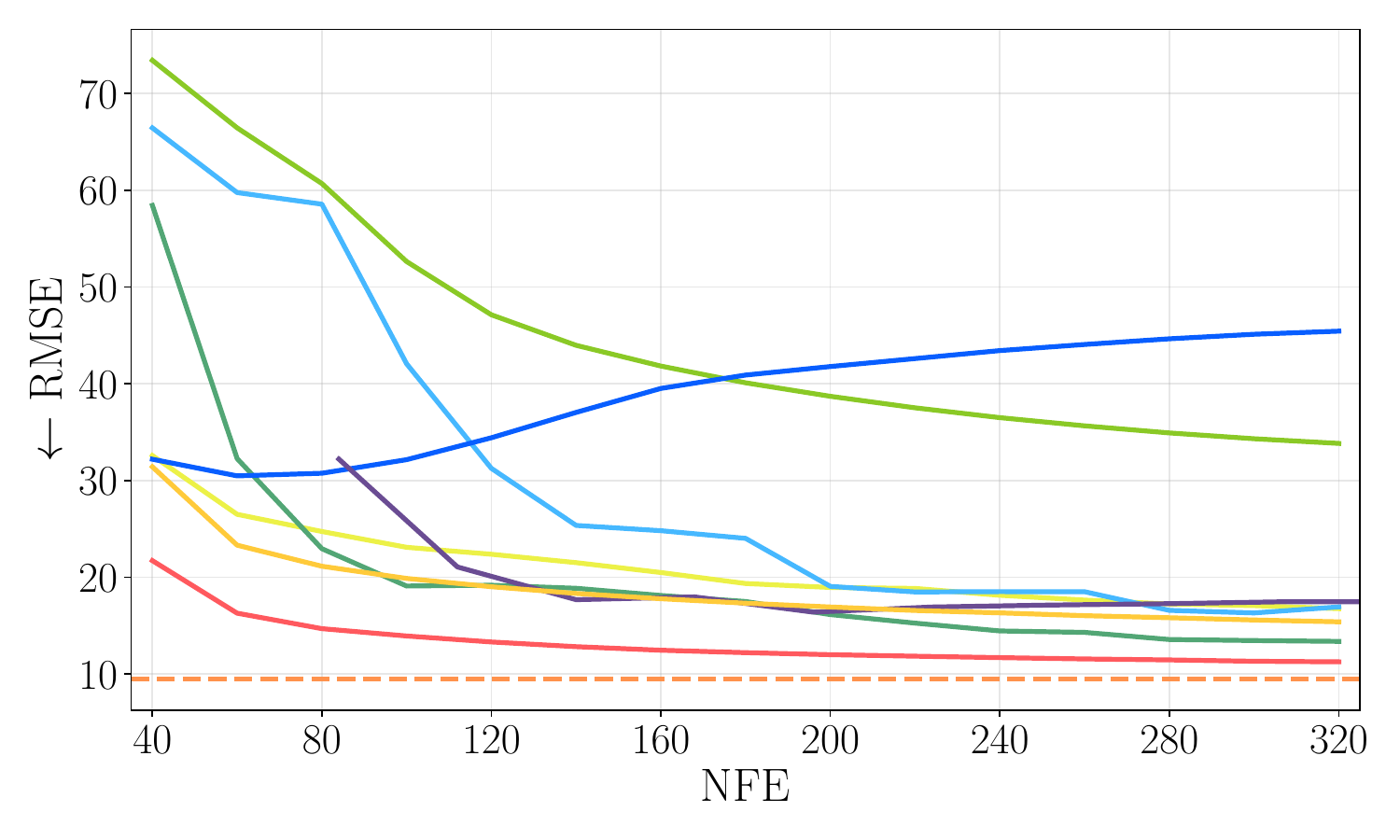}
  \includegraphics[width=.49\textwidth]{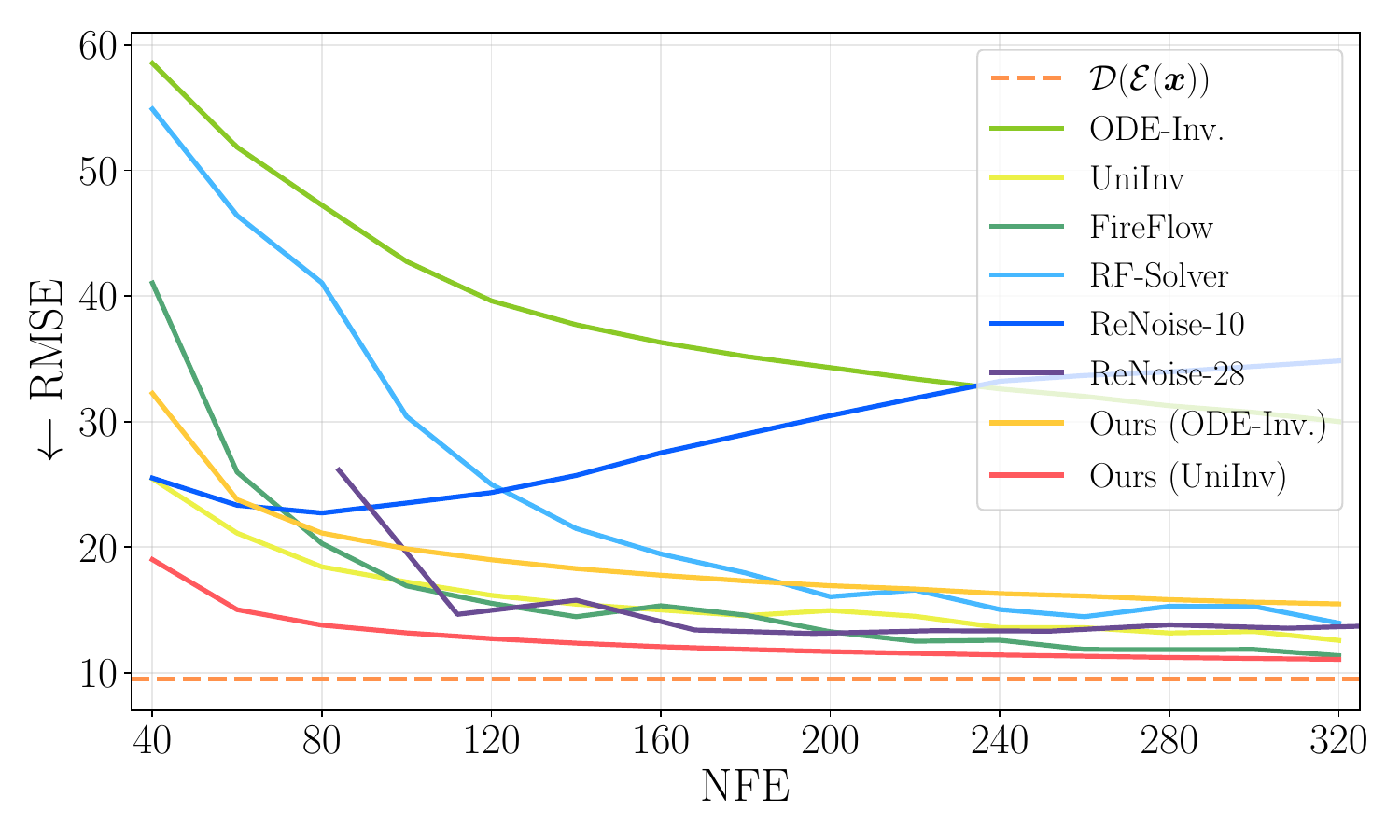}
\end{subfigure}
\begin{subfigure}[t]{\linewidth}
  \centering
  \includegraphics[width=.49\textwidth]{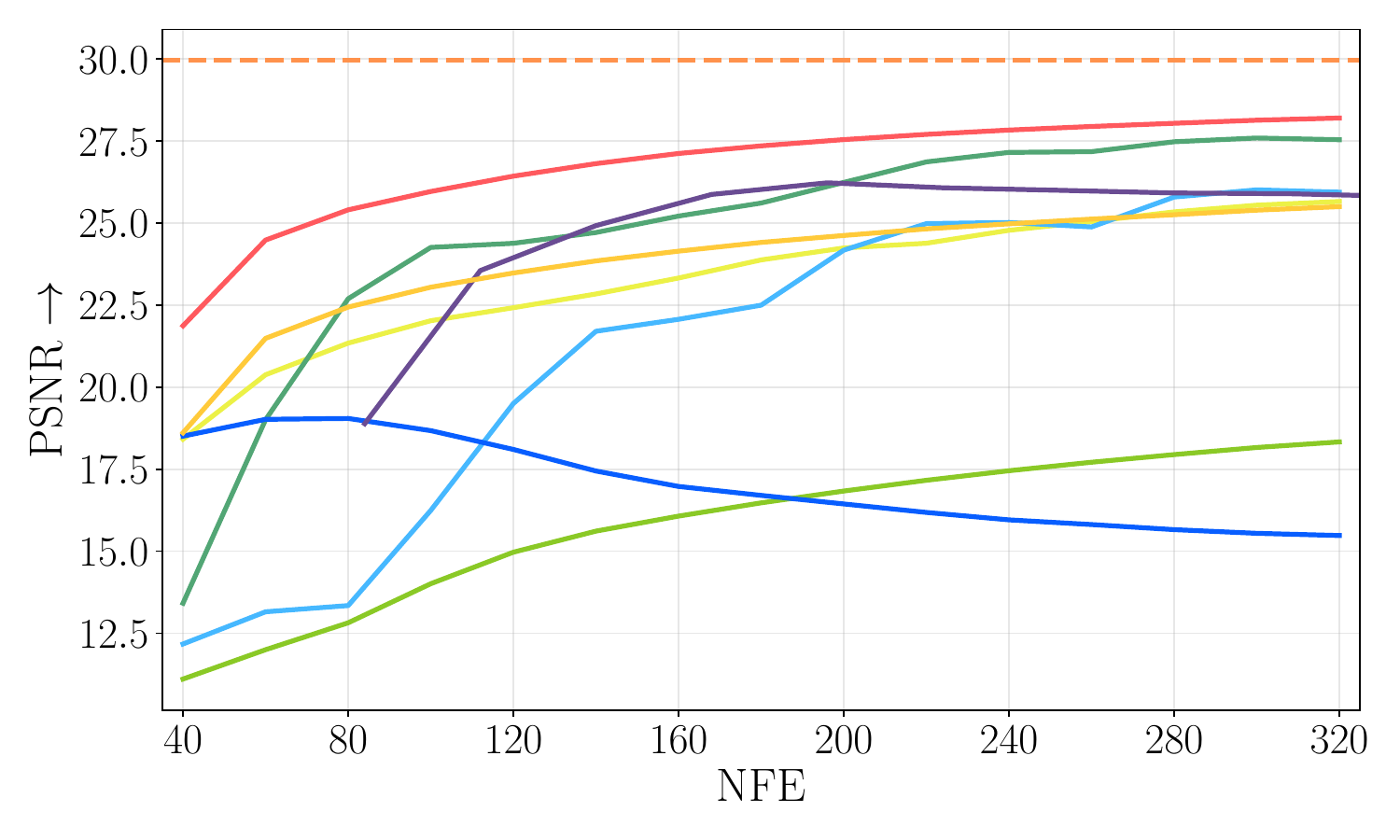}
  \includegraphics[width=.49\textwidth]{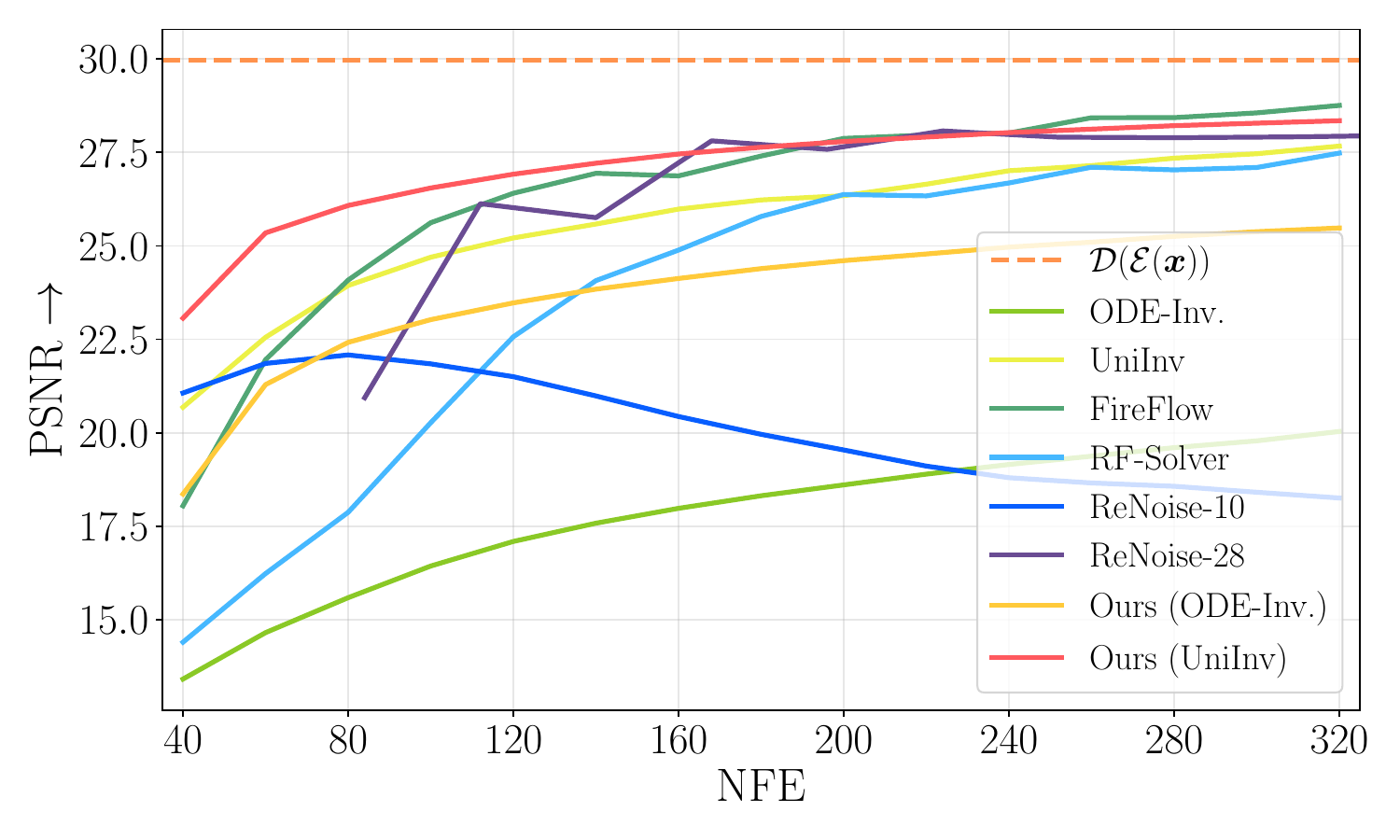}
\end{subfigure}
\begin{subfigure}[t]{\linewidth}
  \centering
  \includegraphics[width=.49\textwidth]{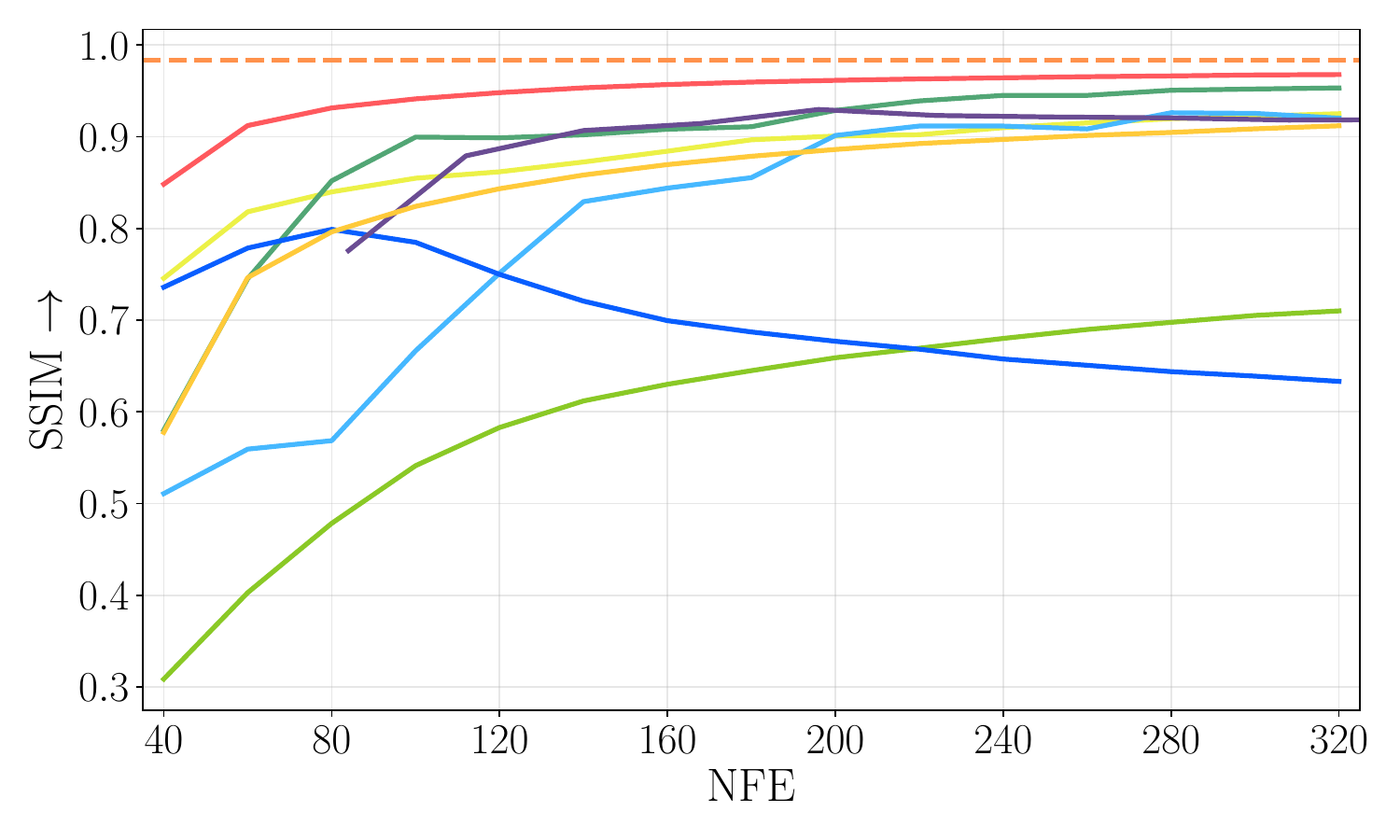}
  \includegraphics[width=.49\textwidth]{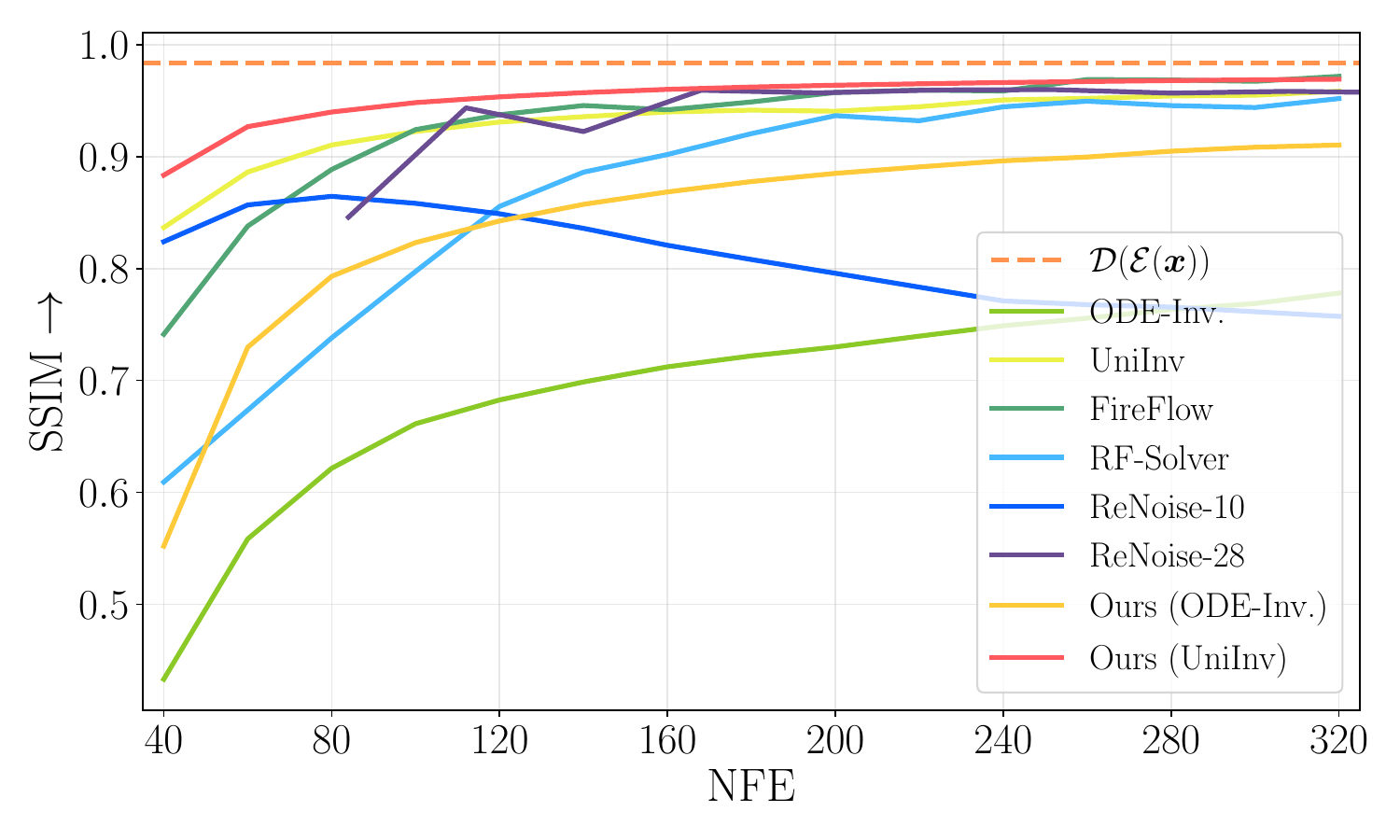}
\end{subfigure}
\begin{subfigure}[t]{\linewidth}
  \centering
  \includegraphics[width=.49\textwidth]{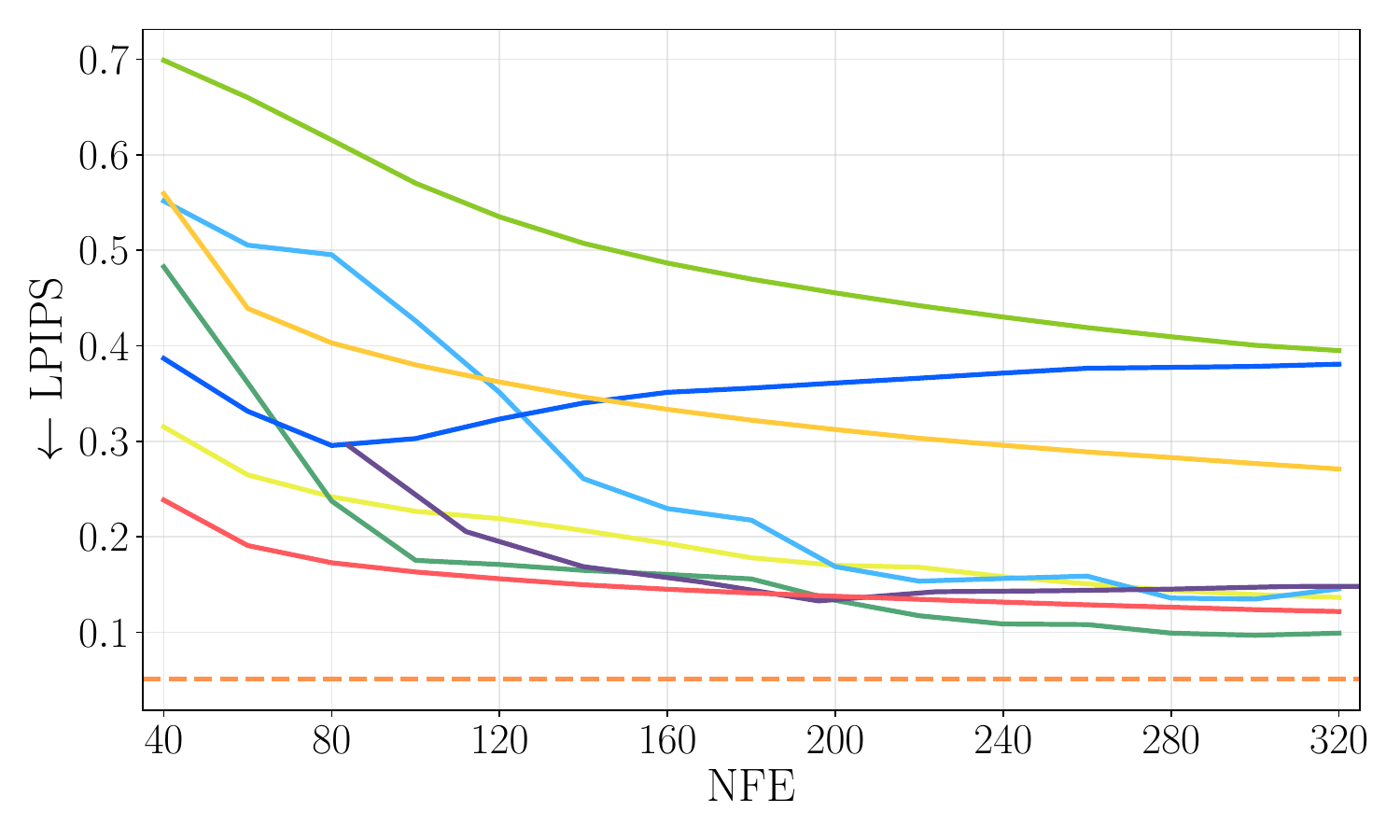}
  \includegraphics[width=.49\textwidth]{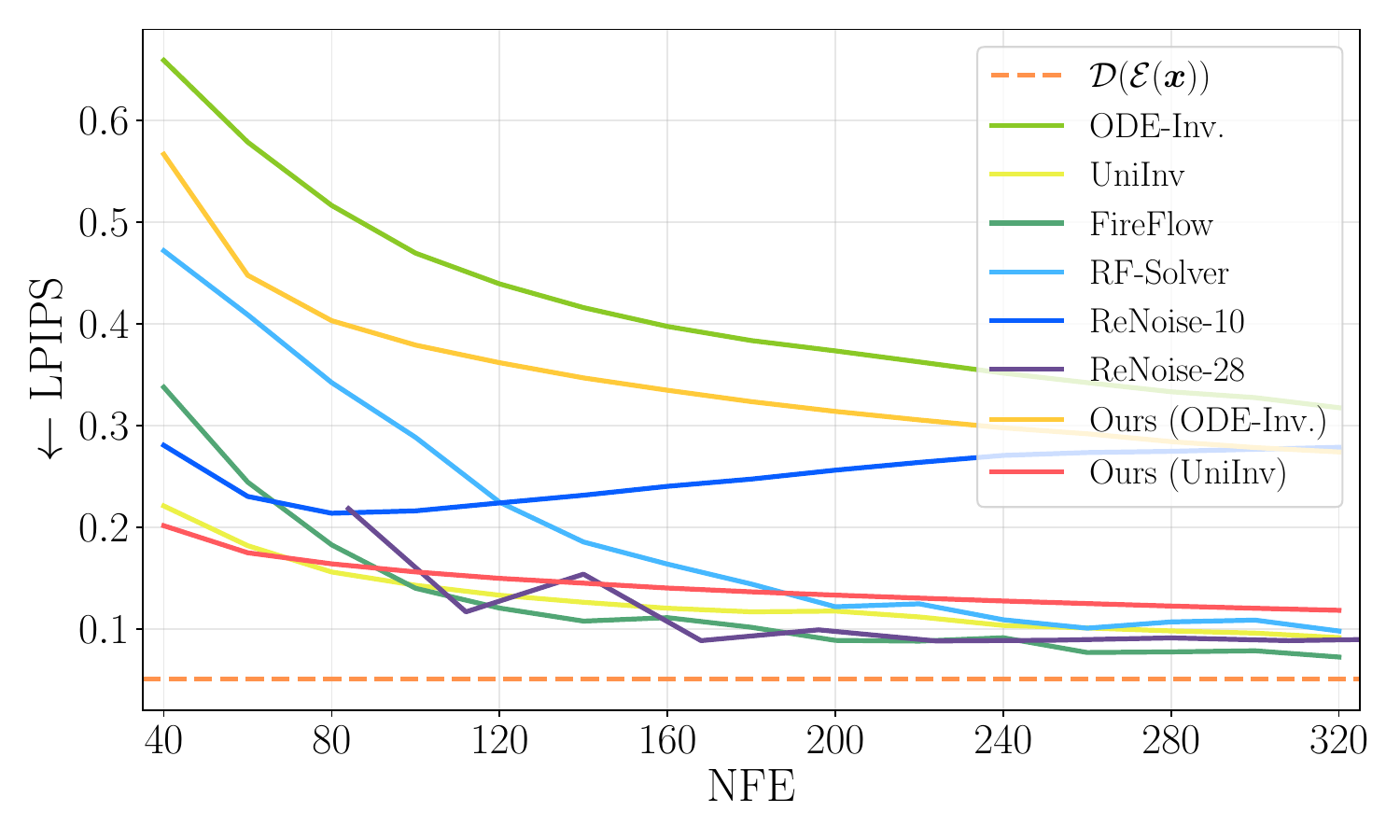}
\end{subfigure}
\caption{\textbf{Reconstruction quantitative comparisons (SD3).} Pixel-space RMSE (first row), PSNR (second row), SSIM (third row), and LPIPS (last row) as functions of the number of NFEs for several inversion methods, for unconditional (left) and text-conditional (right) sampling. The red and yellow curves correspond to our \oursshort{} initialized with the UniInv and ODE Inversion methods, respectively. The dashed orange horizontal line is the average of forwarding the images through the encoder and decoder of the model.}
\label{fig:reconstruction_sd3}
\end{figure}

\paragraph{Quantitative evaluation.}
The reconstruction results of \oursshort{}, as well as competing the methods are provided in Fig.~\ref{fig:reconstruction_sd3}  both for the unconditional and the conditional case. We can see that our method achieves the best reconstruction results for various NFE values, both for unconditional and for conditional sampling. 
We can also see that the initialization affects the results achieved by our method, with UniInv leading to better results than naive ODE Inversion and outperforming the competing methods.

\FloatBarrier
\clearpage

\subsection{Image Editing}
\paragraph{Implementation details.}
We set the number of denoisers to $T = 15$, and evaluate our method for various $n_{\max}$ values. Specifically, we use $n_{\max} \in \{ 13, 12 \}$. We set the CFG to the default value, \ie $\text{CFG} = 3.5$. We evaluate our method for various number of iterations, $N \in \{ 2, 3, 4, 5 \}$. For all figures we present the results obtained with $n_{\max} = 12$.

\paragraph{Dataset.}
We use the same dataset used for evaluating FLUX -- about $400$ text-image pairs. The dataset consists of $90$ real images of dimensions $1024 \times 1024$, which were captioned by LLaVA-1.5, and manually refined. The target prompts for editing the images were handcrafted.

\paragraph{Competing methods.}
We compare our method against ODE Inversion, UniEdit and FlowEdit. As there was no official implementation for UniEdit for SD3, we implemented it by ourselves. For ODE Inversion, we apply the same number of NFEs used for our method. For all methods, we performed hyperparameters search. Additional details regarding the hyperparameters are provided below.

\begin{figure}[ht]
\centering
\begin{subfigure}[t]{\linewidth}
  \centering 
  \includegraphics[width=.327\textwidth]{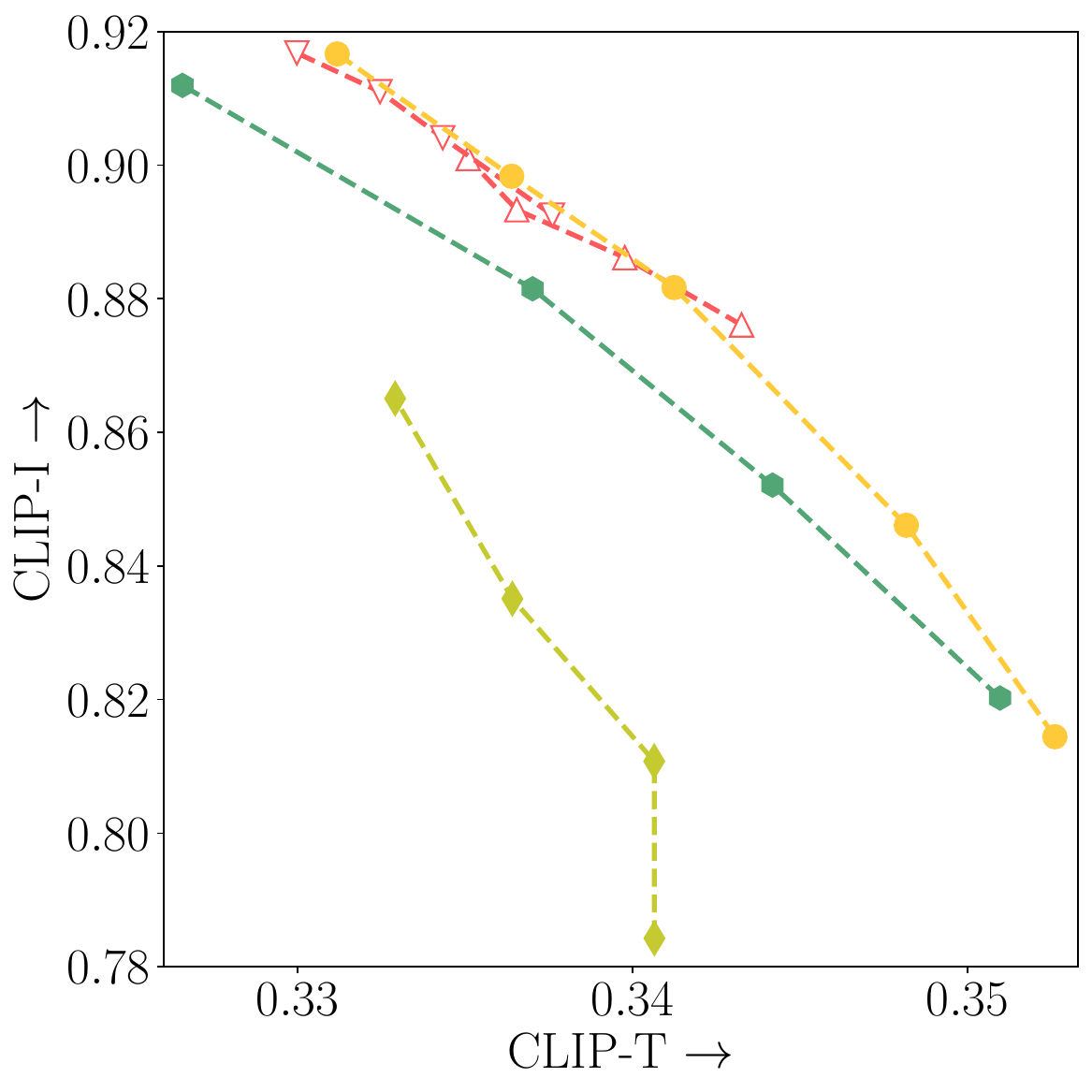}
  \includegraphics[width=.327\textwidth]{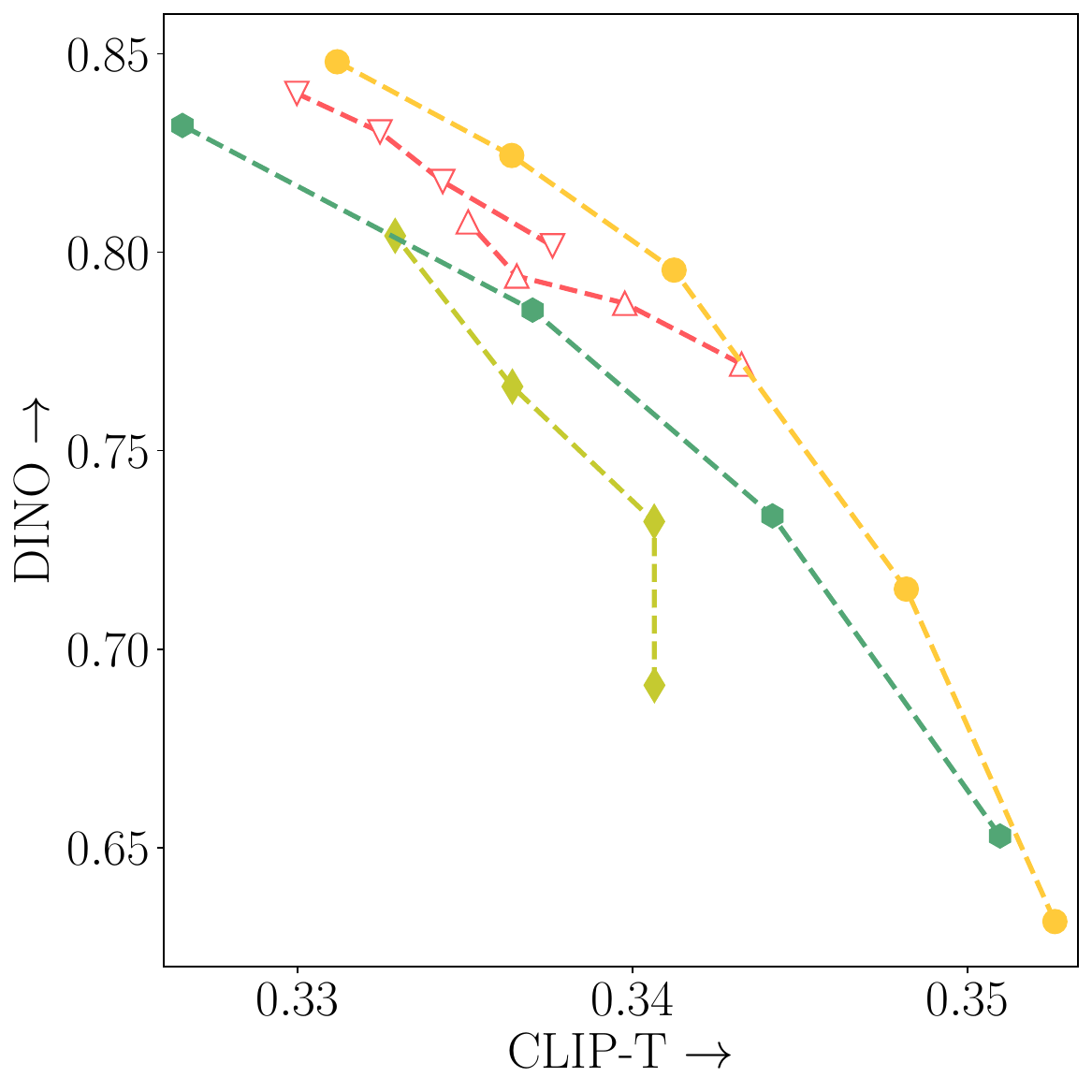}
  \includegraphics[width=.327\textwidth]{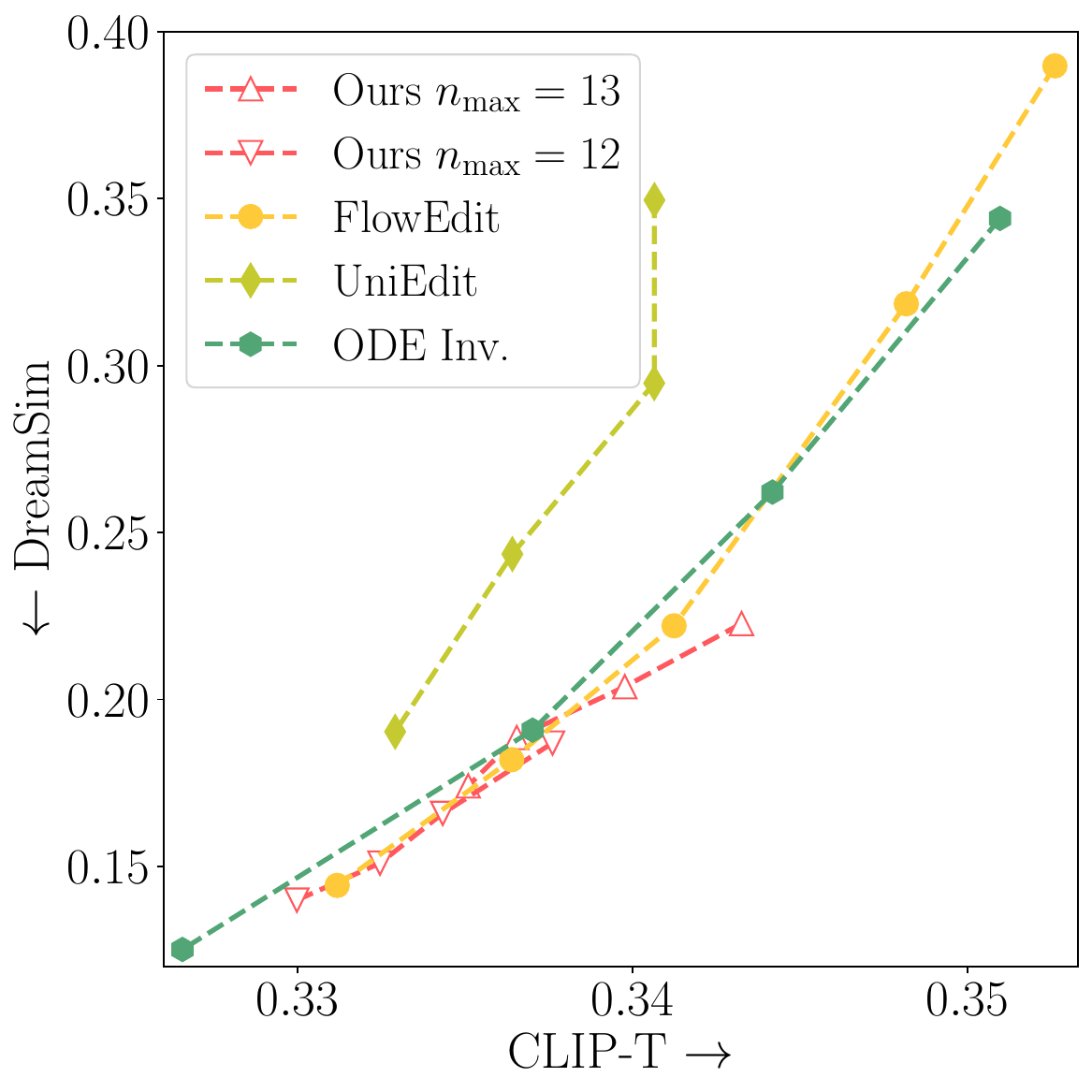}
\end{subfigure}
\caption{\textbf{Editing quantitative comparisons (SD3).}  Text adherence is measured by CLIP-Text (x-axis) for all figures. Image adherence (y-axis) is measured by CLIP-Image (left), DINOv3 (center), and DreamSim (right). Connected markers represent different hyperparameters.}
\label{fig:editing_quantitative_sd3}
\end{figure}

\paragraph{Quantitative evaluation.}
We evaluate the results of all methods using the same measures reported for FLUX in Sec.~\ref{sec:experiments}. The results are presented in Fig.~\ref{fig:editing_quantitative_sd3}.
We see that our method achieves results comparable to FlowEdit, and achieves better results than other competing methods.

\begin{figure}[ht!]
    \centering
    \includegraphics[width=\linewidth]{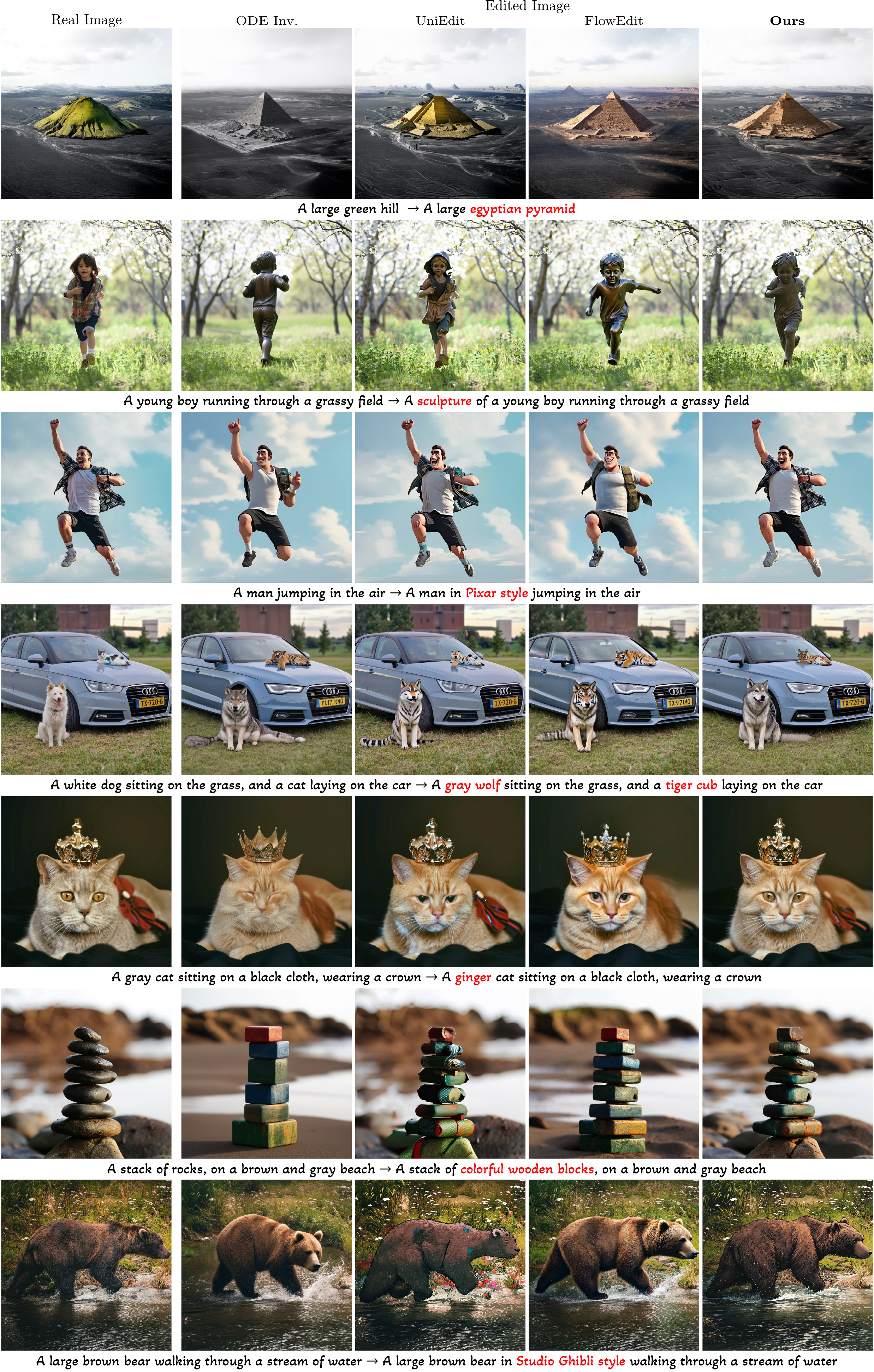}
    \caption{\textbf{Qualitative comparisons (SD3).}  Fine details are visible upon zooming in.}
    \label{fig:supp_qualitative_comparison_sd3}
\end{figure}

\begin{figure}[ht!]
    \centering
    \includegraphics[width=\linewidth]{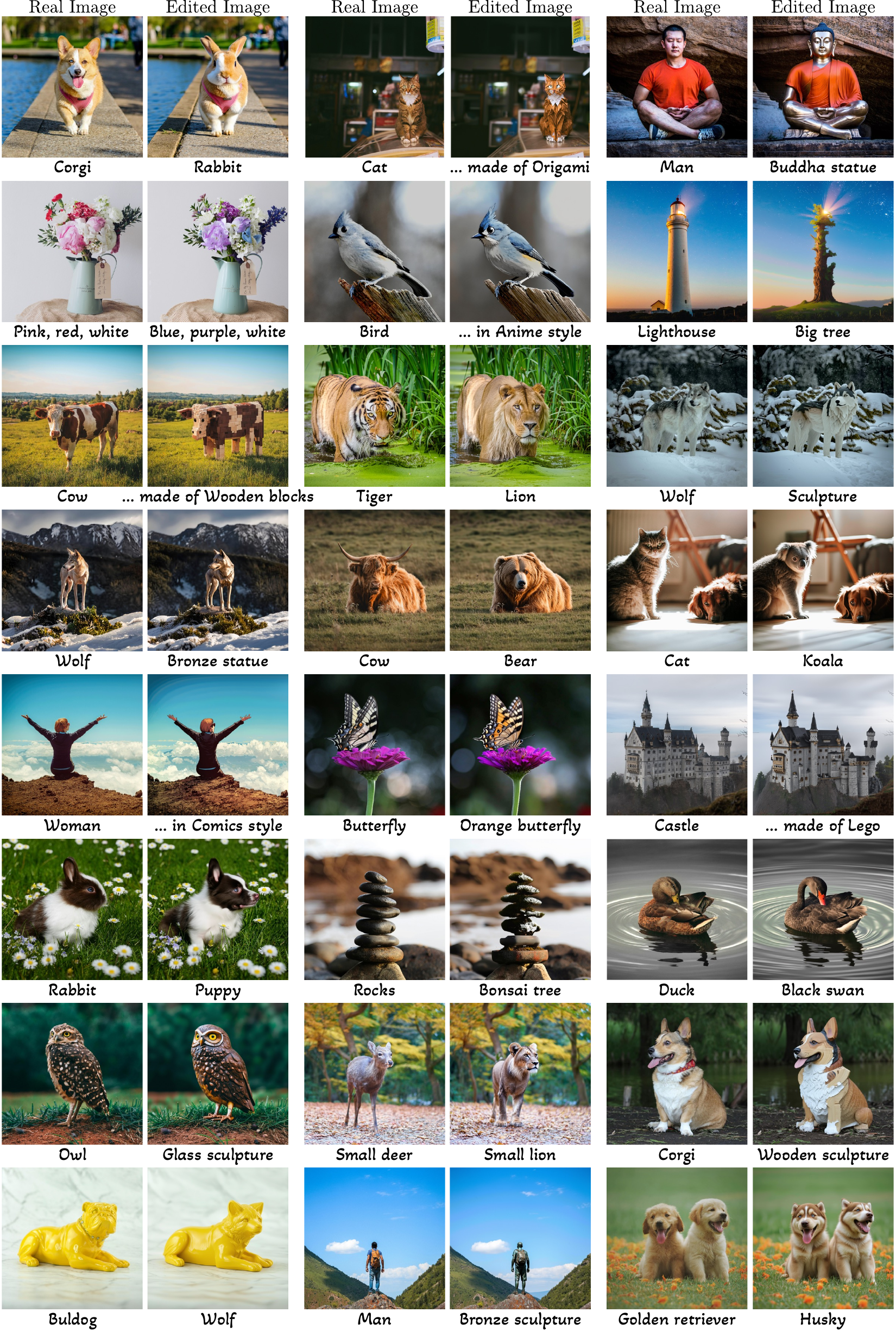}
    \caption{\textbf{Additional \oursshort{} results (SD3).} Fine details are visible upon zooming in.}
    \label{fig:supp_results_sd3}
\end{figure}

\paragraph{Qualitative evaluation.}
Figure \ref{fig:supp_qualitative_comparison_sd3} shows comparisons between \oursshort{} method and the competing methods. More details about the hyperparameters used to construct this figure are provided in Sec.~\ref{subsec:sd3_hyperparameters}. We can see that our method achieves at least comparable results to other methods, for both object editing and style editing. For example, \oursshort{} is the only method that preserves the structure of the scene and the running kid (second row), and successfully turns him into a sculpture. Moreover, our method is the only one that preserves the cat and the crown structure (fifth row), and successfully edits its color. Additional results of our method are provided in Fig.~\ref{fig:supp_results_sd3}.

\FloatBarrier

\subsubsection{SD3 hyperparameters.}
\label{subsec:sd3_hyperparameters}
The hyperparameters presented in Fig.~\ref{fig:supp_qualitative_comparison_sd3} for FlowEdit, ODE Inversion, and \oursshort{} are listed in Tab.~\ref{tab:hyperparams_sd3}, where the chosen hyperparameters for the displayed figures are marked in bold.

\begin{table*}[ht!]
\centering
\caption{\textbf{SD3 hyperparameters.}}
\label{tab:hyperparams_sd3}
\begin{tabular}{@{}ccccccc@{}}
\toprule
 & $T$ & $n_{\max}$ & CFG @ source & CFG @ target & $N$ iterations   \\ \midrule
FlowEdit & $50$ & $45,\;40,\;\textbf{33},\;30,\;27$ & $3.5$ & $13.5$ & -  \\
ODE Inversion & $50$ & $45,\;\textbf{40},\;35,\;30$  & $1$ & $3.5$ & -  \\
\oursshort{} & $15$ & $13,\;\textbf{12}$  & $1$ & $3.5$ & $2,\;3,\;4,\;5$  \\
\bottomrule
\end{tabular}
\end{table*}

For UniEdit, the evaluated hyperparmeters are presented in Tab.~\ref{tab:hyperparams_uniedit_sd3}, with the chosen value for their $\alpha$ parameter marked in bold. In our notation, $\alpha = n_{\max}/ T$.

\begin{table*}[ht!]
\centering
\caption{\textbf{SD3 UniEdit hyperparameters.}}
\label{tab:hyperparams_uniedit_sd3}
\begin{tabular}{@{}ccc@{}}
\toprule
 $T$ & $\alpha$ delay rate & $\omega$ guidance scale  \\ \midrule
 $15$ & $\mathbf{\frac{2}{5}},\;\frac{2}{3},\;\frac{11}{15},\;\frac{3}{5}$ & $5$  \\
\bottomrule
\end{tabular}
\end{table*}

%% file: sections/supplementary/other_losses.tex
\section{Editing with other loss functions}
\label{ap:other_losses}
\begin{figure}[htbp]
    \centering
    \includegraphics[width=\linewidth]{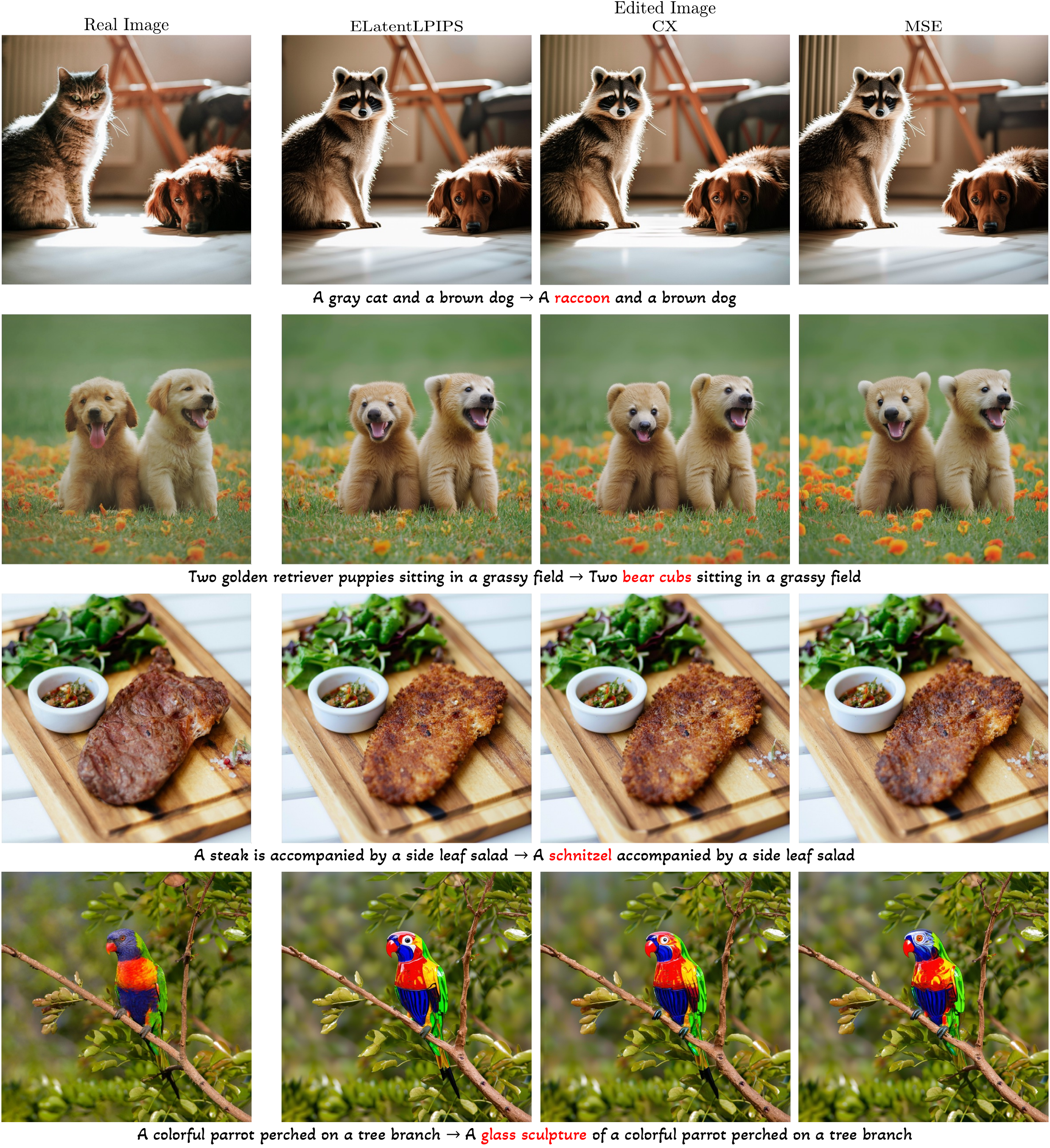}
    \caption{\textbf{Qualitative comparisons using other loss functions (SD3).} The results obtained for the update rule in \eqref{eq:general_loss_update_rule}, for ELatentLPIPS loss (left), contextual (CX) loss (center), and our proposed approach -- MSE loss (right). The results obtained by all losses are similar.}
    \label{fig:supp_loss_comparison_sd3}
\end{figure}

\begin{figure}[htbp]
    \centering
    \includegraphics[width=\linewidth]{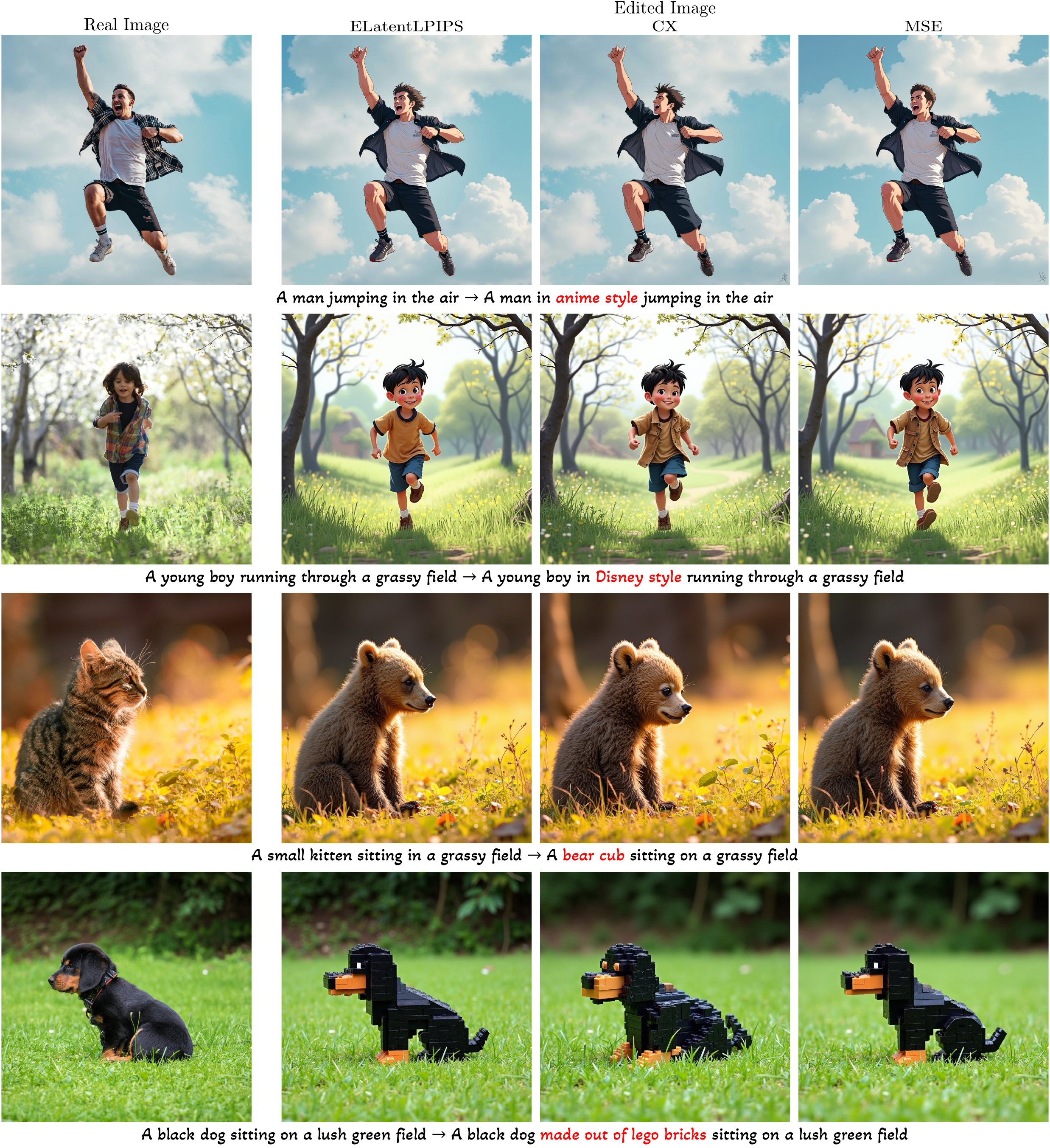}
    \caption{\textbf{Qualitative comparisons using other loss functions (FLUX).} The results obtained for the update rule in \eqref{eq:general_loss_update_rule}, for ELatentLPIPS loss (left), contextual (CX) loss (center), and our proposed approach -- MSE loss (right). The results obtained by all losses are similar.}
    \label{fig:supp_loss_comparison_flux}
\end{figure}

\begin{figure}[htbp]
    \centering
    \includegraphics[width=\linewidth]{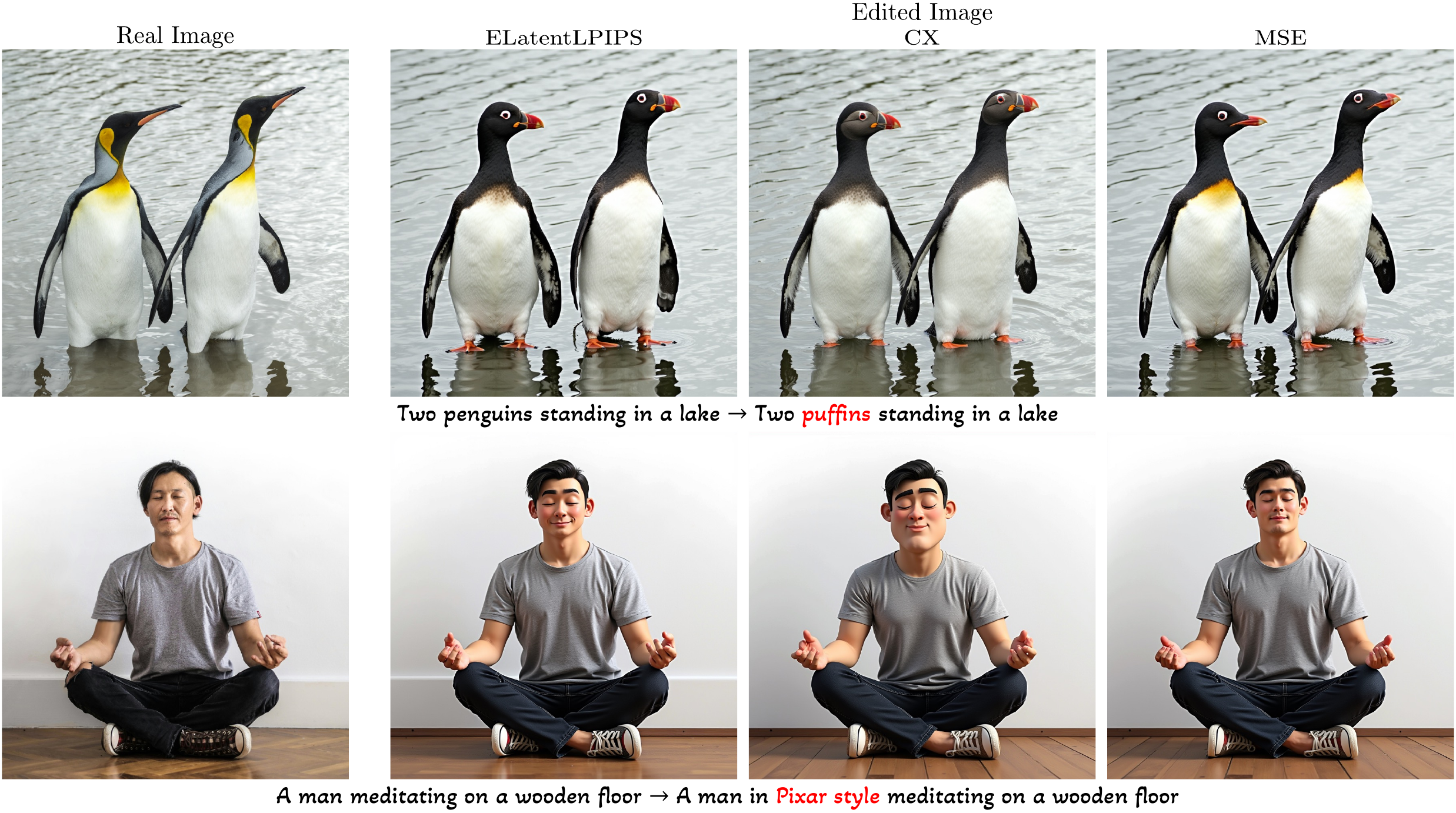}
    \caption{\textbf{Qualitative comparisons using other loss functions (FLUX).} The results obtained for the update rule in \eqref{eq:general_loss_update_rule}, for ELatentLPIPS loss (left), contextual (CX) loss (center), and our proposed approach -- MSE loss (right). Infrequent cases where the results obtained with loss functions other than MSE are better than results obtained with the MSE loss (allowing weaker structure preservation in favor of stronger adherence to the target text prompt). The number of iterations is typically larger for other loss functions.}
    \label{fig:supp_loss_comparison_flux_exception}
\end{figure}

As noted in Sec.~\ref{sec:method}, we can generalize the MSE loss defined in \eqref{eq:mse_loss} to other loss functions $\mathcal{L}(f(\boldsymbol{z}_t), \boldsymbol{y})$. In this case, the update rule in \eqref{eq:mse_update_rule} becomes
\begin{equation}
    \boldsymbol{z}^{(i+1)}_t \gets \boldsymbol{z}^{(i)}_t - \eta \nabla_{f} \mathcal{L}\Bigl( f(\boldsymbol{z}^{(i)}_t), \boldsymbol{y} \Bigr).
\label{eq:general_loss_update_rule}
\end{equation}
Note that \eqref{eq:general_loss_update_rule} uses the gradient of the loss, but not the Jacobian of $f$. That is, it does not require backpropagating through the flow process, though it does require backpropagating through decoder in cases where the loss is defined in pixel-space. However, while seemingly attractive, we have not seen significant advantages for using losses other than the $\normltwo$ loss, except of infrequent cases, as presented in Fig.~\ref{fig:supp_loss_comparison_flux_exception} (for FLUX).
Moreover, we observed that other losses typically achieved satisfying results for larger number of iterations ($N$). Specifically, other losses typically require $\sim 15-30$ iterations, which is significantly more than the $\sim 3 - 5$ iterations that commonly suffice for the $\normltwo$ loss. Therefore, the $\normltwo$ loss has the advantage of achieving satisfying results while being computationally.
Figures \ref{fig:supp_loss_comparison_sd3}, \ref{fig:supp_loss_comparison_flux} present results with different loss functions, for both SD3 and FLUX. These include the contextual loss (CX) \citep{mechrez2018contextual} in pixel space and the ELatentLPIPS \citep{kang2024distilling} in latent space, in addition to our default latent-space $\normltwo$ loss.
For all loss functions, we used the hyperparameters reported in Sec.~\ref{sec:experiments} for FLUX, and in App.~\ref{ap:sd3} for SD3, except for $N$ and $\eta$.
We can see that the CX and ELatentLpips losses achieve similar results to the ones obtained with the $\normltwo$ loss.

\FloatBarrier

%% file: sections/supplementary/contraction_mapping.tex
\section{Contraction mapping}
\label{ap:contraction_mapping}

\subsection{Proof of Theorem \ref{theorem:contraction_mapping}}

    Let $g(\boldsymbol{u}) \triangleq \boldsymbol{u} - \eta ( f(\boldsymbol{u}) - \boldsymbol{y})$. By definition, $g(\boldsymbol{u})$ is a contraction mapping if there exists $\gamma \in [0, 1)$ such that
    \begin{equation}
        \norm{g(\boldsymbol{u}_1) - g(\boldsymbol{u}_2)} \leq \gamma \norm{\boldsymbol{u}_1 - \boldsymbol{u}_2}
    \end{equation}
    for all $\boldsymbol{u}_1, \boldsymbol{u}_2$.
    Substituting $g$, the inequality reads
    \begin{align}
        \norm{\left( \boldsymbol{u}_1 - \eta \left( f(\boldsymbol{u}_1) - \boldsymbol{y} \right) \right) - \left( \boldsymbol{u}_2 - \eta \left( f(\boldsymbol{u}_2) - \boldsymbol{y} \right) \right)} \leq \gamma \norm{\boldsymbol{u}_1 - \boldsymbol{u}_2}.
    \end{align}
    Squaring both sides, we get
    \begin{align}
        \norm{\boldsymbol{u}_1 - \boldsymbol{u}_2 - \eta \left( f(\boldsymbol{u}_1) - f(\boldsymbol{u}_2) \right)}^2 \leq \gamma^2 \norm{\boldsymbol{u}_1 - \boldsymbol{u}_2}^2.
    \end{align}
    Rearranging terms, we get
    \begin{align}
        \norm{\boldsymbol{u}_1 - \boldsymbol{u}_2}^2 + \eta^2 \norm{f(\boldsymbol{u}_1) - f(\boldsymbol{u}_2)}^2 - 2 \eta \left\langle \boldsymbol{u}_1 - \boldsymbol{u}_2, f(\boldsymbol{u}_1) - f(\boldsymbol{u}_2) \right\rangle \leq \gamma^2 \norm{\boldsymbol{u}_1 - \boldsymbol{u}_2}^2.
    \end{align}
    Defining $\kappa = 1 - \gamma^2 \in (0, 1]$, we get a quadratic inequality in $\eta$, 
    \begin{align}
         \norm{f(\boldsymbol{u}_1) - f(\boldsymbol{u}_2)}^2 \eta^2 - 2 \left\langle \boldsymbol{u}_1 - \boldsymbol{u}_2, f(\boldsymbol{u}_1) - f(\boldsymbol{u}_2) \right\rangle \eta + \kappa \norm{\boldsymbol{u}_1 - \boldsymbol{u}_2}^2 \leq 0.
    \end{align}
    For each given pair of $\boldsymbol{u}_1, \boldsymbol{u}_2$, the set of $\eta$'s that satisfy the inequality is $\eta \in \left[ \eta_1(\boldsymbol{u}_1, \boldsymbol{u}_2), \eta_2(\boldsymbol{u}_1, \boldsymbol{u}_2) \right]$, where
    \begin{align}
        \eta_{1,2}(\boldsymbol{u}_1, \boldsymbol{u}_2) &= \frac{\left\langle \boldsymbol{u}_1 - \boldsymbol{u}_2, f(\boldsymbol{u}_1) - f(\boldsymbol{u}_2) \right\rangle}{\norm{f(\boldsymbol{u}_1) - f(\boldsymbol{u}_2)}^2} \pm \sqrt{\left(\frac{\left\langle \boldsymbol{u}_1 - \boldsymbol{u}_2, f(\boldsymbol{u}_1) - f(\boldsymbol{u}_2) \right\rangle}{\norm{f(\boldsymbol{u}_1) - f(\boldsymbol{u}_2)}^2}\right)^2 - \kappa \left(\frac{\norm{\boldsymbol{u}_1 - \boldsymbol{u}_2}}{\norm{f(\boldsymbol{u}_1) - f(\boldsymbol{u}_2)}}\right)^2} \nonumber\\
        &= \frac{\left\langle \boldsymbol{u}_1 - \boldsymbol{u}_2, f(\boldsymbol{u}_1) - f(\boldsymbol{u}_2) \right\rangle}{\norm{f(\boldsymbol{u}_1) - f(\boldsymbol{u}_2)}^2} \left( 1 \pm \sqrt{1 - \kappa \left( \frac{\norm{f(\boldsymbol{u}_1) - f(\boldsymbol{u}_2)}\norm{\boldsymbol{u}_1 -\boldsymbol{u}_2}}{\left\langle \boldsymbol{u}_1 - \boldsymbol{u}_2, f(\boldsymbol{u}_1) - f(\boldsymbol{u}_2) \right\rangle} \right)^2} \right).
    \end{align}
    Therefore, if we choose
    \begin{equation}\label{eq:eta_bar_sup_inf}
     \eta \in (\bar\eta_1, \bar\eta_2) \subset \left[\sup_{\boldsymbol{u}_1, \boldsymbol{u}_2} \eta_1(\boldsymbol{u}_1, \boldsymbol{u}_2), \inf_{\boldsymbol{u}_1, \boldsymbol{u}_2} \eta_2(\boldsymbol{u}_1, \boldsymbol{u}_2)\right],  
    \end{equation}
    then the iterations are guaranteed to converge. To choose $\bar\eta_2$, we note that 
    \begin{align}
        &\inf_{\boldsymbol{u}_1, \boldsymbol{u}_2} \eta_2(\boldsymbol{u}_1, \boldsymbol{u}_2) \nonumber\\
        &\geq \inf_{\boldsymbol{u}_1,\boldsymbol{u}_2} \frac{\left\langle \boldsymbol{u}_1 - \boldsymbol{u}_2, f(\boldsymbol{u}_1) - f(\boldsymbol{u}_2) \right\rangle}{\norm{f(\boldsymbol{u}_1) - f(\boldsymbol{u}_2)}^2}  \inf_{\boldsymbol{u}_1, \boldsymbol{u}_2} \left(1+ \sqrt{1 - \kappa \left( \frac{\norm{f(\boldsymbol{u}_1) - f(\boldsymbol{u}_2)} \norm{\boldsymbol{u}_1 - \boldsymbol{u}_2}}{\left\langle \boldsymbol{u}_1 - \boldsymbol{u}_2, f(\boldsymbol{u}_1) - f(\boldsymbol{u}_2) \right\rangle} \right)^2} \right) \nonumber\\
        &= \inf_{\boldsymbol{u}_1, \boldsymbol{u}_2} \frac{\left\langle \boldsymbol{u}_1 - \boldsymbol{u}_2, f(\boldsymbol{u}_1) - f(\boldsymbol{u}_2) \right\rangle}{\norm{f(\boldsymbol{u}_1) - f(\boldsymbol{u}_2)}^2} \left( 1 + \sqrt{1 - \kappa \sup_{\boldsymbol{u}_1, \boldsymbol{u}_2} \left( \frac{\norm{f(\boldsymbol{u}_1) - f(\boldsymbol{u}_2)} \norm{\boldsymbol{u}_1 -\boldsymbol{u}_2}}{\left\langle \boldsymbol{u}_1 - \boldsymbol{u}_2, f(\boldsymbol{u}_1) - f(\boldsymbol{u}_2) \right\rangle} \right)^2} \right) \nonumber\\
        & \geq \inf_{\boldsymbol{u}_1, \boldsymbol{u}_2} \frac{\left\langle \boldsymbol{u}_1 - \boldsymbol{u}_2, f(\boldsymbol{u}_1) - f(\boldsymbol{u}_2) \right\rangle}{\norm{f(\boldsymbol{u}_1) - f(\boldsymbol{u}_2)}^2} \left( 1 + \sqrt{1 - \frac{\kappa}{\beta^2}} \right) \nonumber\\
        & \triangleq \bar\eta_2,
    \end{align}
    where we denoted $\beta = \inf_{\boldsymbol{u}_1 \neq \boldsymbol{u}_2} \frac{ \left\langle \boldsymbol{u}_1 - \boldsymbol{u}_2, f(\boldsymbol{u}_1) - f(\boldsymbol{u}_2) \right\rangle }{\norm{\boldsymbol{u}_1 - \boldsymbol{u}_2} \norm{f(\boldsymbol{u}_1) - f(\boldsymbol{u}_2)}}$ and used the assumption of the theorem that $\beta > 0$. Note that the first inequality here follows from the fact that both multiplicands are nonnegative, as $\inf_{\boldsymbol{u}_1 \neq \boldsymbol{u}_2}\frac{ \left\langle \boldsymbol{u}_1 - \boldsymbol{u}_2, f(\boldsymbol{u}_1) - f(\boldsymbol{u}_2) \right\rangle }{\norm{f(\boldsymbol{u}_1) - f(\boldsymbol{u}_2)}^2} > 0$ from the assumption of \eqref{eq:contraction_mapping_sufficient} in the theorem. In a similar manner, we can choose $\bar\eta_1$ by noting that
    \begin{align}
    \sup_{\boldsymbol{u}_1, \boldsymbol{u}_2} \eta_1(\boldsymbol{u}_1,\boldsymbol{u}_2) \leq \sup_{\boldsymbol{u}_1, \boldsymbol{u}_2} \frac{\left\langle \boldsymbol{u}_1 - \boldsymbol{u}_2, f(\boldsymbol{u}_1) - f(\boldsymbol{u}_2) \right\rangle}{\norm{f(\boldsymbol{u}_1) - f(\boldsymbol{u}_2)}^2} \left( 1 - \sqrt{1 - \frac{\kappa}{\beta^2}}\right) \triangleq \bar\eta_1.
    \end{align}
    Now, since $\kappa > 0$ can be chosen arbitrarily small, we take the upper bound to be 
    \begin{equation}
        \lim_{\kappa \to 0} \bar\eta_2 = 2 \inf_{\boldsymbol{u}_1, \boldsymbol{u}_2} \frac{\left\langle \boldsymbol{u}_1 - \boldsymbol{u}_2, f(\boldsymbol{u}_1) - f(\boldsymbol{u}_2) \right\rangle}{\norm{f(\boldsymbol{u}_1) - f(\boldsymbol{u}_2)}^2},
    \label{eq:inner_product_sufficient}
    \end{equation}
    and the lower bound to be
    \begin{equation}
        \lim_{\kappa \to 0} \bar\eta_1 = 0. 
    \end{equation}
    This is allowed since for any $\eta \in ( \lim_{\kappa \to 0} \bar\eta_1, \lim_{\kappa \to 0} \bar\eta_2)$, there exists a fixed $\kappa > 0$ small enough such that \eqref{eq:eta_bar_sup_inf} is satisfied with that particular $\kappa$. This completes the proof of the theorem.

\subsection{Step size upper bound}
To verify our choice of $\eta$, we drew many pairs of samples $\boldsymbol{u}_1, \boldsymbol{u}_2$, as we detail next. Specifically, we generated nonidentical $2000$ text prompts using ChatGPT4 \citep{achiam2023gpt}, drew two different random white Gaussian noises $\boldsymbol{u}_1, \boldsymbol{\varepsilon}$ for each text prompt, and defined $\boldsymbol{u}_2$ as
\begin{equation}
    \boldsymbol{u}_2 = \sqrt{\alpha} \boldsymbol{u}_1 + \sqrt{1 - \alpha} \boldsymbol{\varepsilon},
\end{equation}
for various $\alpha$ values, so that both $\boldsymbol{u}_1$ and $\boldsymbol{u}_2$ are distributed $\sim \mathcal{N}(\boldsymbol{0}, \boldsymbol{I})$ (an isotropic Gaussian).

\begin{figure}[htbp]
    \centering
    \begin{subfigure}[t]{\linewidth}
      \centering 
      \includegraphics[width=.49\textwidth]{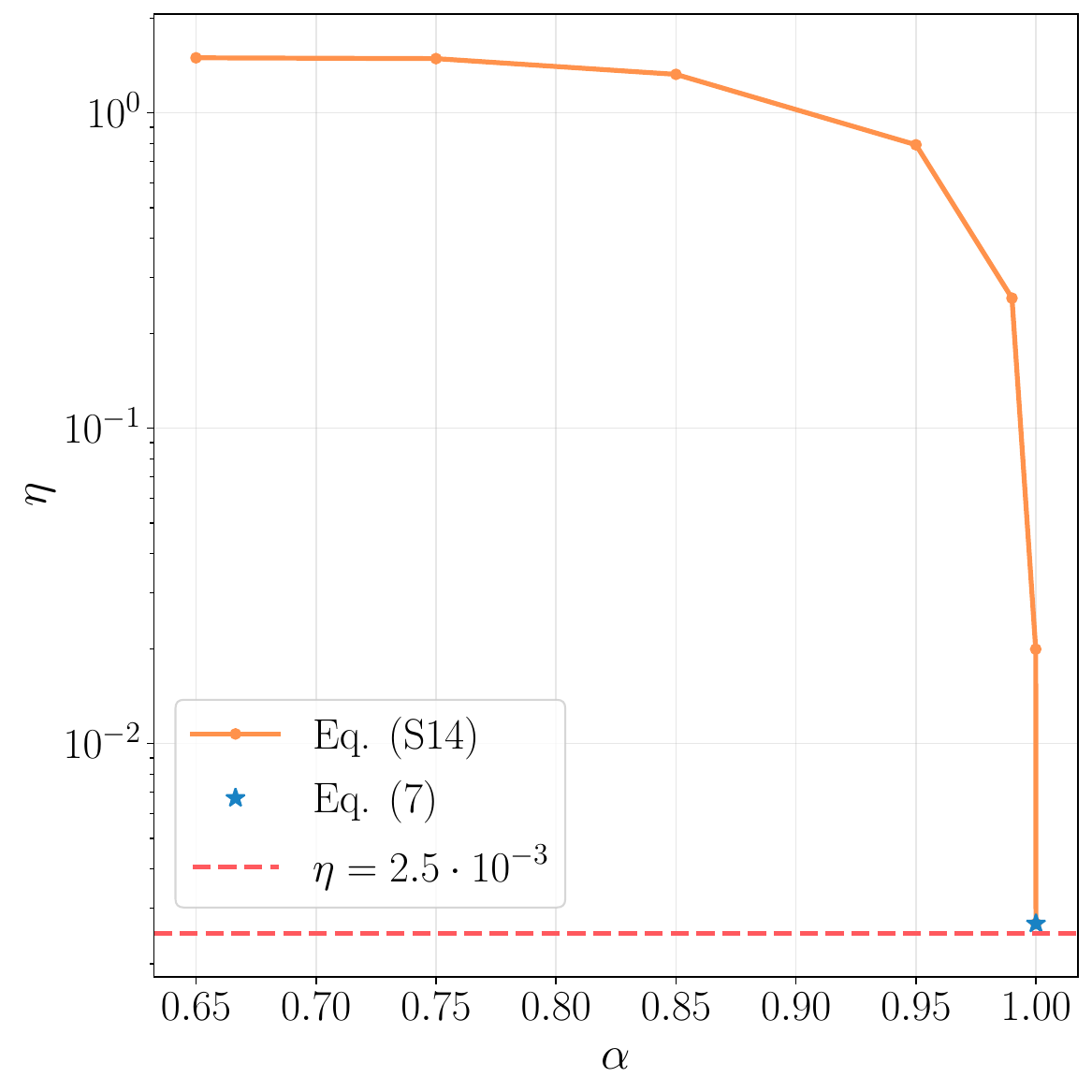}
      \includegraphics[width=.49\textwidth]{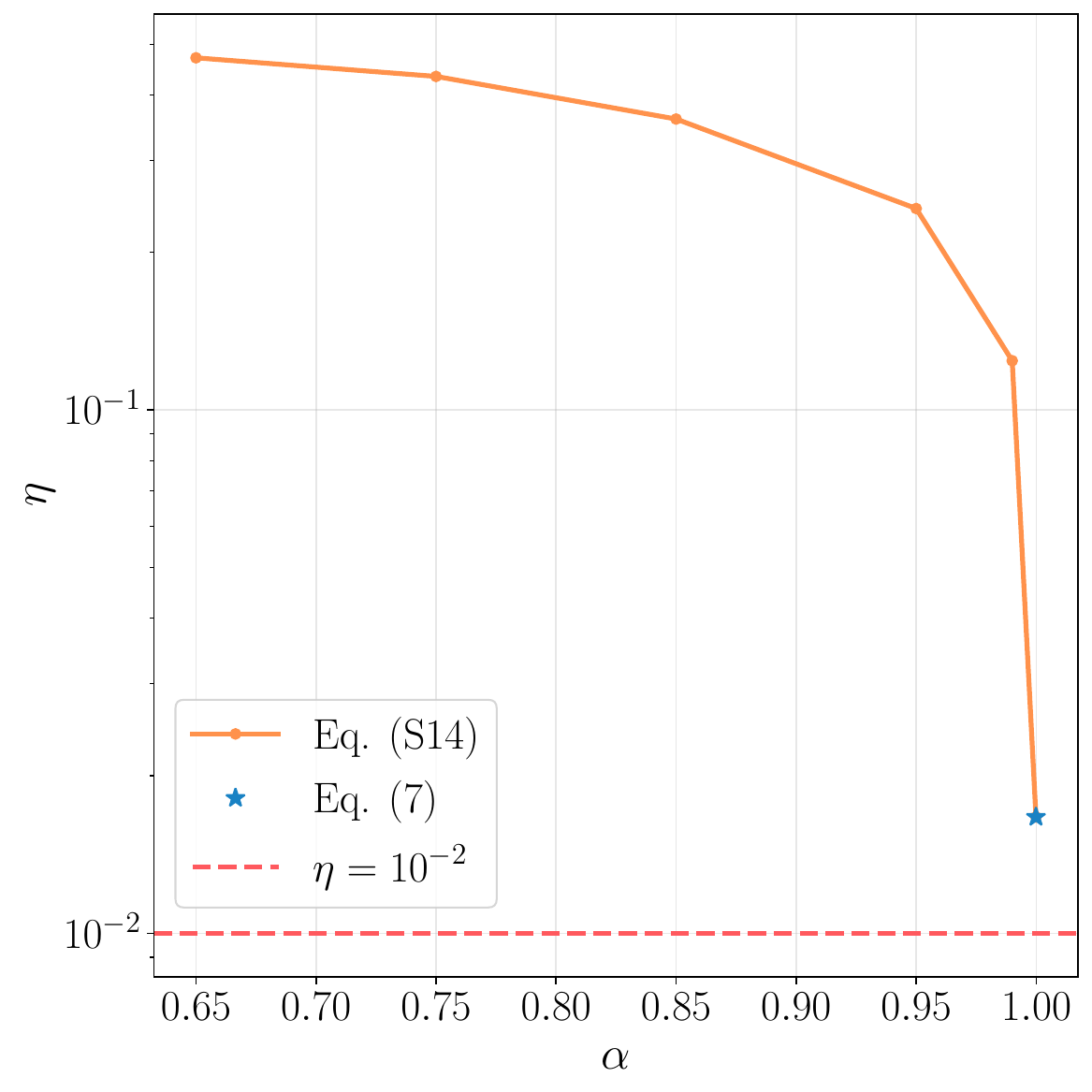}
    \end{subfigure}
    
    \caption{\textbf{Step size upper bounds}. The orange line is the minimum obtained over $2000$ noise realizations in \eqref{eq:sufficient_condition_alpha}, achieved for various $\alpha$ values. The approximation for the upper bound (\eqref{eq:contraction_mapping_sufficient}) is the starred blue point, and the dashed red line is our step size ($\eta$) choice (which is below the upper bound), for FLUX (left) and SD3 (right).}
    \label{fig:supp_lip_inner_prod}
\end{figure}

By substituting $\boldsymbol{u}_2$ into \eqref{eq:inner_product_sufficient} we obtain, for each $\alpha$ value
\begin{equation}
    \min_{\boldsymbol{u}_1, \boldsymbol{\varepsilon}}{2 \frac{\Bigl\langle \left( 1 - \sqrt{\alpha} \right) \boldsymbol{u}_1 - \sqrt{1 - \alpha} \boldsymbol{\varepsilon}, f(\boldsymbol{u}_1) - f(\sqrt{\alpha} \boldsymbol{u}_1 + \sqrt{1 - \alpha} \boldsymbol{\varepsilon}) \Bigr\rangle}{\norm{f(\boldsymbol{u}_1) - f(\sqrt{\alpha} \boldsymbol{u}_1 + \sqrt{1 - \alpha} \boldsymbol{\varepsilon})}^2}},
\label{eq:sufficient_condition_alpha}
\end{equation}
Figure \ref{fig:supp_lip_inner_prod} presents the minimum in \eqref{eq:sufficient_condition_alpha} obtained over the $2000$ different realizations, as a function of $\alpha$, for both FLUX and SD3.
The global minimum, marked by a blue star, is our approximation for the upper bound of \eqref{eq:contraction_mapping_sufficient}. We can see that the minimum is obtained when $\norm{\boldsymbol{u}_1 - \boldsymbol{u}_2}$ is small ($\alpha$ close to $1$).
Our choice for $\eta$, which is presented as a dashed red line, is below this upper bound.

\begin{figure}[htbp]
  \centering
  \includegraphics[width=.6\textwidth]{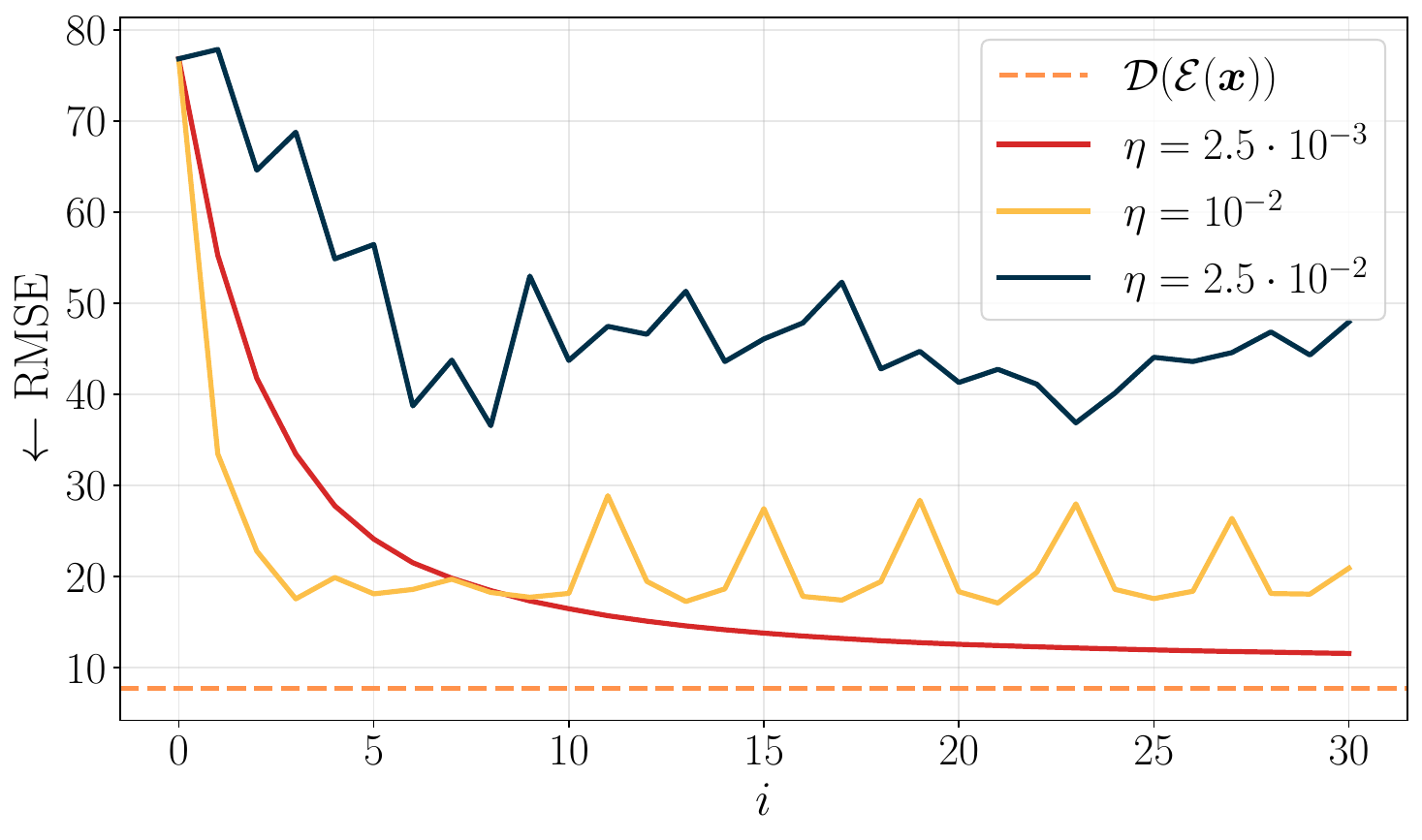}
\caption{\textbf{Convergence analysis (FLUX).} The plot shows RMSE in pixel space vs.~number of iterations for the task of inversion, averaged over a dataset. The step size we use (red) satisfies the sufficient condition of \eqref{eq:contraction_mapping_sufficient} and thus leads to convergence. Step sizes that are $4 \times$ and $10 \times$ larger (yellow and black) do not satisfy the condition and do not lead to convergence. The dashed orange line is the minimal RMSE achievable in this setting. It corresponds to passing images through the encoder and decoder.}
\label{fig:eta_comparisons_flux}
\end{figure}

Figure \ref{fig:eta_comparisons_flux} is the same as Fig.~\ref{fig:eta_comparisons_sd3}, but for FLUX instead of SD3, and the comparisons are to step sizes that are $4 \times$ and $10 \times$ larger than our choice, namely $\eta \in \{ 1.0 \times 10^{-2}, 2.5 \cdot 10^{-2} \}$.

We note that this experiment was evaluated for $T = 10$ both for SD3 and FLUX, and for latent variables $\{ \boldsymbol{z}_t \}$ corresponding to images with dimensions of $1024 \times 1024$. For a different number of denoisers, or images of other resolutions, the experiment should be redone.

\FloatBarrier

%% file: sections/supplementary/ddim_coefficients.tex
\section{Probability flow ODE coefficients}
\label{ap:pf_coeffs}
Each denoising step by the flow formulation is given by
\begin{equation}
    \boldsymbol{z}_{t + \Delta t} = \boldsymbol{z}_t + \boldsymbol{v}_t(\boldsymbol{z}_t) \Delta t,
\label{eq:flow_step}
\end{equation}
where for notational convenience we omit the condition $c$.

However, each denoising step by the DDIM formulation is given by
\begin{equation}
    \boldsymbol{z}_{t-1} = \sqrt{\alpha_{t-1}} \biggl( \frac{\boldsymbol{z}_t - \sqrt{1 - \alpha_t} \epsilon_\theta^t(\boldsymbol{z}_t)}{\sqrt{\alpha_t}} \biggr) + \sqrt{1 - \alpha_{t-1}}\epsilon_\theta^t(\boldsymbol{z}_t),
\label{eq:ddim_step}
\end{equation}
where $\alpha_t$ are the diffusion coefficients as defined by \citet{song2021denoising}, and $\epsilon_\theta^t(\boldsymbol{z}_t)$ is the predicted noise for the current observation $\boldsymbol{z}_t$, replacing the learned vector field $\boldsymbol{v}_t(\boldsymbol{z}_t)$ of the flow formulation. Rearranging \eqref{eq:ddim_step}, we get
\begin{equation}
    \boldsymbol{z}_{t-1} = \frac{\sqrt{\alpha_{t-1}}}{\sqrt{\alpha_t}} \boldsymbol{z}_t + \biggr( \sqrt{1 - \alpha_{t-1}} - \frac{\sqrt{1 - \alpha_t}}{\sqrt{\alpha_t}} \biggl) \epsilon_\theta^t(\boldsymbol{z}_t).
\label{eq:arrange_ddim_step}
\end{equation}
As we relate to the entire process as a black box, and a $\texttt{stop-grad}$ operator is applied on the output of each of the noise-predicting networks, the terms $\epsilon_\theta^t(\boldsymbol{z}_t)$ vanish under differentiation. Stacking all timesteps one after the other, the formulation remains the same as flows, but with a multiplicative coefficient that corresponds to the product of the coefficients multiplying  $\boldsymbol{z}_t$ in each of the timesteps, 
\begin{equation}
    \delta \triangleq \prod_{t=1}^T{\frac{\sqrt{\alpha_{t-1}}}{\sqrt{\alpha_t}}} = \sqrt{\frac{\alpha_0}{\alpha_T}} = \frac{1}{\sqrt{\alpha_T}}.
\label{eq:ddim_multiplicative_coeff}
\end{equation}
Therefore, for example, the update rule for the $\normltwo$ loss in \eqref{eq:mse_loss} for any condition $c$, is given by
\begin{equation}
    \boldsymbol{z}^{(i+1)}_t \gets \boldsymbol{z}^{(i)}_t - \eta \delta \Bigl( f(\boldsymbol{z}^{(i)}_t, c) - \boldsymbol{y} \Bigr).
\end{equation}

%% file: sections/supplementary/teaser_hyperparameters.tex
\section{Hyperparameters used for Figure \ref{fig:teaser}}
\label{ap:teaser_hyperparameters}
The results presented in Fig.~\ref{fig:teaser} were achieved by the hyperparameters provided in Tab.~\ref{tab:teaser_hyperparameters}.

\begin{table*}[ht!]
\centering
\caption{\textbf{Figure \ref{fig:teaser} hyperparameters.}}
\label{tab:teaser_hyperparameters}
\begin{tabular}{@{}cccc@{}}
\toprule
 & Model & $n_{\max}$ & $N$ iterations   \\ \midrule
Owls $\rightarrow$ Cardboard & FLUX & $11$ & $5$  \\
Corgi $\rightarrow$ Lego & FLUX & $13$ & $8$  \\
Forest $\rightarrow$ Paved pathway & FLUX & $13$ & $3$  \\

Penguins $\rightarrow$ Glass sculpture & SD3 & $12$ & $4$  \\
Owl $\rightarrow$ in Anime style & SD3 & $12$ & $5$  \\
Wolf $\rightarrow$ Deer & SD3 & $12$ & $4$  \\

Cow $\rightarrow$ Colorful toy bricks & FLUX & $12$ & $6$  \\
Lizard $\rightarrow$ Crochet & FLUX & $12$ & $5$  \\
Corgi $\rightarrow$ in Pixar style & FLUX & $11$ & $5$  \\

\bottomrule
\end{tabular}
\end{table*}

\clearpage

%% file: sections/supplementary/editing_by_inversion.tex
\section{Editing by inversion}
\label{ap:editing_by_inversion}
\begin{figure}[htbp]
\centering
\begin{subfigure}[t]{\linewidth}
  \centering 
  \includegraphics[width=.327\textwidth]{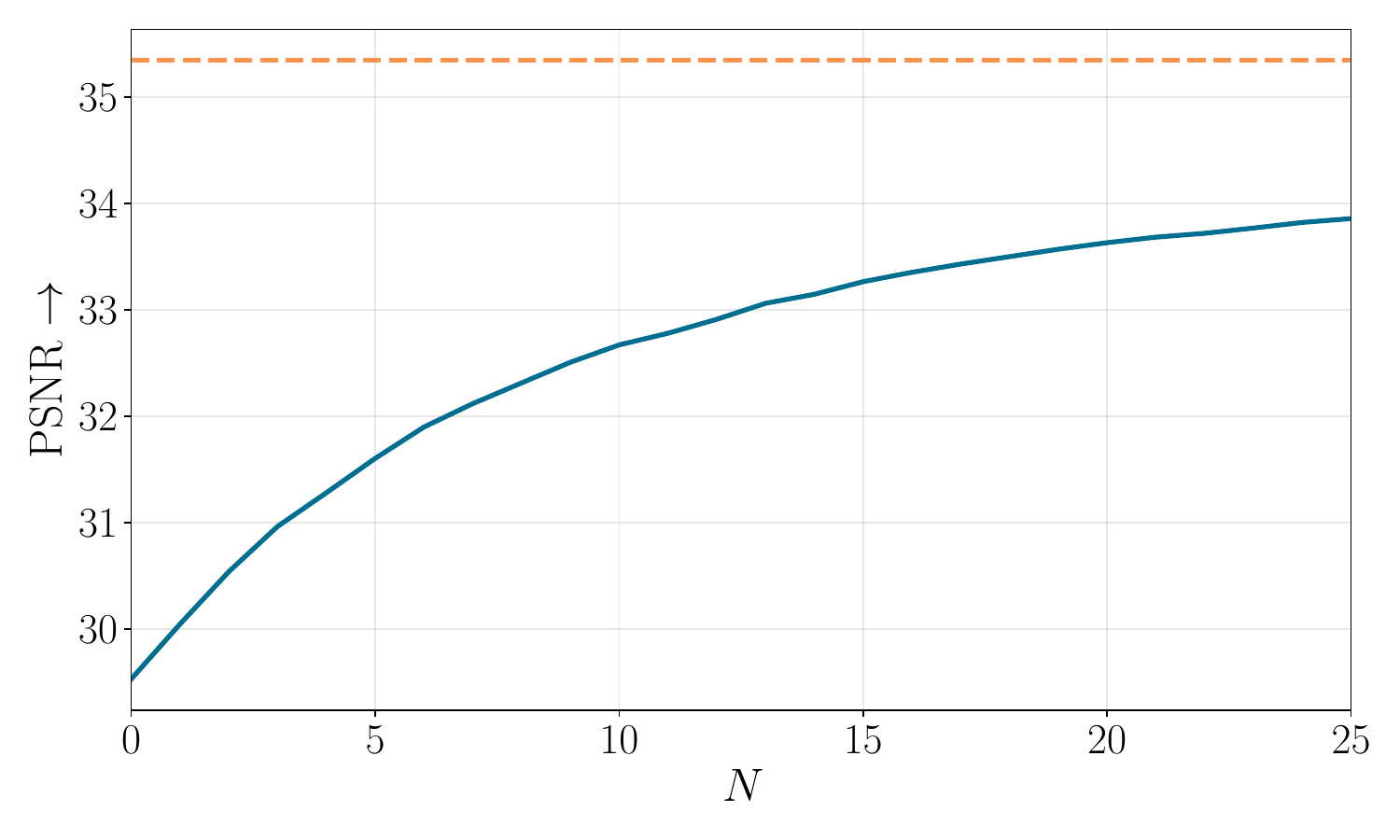}
  \includegraphics[width=.327\textwidth]{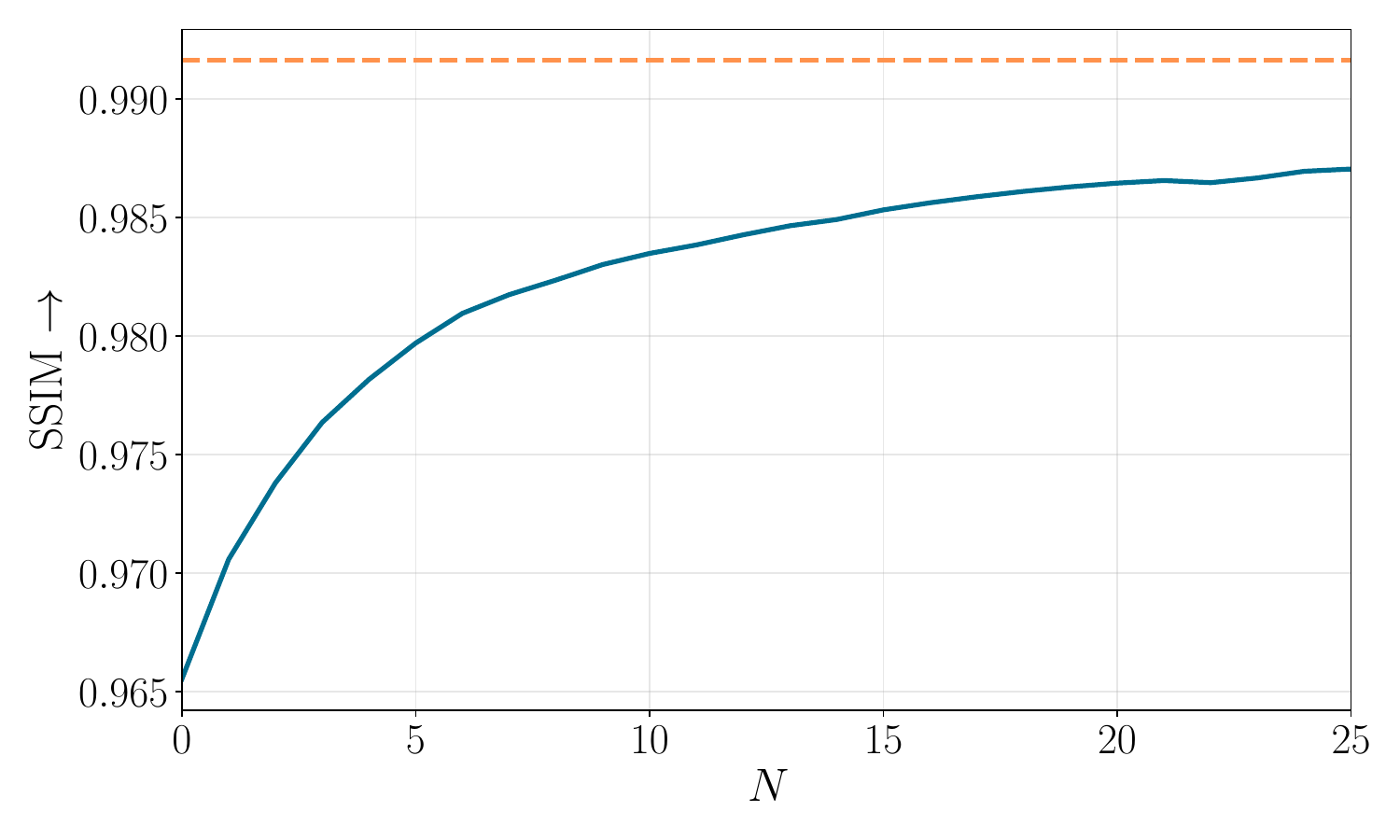}
  \includegraphics[width=.327\textwidth]{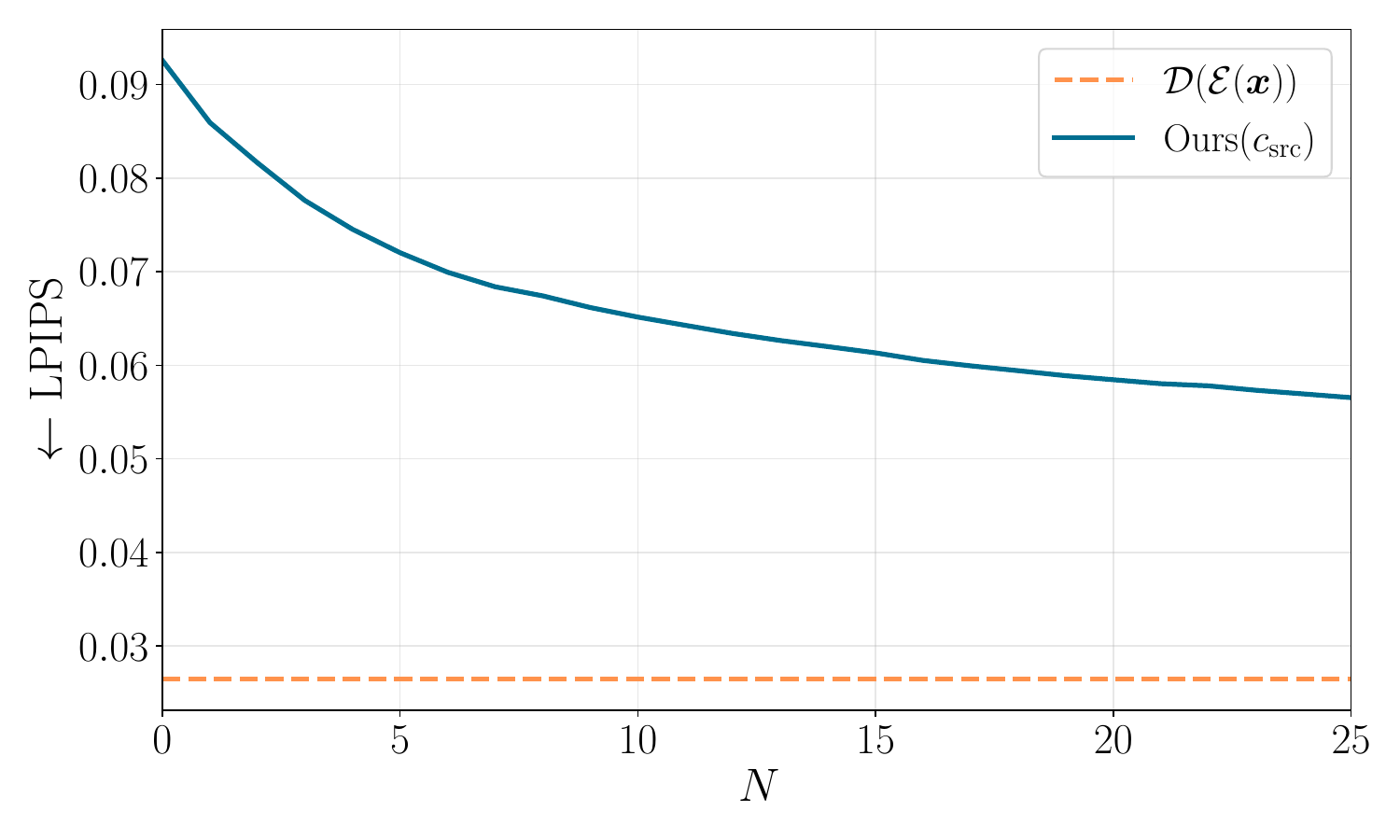}
\end{subfigure}
\hfill
\begin{subfigure}[t]{\linewidth}
  \centering 
  \includegraphics[width=.49\textwidth]{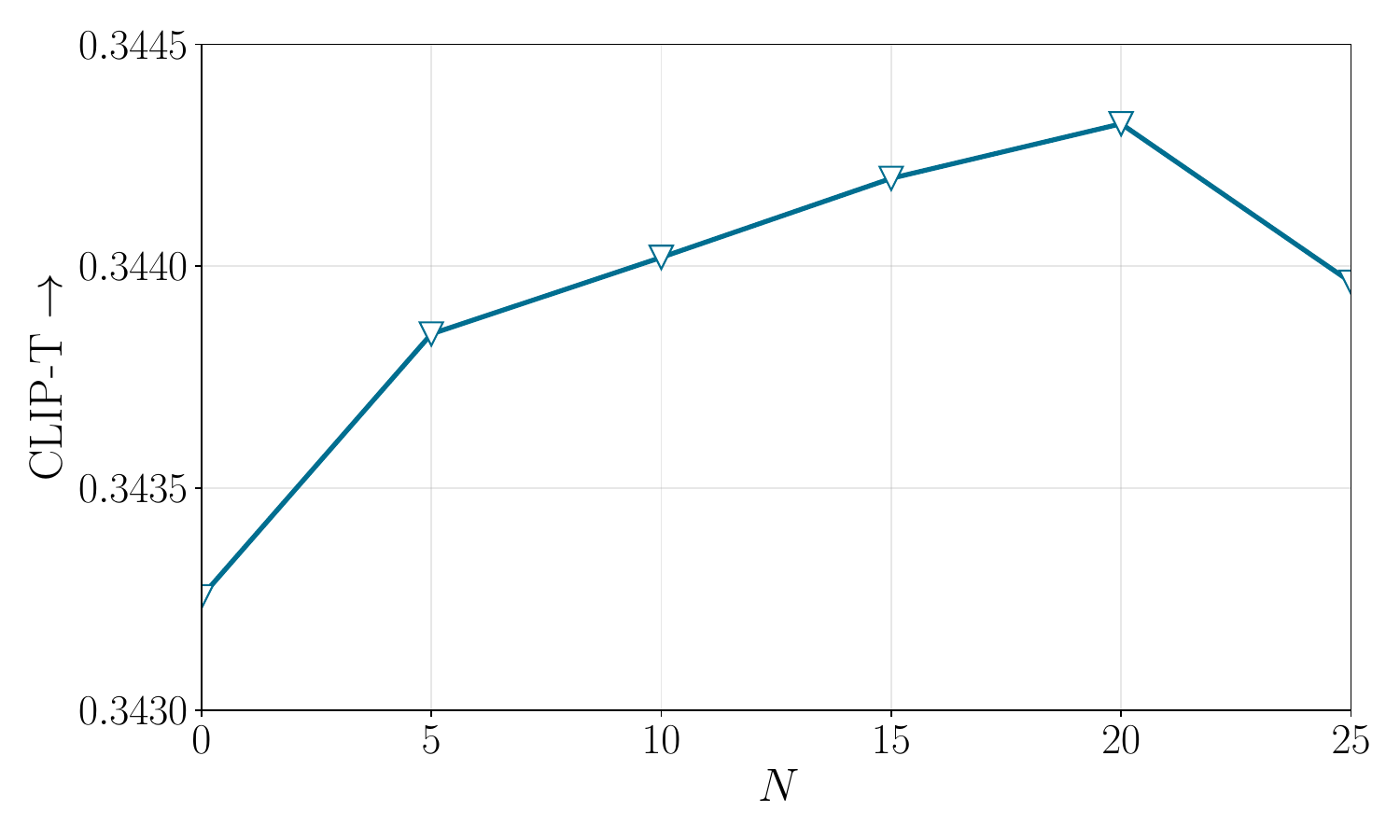}
  \includegraphics[width=.49\textwidth]{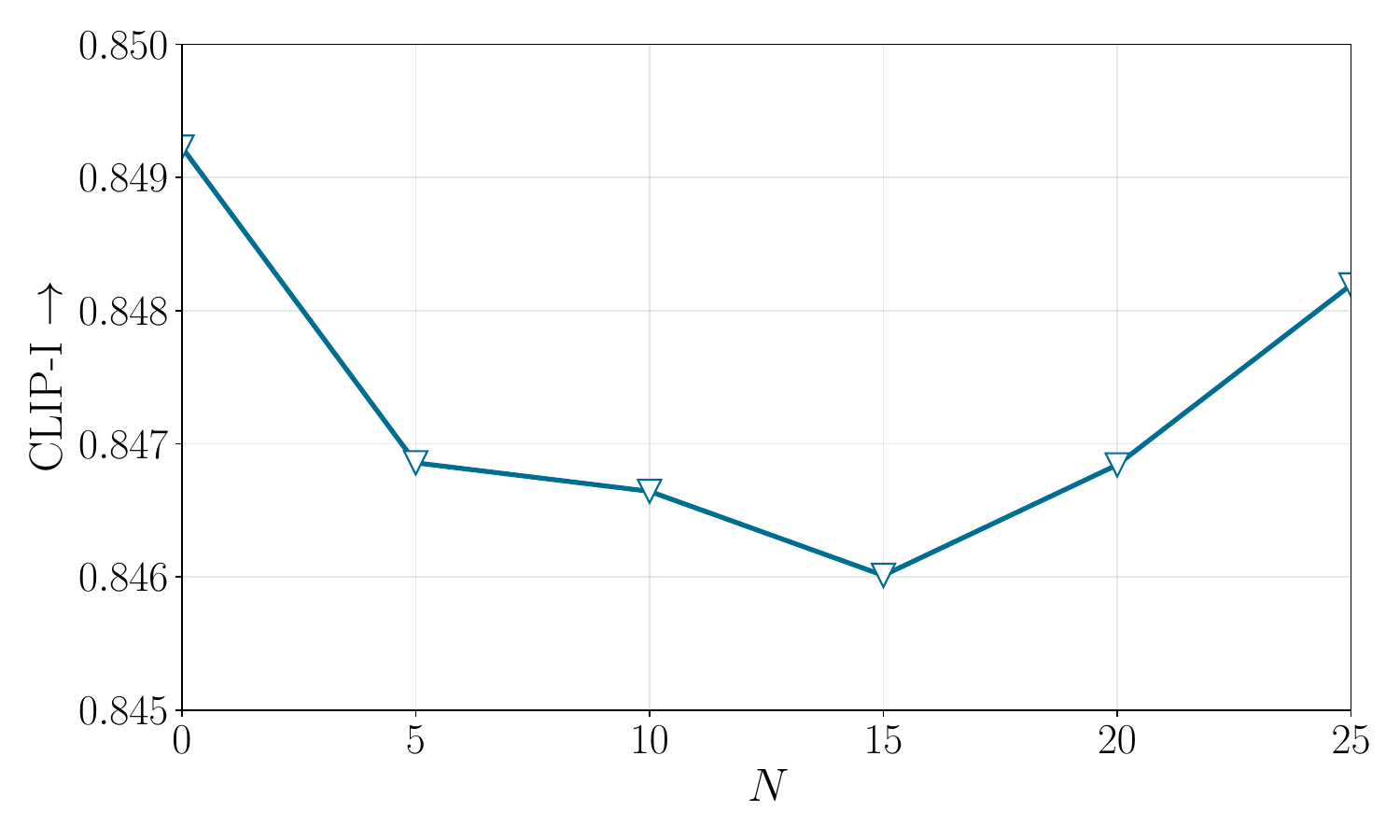}
\end{subfigure}
\hfill
\begin{subfigure}[t]{\linewidth}
  \centering 
  \includegraphics[width=.49\textwidth]{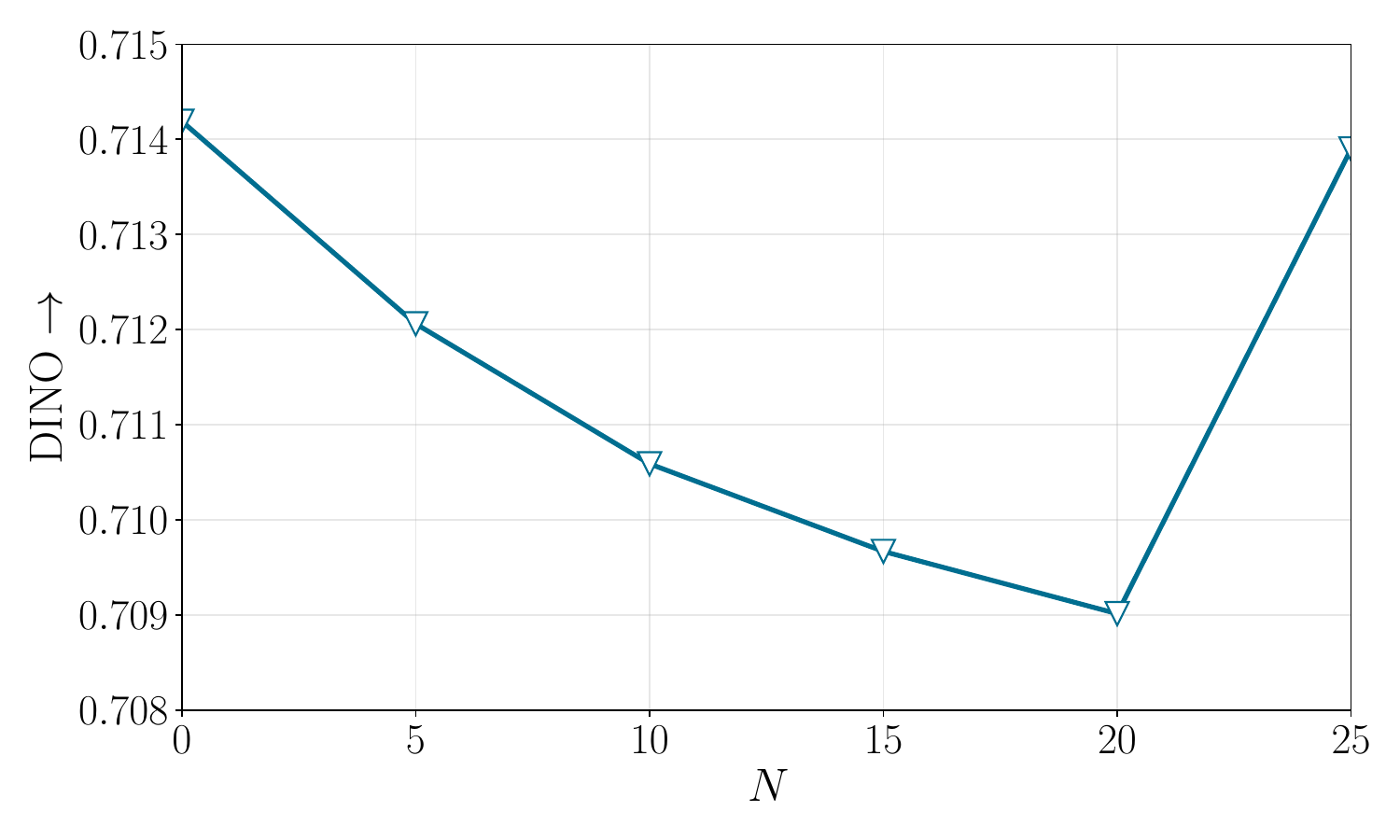}
  \includegraphics[width=.49\textwidth]{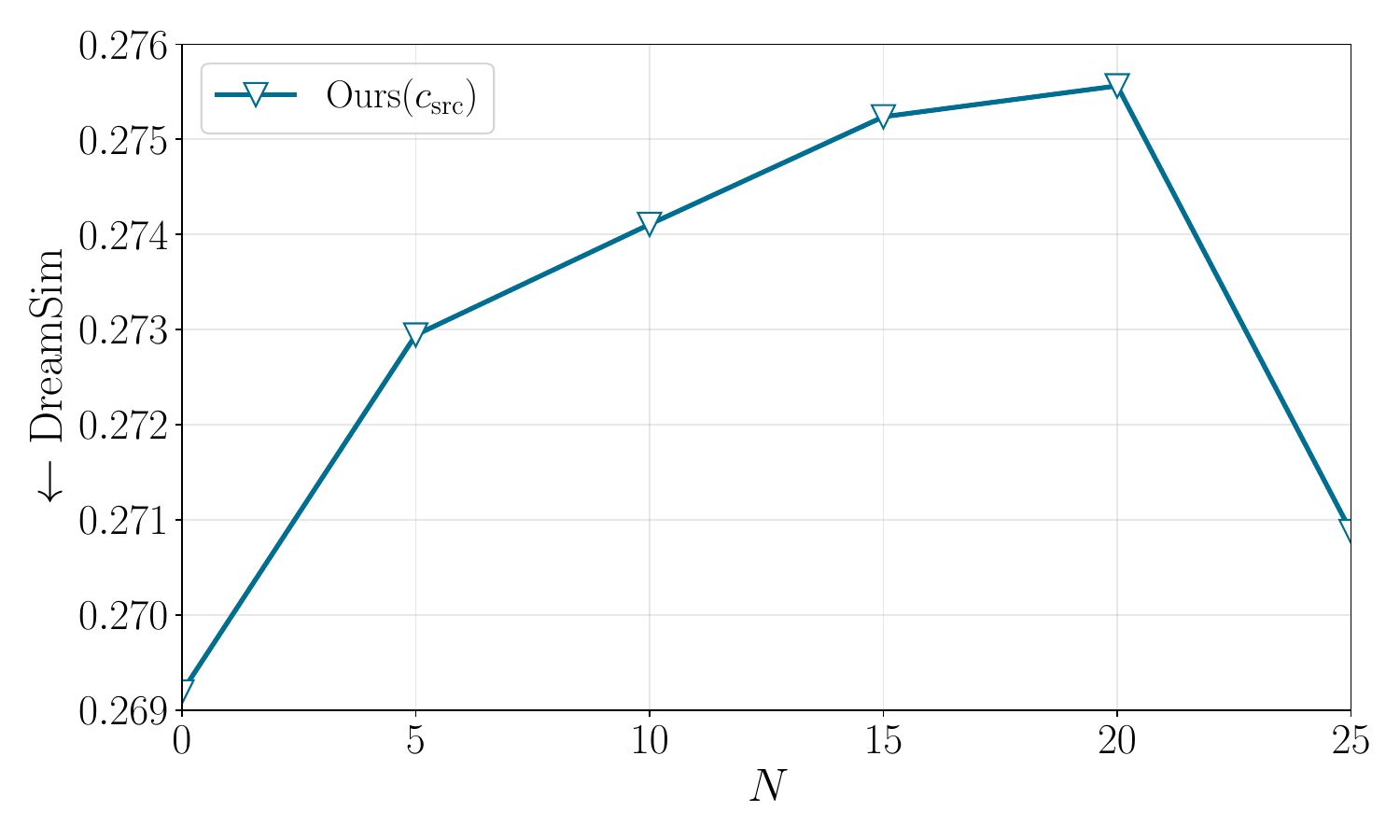}
\end{subfigure}
\caption{\textbf{Editing by inversion (FLUX).} Reconstruction metrics (top) -- pixel-space PSNR, SSIM and LPIPS (left to right), and editing metrics (last two rows) -- CLIP-Text, CLIP-Image, DINOv3 and DreamSim (left to right, top to bottom), as a function of the number of iterations ($N$) by using $c = c_{\text{src}}$ in \eqref{eq:mse_loss}, and in the final sampling step using $c = c_{\text{tar}}$. The dashed orange horizontal line is the average of forwarding the images through the encoder and decoder of the model. Better Inversion (first row) doesn't imply better editing -- no improvement trend is observed in the editing metrics, even as reconstruction quality improves.}
\label{fig:editing_by_inversion}
\end{figure}

In this section we show that even if we have a good inversion method, we do not necessarily get good editing performance with the naive editing-by-inversion paradigm. 
Our definition for good inversion is that we have a noise map $\boldsymbol{z}_t$, such that if we would forward it through the chain of denoisers, $f(\boldsymbol{z}_t, c_{\text{src}})$, we would get almost the same original image $\boldsymbol{z}_0$.
Figure~\ref{fig:editing_by_inversion} demonstrates this using the inversion map obtained by \eqref{eq:mse_update_rule} with $c = c_{\text{src}}$, where in the final iteration we perform sampling with $c = c_{\text{tar}}$. As can be seen, while increasing the number of iterations $N$ leads to better ivnersion (top row), it does not monotonically improve the editing measures (bottom plots). Here, we used $T = 15$ and $n_{\max} = 13$ with the same dataset as that we used for the image editing experiment of Sec.~\ref{sec:experiments}.

\clearpage

%% file: sections/supplementary/limitations.tex
\section{Limitations}
\label{ap:limitations}
\begin{figure}[htbp]
    \centering
    \includegraphics[width=\textwidth]{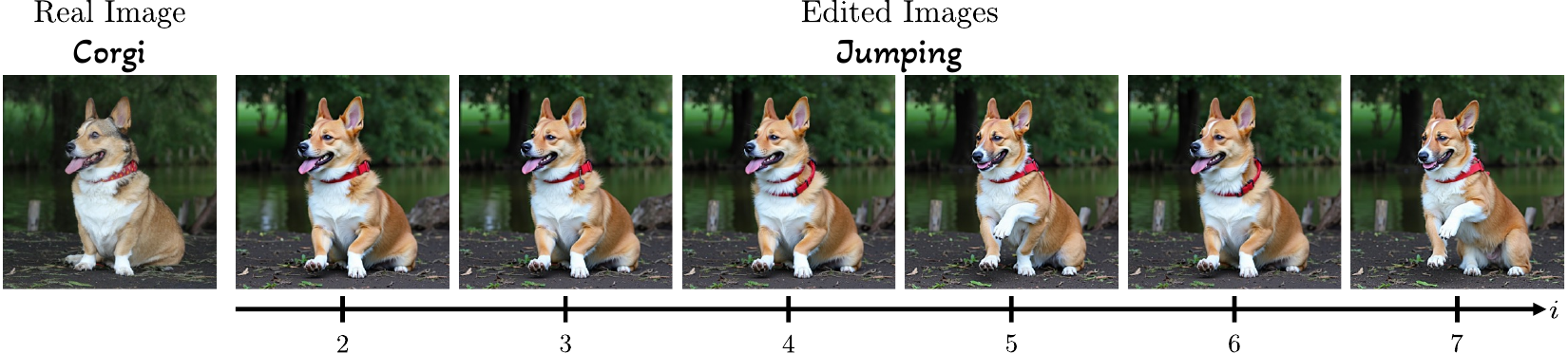}
    \caption{\textbf{Pose limitation}. \oursshort{} often fails in preserving the identity of the edited object for pose editing.}
    \label{fig:pose_limitation}
\vspace{-0.5em}
\end{figure}

Our method, similarly to other structure preserving editing methods, has limited performance when large geometrical changes to the image are required.
For instance, for pose editing, a larger number of optimization steps is required to deviate from the original geometry, and this eventually comes at the cost of diverging from the subject's identity. This can be observed in Fig.~\ref{fig:pose_limitation} where we attempt to make the dog jump. As can be seen, while the dog begins to perform the jumping action in the later optimization steps, its identity begins to drift away from the original. Moreover, as seen in iteration $i = 5$, the edited dog has $5$ legs. This is because our optimization tries to keep the source and target images close in terms of MSE, pushing towards strict alignment, while also trying to adhere to the text, resulting in deviations from the natural image domain. Choosing a more semantic loss function for optimization might remedy these issues, and we leave this for future work. 

\begin{figure}[htbp]
    \centering
    \includegraphics[width=\textwidth]{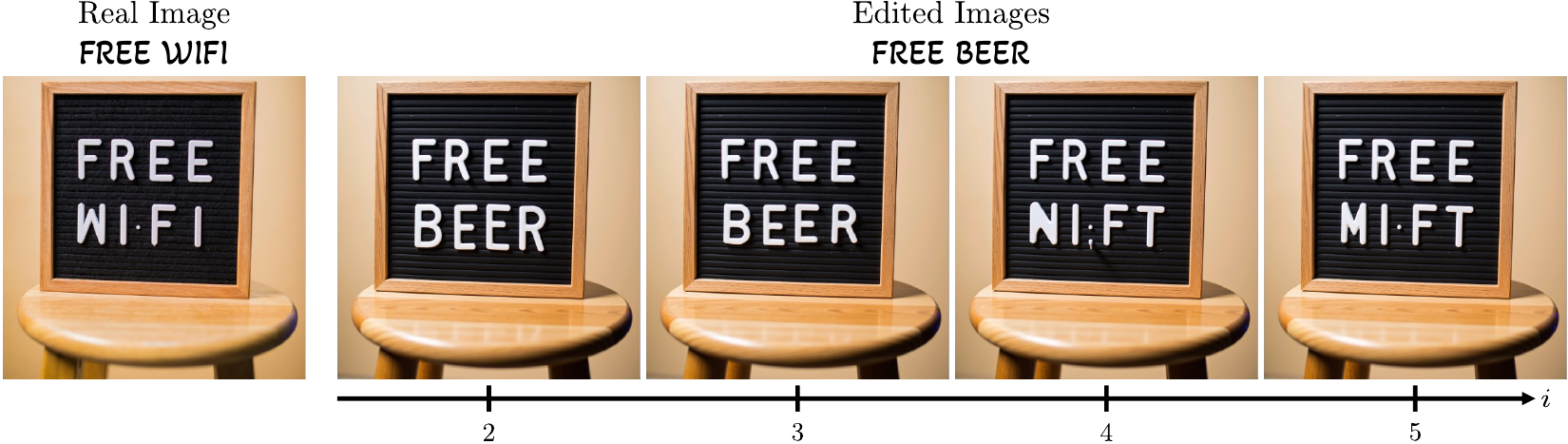}
    \caption{\textbf{Text limitation}. \oursshort{} often fails in text editing as the iterations progress -- it would make the result overly similar to the original image.}
    \label{fig:text_limitation_1}
\vspace{-0.5em}
\end{figure}

\begin{figure}[htbp]
    \centering
    \includegraphics[width=\textwidth]{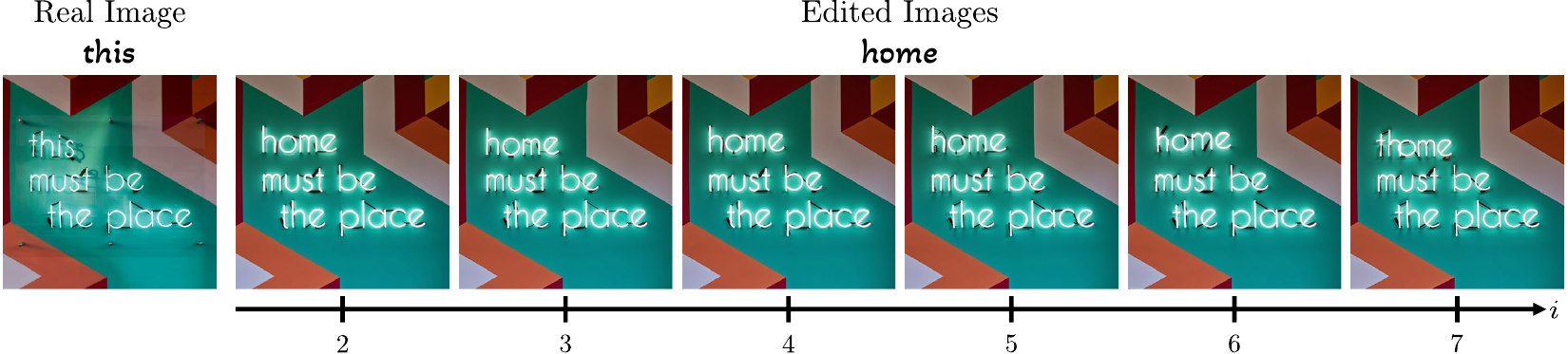}
    \caption{\textbf{Text limitation}. \oursshort{} often fails in text editing, as it struggles to preserve the structure of the original image and adhere to the target text prompt simultaneously.}
    \label{fig:text_limitation_2}
\vspace{-1em}
\end{figure}

Another shortcoming of our method is the ability to edit existing text content in images, such as signs, as demonstrated in Fig.~\ref{fig:text_limitation_1}. We can see that in the first iterations of our optimization process, the method successfully edits the image. Although after several additional iterations, the edit reverts to the original text, failing to adhere to the requested prompt. However, as discussed in Sec.~\ref{sec:method}, the versatility of our algorithm allows the user to choose between all intermediate editing iterations, mediating this downside.
An additional text editing example is presented in Fig.~\ref{fig:text_limitation_2}, where similar behavior is observed.

